%% file: iclr2025_conference.tex
\documentclass{article} 
\usepackage{iclr2025_conference,times}

\input{math_commands.tex}

\usepackage{url}
\usepackage{marvosym}
\usepackage[colorlinks,
            linkcolor=magenta,
            anchorcolor=blue,
            citecolor=gray
            ]{hyperref}
\definecolor{emphypurple}{rgb}{0.302, 0.055, 0.659}
\usepackage{hyperref}
\usepackage{url}
\usepackage{hyperref}
\usepackage{xcolor}
\usepackage{pifont}
\usepackage{soul}
\usepackage{url}
\usepackage[T1]{fontenc}
\usepackage{siunitx}
\usepackage{graphicx}
\usepackage{float}
\usepackage{multirow}
\usepackage{color}
\usepackage{wrapfig}
\usepackage{booktabs}
\usepackage{amssymb}
\usepackage{subfigure}
\usepackage{listings}
\usepackage{makecell}
\usepackage{amssymb,mathrsfs,amsmath}
\usepackage{amsmath, bm}
\usepackage{colortbl}
\definecolor{xred}{HTML}{BD4242}
\definecolor{xblue}{HTML}{C7A085}
\definecolor{xblues}{HTML}{52B256}
\definecolor{xgreen}{HTML}{52B256}
\definecolor{xpurple}{HTML}{7F52B2}
\definecolor{xorange}{HTML}{FD9337}
\definecolor{xdotted}{HTML}{999999}
\definecolor{xgray}{HTML}{777777}
\definecolor{xcyan}{HTML}{80F5DC}
\definecolor{xpink}{HTML}{f690ea}
\definecolor{xgraycyan}{HTML}{82bceb}
\usepackage{tikz}
\usetikzlibrary{positioning, calc}
\usepackage[most]{tcolorbox}

\tcbset{
	common/.style n args={2}{
		colframe={#1},
		colback={#1!5},
		colbacktitle={#1},
		lower separated=false,
		coltitle=white,
		boxrule=1pt,
		fonttitle=\bfseries,
		enhanced,
		breakable,
		top=8pt,
		before skip=8pt,
		attach boxed title to top left={
			yshift=-0.25cm,
			xshift=0.38cm,
		},
		boxed title style={
			boxrule=0pt,
			colframe=white,
			arc=5pt,
			outer arc=4pt,
		},
		separator sign={~~},
		overlay unbroken and last={
			\node[text=white, align=right, rounded corners, fill=#1, xshift=-4mm, minimum height=6mm, anchor=east] at (frame.south east) {#2};
		}
	},
	defstyle/.style={
		common={xpurple}{$\bm{\pi}$},
	},
	theoremstyle/.style={
		common={xgraycyan}{$\star$},
	},
	proofstyle/.style={
		common={xgreen}{\textbf{QED}},
	},
	examplestyle/.style={
		common={xblue}{$\bigstar$},
	},
    examplestyles/.style={
		common={xblues}{$\bigstar$},
	},
	notestyle/.style={
		common={xred}{\textbf{!}},
	},
	challangestyle/.style={
		common={xorange}{\textbf{?}},
	},
}

\newtcbtheorem{definition}{Definition}{defstyle}{def}
\newtcbtheorem{theorem}{Theorem}{theoremstyle}{theorem}
\newtcbtheorem{proof}{Proof}{proofstyle}{proof}
\newtcbtheorem{example}{Example}{examplestyle}{example}
\newtcbtheorem{examples}{Examples}{examplestyle}{example}
\newtcbtheorem{note}{Note}{notestyle}{note}
\newtcbtheorem{challange}{Challange}{challangestyle}{challange}

\usepackage[utf8]{inputenc} 
\usepackage[T1]{fontenc}    
\usepackage{hyperref}       
\usepackage{url}            
\usepackage{booktabs}       
\usepackage{amsfonts}       
\usepackage{nicefrac}       
\usepackage{microtype}      
\usepackage{xcolor}         
\usepackage[utf8]{inputenc} 
\usepackage[T1]{fontenc}    
\usepackage{hyperref}       
\usepackage{url}            
\usepackage{booktabs}       
\usepackage{amsfonts}       
\usepackage{nicefrac}       
\usepackage{microtype}      
\usepackage{xcolor}         
\usepackage{csquotes}
\usepackage{xcolor}

\newcommand{\method}{PAS\space}

\title{Personality Alignment of Large Language Models}

\author{Minjun Zhu$^{1,2}$, Yixuan Weng$^{2}$, \textbf{Linyi Yang$^{3,4}$}, \textbf{Yue Zhang$^{2}\thanks{Correspondence to: Yue Zhang (zhangyue@westlake.edu.cn)}$} \\ $^1$Zhejiang University  $^2$School of Engineering, Westlake University \\$^{3}$University College London 
$^{4}$Huawei Noah’s Ark Lab\\
\texttt{zhuminjun,zhangyue@westlake.edu.cn}, \texttt{wengsyx@gmail.com}
\\ 
\texttt{yanglinyiucd@gmail.com}}

\iclrfinalcopy 
\begin{document}

\maketitle

\begin{abstract}
Aligning large language models (LLMs) typically aim to reflect general human values and behaviors, but they often fail to capture the unique characteristics and preferences of individual users. To address this gap, we introduce the concept of Personality Alignment. This approach tailors LLMs' responses and decisions to match the specific preferences of individual users or closely related groups. Inspired by psychometrics, we created the Personality Alignment with Personality Inventories (PAPI) dataset, which includes data from over 320,000 real subjects across multiple personality assessments - including both the Big Five Personality Factors and Dark Triad traits. This comprehensive dataset enables quantitative evaluation of LLMs' alignment capabilities across both positive and potentially problematic personality dimensions. Recognizing the challenges of personality alignments—such as limited personal data, diverse preferences, and scalability requirements—we developed an activation intervention optimization method. This method enhances LLMs' ability to efficiently align with individual behavioral preferences using minimal data and computational resources. Remarkably, our method, PAS, achieves superior performance while requiring only 1/5 of the optimization time compared to DPO, offering practical value for personality alignment. Our work paves the way for future AI systems to make decisions and reason in truly personality ways, enhancing the relevance and meaning of AI interactions for each user and advancing human-centered artificial intelligence. The dataset and code are released at \url{https://github.com/zhu-minjun/PAlign}.
\end{abstract}

\section{Introduction}

\begin{figure}[t]
\begin{center}
	\includegraphics[width=\textwidth]{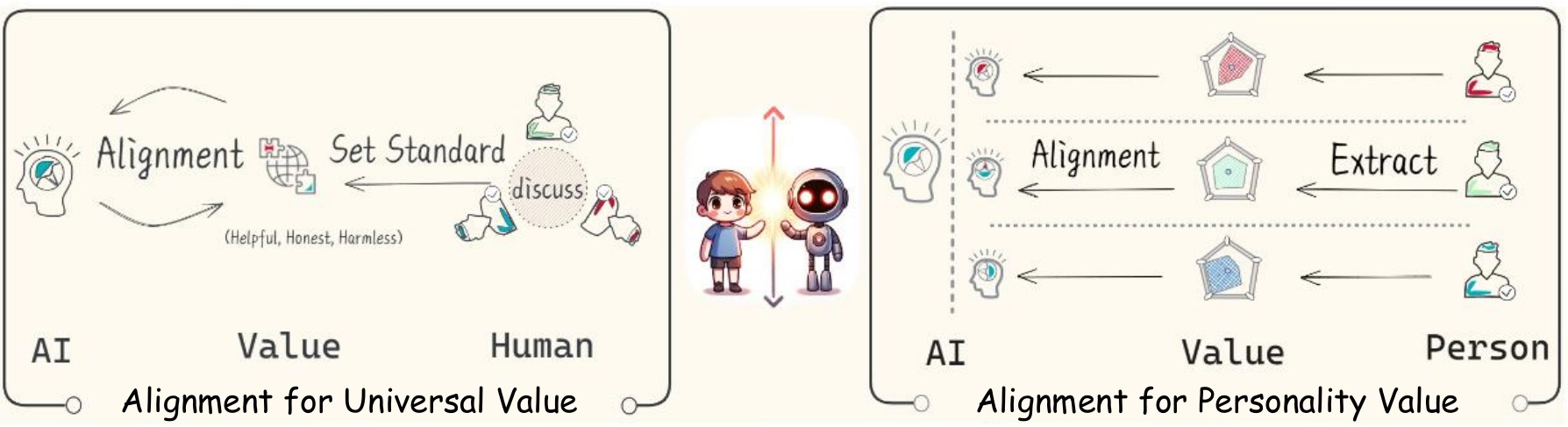}
\end{center}
 \vspace{-0.56cm}
 \caption{On the left, AI aligns to broad human values like helpfulness, honesty, and harmlessness using a standard set. On the right, the focus is on aligning AI behavior with individual users' specific traits and preferences, using detailed profiles to reflect unique personal values.}
 \vspace{-0.4cm}
 \label{fig:first}
\end{figure}

The alignment with human preferences has been a focal point of both theoretical and applied research for AI systems \citep{shen2023large,ji2023ai,hendrycks2023overview}. However, existing alignment approaches often treat society values as a monolithic construct, lacking the granularity needed to cater to individual differences. This broad approach may overlook the nuanced preference of individual users or closely related groups, leading to a one-size-fits-all model that fails to effectively serve diverse user needs~\citep{fernandes2023bridging,carranza2023deceptive,vanWynsberghe2021SustainableAA}. This gap is particularly critical as personality alignment can enhance user satisfaction, trust, and engagement by making interactions more meaningful. As shown in Figure \ref{fig:first}, we propose "\textbf{Personality Alignment}": 

\begin{displayquote}
\textit{Aligning AI behavior to match an individual's unique preferences and priorities of specific individuals, mirroring their behavior, thinking, and decision-making. It ensures AI responses not only are precise but also resonate with the user's traits, understanding their communication nuances and preferences.}
\end{displayquote}

As large language models (LLMs) advance \citep{solaiman2019release,farina2023chatgpt}, AI systems increasingly exhibit societal values and human-like personality traits \citep{ng2024llms,bai2022constitutional,durmus2023towards,zhang2024heterogeneous,huang2024on,jang2023personalizedsoupspersonalizedlarge}. The Role-playing tasks \citep{li2023chatharuhi,salemi2024lamp,gupta2024bias,yu2024neeko} highlight this trend. However, human preferences are complex, every individual has unique preferences and needs that must be considered to effectively meet user expectations in real-world applications \citep{pfeifer2006body}. For example, in customer service, some users prioritize politeness, while others prefer efficiency. Therefore, personalization is crucial for aligning AI systems with user preferences. Traditional Role-Playing Agents simulate specific professions or roles, rather than deeply aligning with individual preferences \citep{chen2024persona,Kulveit2018,hubinger2019risks,stray2020aligning}. The lack of large-scale, theoretically supported datasets for aligning AI with diverse human personalities and preferences underscores the need for further research in personality and personality alignment \citep{Jang2023PersonalizedSP,Lou2024SPOMP,Li2024PersonalizedLM}.

There remains a practical challenge: aligning a model like Llama-2 typically requires around 3 million sentences \citep{touvron2023llama}. For personality alignment, collecting such an enormous set of behavioral data from individuals is both expensive and unrealistic. Instead, we propose a training-free personal-alignment paradigm, providing over 320,000 unique preference samples across two widely-used personality assessments. Specifically, we utilize the IPIP-NEO-120 questionnaire \citep{johnson2014measuring}, which quantifies individual behavior patterns across five dimensions: Openness, Conscientiousness, Extraversion, Agreeableness, and Neuroticism \citep{de2000big,roccas2002big}. This questionnaire enables subjects to provide detailed personality measures within 10-20 minutes \citep{aluja2004relationships}. We collect preference data from over 300,000 subjects completing both IPIP-NEO-120 and IPIP-NEO-300 \citep{goldberg1999broad} questionnaires. Additionally, we incorporate 18,192 independent samples from the Short Dark Triad \citep{jones2014introducing} assessment, which measures three negative personality traits (Machiavellianism, Narcissism, and Psychopathy) through 27 questions. We manipulate these datasets to build the Personality Alignment with Personality Inventories (PAPI) dataset. For evaluation, we selected 300 representative subjects from each assessment as independent test sets, establishing a comprehensive framework for assessing LLMs' personality alignment capabilities across both positive and negative personality dimensions.

To manage the behaviors of millions, the system must be scalable, ensuring quality and efficient alignment. Traditional RLHF techniques, which lack data and computational efficiency \citep{yao2023deepspeedchat}, are unsuitable for this task. Similarly, Prompt Engineering, though efficiently, often varies too greatly between models to be reliable for precise alignment \citep{zhou2023large}. While recent activation-based approaches have shown promise in steering model behaviors \citep{turner2023activation, li2024inference}, they typically focus on maximizing a single objective (e.g., safety or truthfulness). In contrast, we propose a more nuanced approach that simultaneously searches for optimal activation value offset directions and distances across multiple personality dimensions, ensuring appropriate alignment levels. This balanced activation steering is crucial for maintaining authentic personality expression while achieving desired alignment. We propose a simple and effective personality alignment method: Personality Activation Search. PAS identifies the most effective attention heads in language models (LLMs) using established behavioral patterns that correlate with the five major personality traits. It then adjusts the activations in these specific directions and optimizes the magnitude of movement to achieve the best results. This approach is computationally efficient because it doesn't rely on backpropagation to modify model parameters. Instead, PAS offers a minimally invasive intervention, resulting in greater data efficiency and more controlled personality alignment than traditional Prompt Engineering \citep{jiang2024evaluating}.

We present a novel systematic and comprehensive personality alignment framework that enables LLMs to understand and align precisely with specific values, preferences, and behaviors required. Compared to aligning with universal values, personality alignment requires LLMs to align with individual values, preferences, and behaviors, which ensures that the LLMs can accurately match or mimic the specific individual characteristics that users need and pursue. To achieve this goal, we introduce the PAPI dataset, which covers over 320,000 subjects of different ages, genders, and periods using multiple-choice questions to comprehensively measure both positive and negative personality traits. Based on that, we propose the PAS method, which aims to address the challenges of scarce individual characteristic data and high scalability in personality alignment. Experiments have shown its high alignment efficiency compared to DPO and PPO \citep{schulman2017proximal,rafailov2023direct} where it requires only 1/6 of the time to achieve superior performance. It even outperforms the GPT-4o model using PAS based on the Llama-3-8B model.

\section{Related Work}
\textbf{Alignment with language models.} The concept of integrating human values into AI systems, initially proposed by Dewey \cite{dewey2011learning}, underscores the necessity of value learning, which entails the AI's ability to process and prioritize a broad spectrum of human values and preferences. In the context of LLMs—functioning as high-performance assistants—aligning these models with human values is crucial, especially since they operate in complex reasoning tasks across various domains \citep{hadar2024assessing, wei2022chain, wei2022emergent, weng-etal-2023-large}. Methods such as RLHF \citep{NEURIPS2020_1f89885d,bai2022training,dong2023raft} and RLAIF \citep{lee2023rlaif,sharma2024critical} seek to align LLMs by employing strategies like PPO \citep{schulman2017proximal} and DPO \citep{rafailov2023direct}. These approaches focus on maximizing cumulative rewards that are closely aligned with predefined human values \citep{azar2023general,lin2023unlocking,dai2024safe,swamy2024minimaximalist,dai2024safe}. Contrary to these broad approaches, our research focuses on personality alignment, which tailors the LLMs' behavior to each individual's complex value set, thereby enhancing the model's capacity to understand and reflect individual human values.

\textbf{Preferences of language models.}  The community of LLMs has started using personality testing tools for both qualitative and quantitative evaluations \citep{Hoover2019MoralFT,jiang2022machines,Qiu_2022,xu2023cvalues,salemi2024lamp,wang2024incharacter}. For example, the Machine Personality Inventory (MPI) \cite{jiang2024evaluating}, utilizes the Big-Five Personality Factors to standardize the assessment of LLMs. This approach analyzes the behaviors of LLMs across the five dimensions of the Big-Five, providing a quantitative interpretation of the values and personality traits of LLMs. Such tools are grounded in widely accepted theories, including the Big-Five and the 16 Personality Factors \citep{cattell2008sixteen}, helping to contextualize LLMs' personality traits in terms of human individual differences \citep{gupta2024selfassessment,wang2023rolellm,aher2023using,liu2024dynamic,li2023chatharuhi}. These studies indicate that aligned LLMs share similar values with humans \citep{mehta2020bottom,pan2023llms,mao2023editing,go2023compositional,chen2024persona}. \cite{Jang2023PersonalizedSP} has also mentioned the similar term "personality alignment", which considers a limited scope of preference personalization \citep{Wang2024ArithmeticCO,Zhuang2024HYDRAMF,Zeng2023OnDP}, focusing on only 8 different conflicting user preferences derived from combinations of three dimensions: Expertise (elementary vs. expert), Informativeness (concise vs. informative), and Style (friendly vs. unfriendly). Our work extends these theories and methods by leveraging psychological measurement to assess the value alignment gap between aligned LLMs and individual users. We aim to understand how closely aligned LLMs are with individual values through personality and personalized adjustments.

\textbf{Interventions Activate.} Activation interventions in neural networks trigger specific changes in internal activations \citep{olsson2022incontext,wu2024pyvene} and have emerged as a pivotal tool in model robustness \citep{he2019parametric}, editing \citep{meng2022locating}, circuit finding \citep{goldowsky2023localizing}, and knowledge tracing \citep{geva-etal-2023-dissecting}. These interventions are advantageous due to their adjustable nature and minimal invasiveness \citep{wu2024interpretability}. Prior research has demonstrated that activation interventions can effectively identify the crucial heads and directions in a model's internal representations \citep{burns2022discovering,Liu2024DecodingtimeRO}, which offers better data efficiency and faster optimization than traditional reinforcement learning approaches \citep{turner2023activation,rimsky2023steering,Shi2024DecodingTimeLM,weng2024controllm}. Prior activation steering work has focused on maximizing single objectives like safety \citep{wang2024trojanactivationattackredteaming,zheng2024promptdrivensafeguardinglargelanguage} or truthfulness \citep{li2024inference}.  Building on these findings, our method enhances scalability by pinpointing value directions of different individuals and calculating optimal offset distances, aiming to achieve appropriate alignment for each dimension while maintaining model capabilities.

\section{Personality Alignment Dataset Construction}
\vspace{-0.1cm}

\begin{figure}[h]
\vspace{-0.1cm}
\begin{center}
	\includegraphics[width=\textwidth]{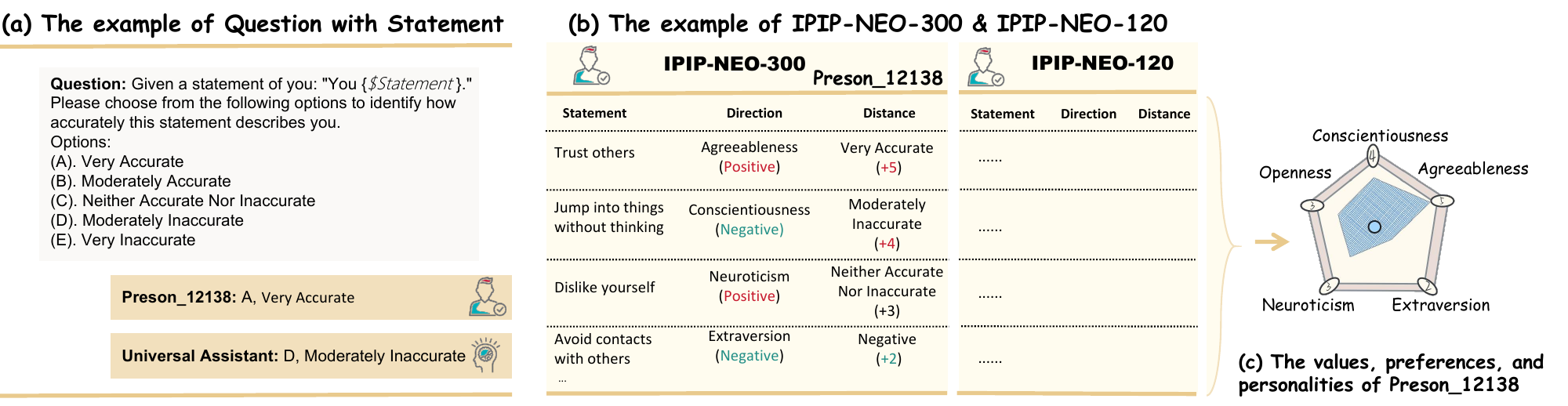}
\end{center}
 \vspace{-0.5cm}
 \caption{Overview of the PAPI dataset. (a) Illustrates the comparison between the subject's self-assessment and the AI's assessment of a specific question. (b) Shows an example of the IPIP-NEO-120 and IPIP-NEO-300 questionnaire responses. (c) Depicts the Big Five personality traits profile. }
 \vspace{-0.3cm}
 \label{fig:2}
\end{figure}

\textbf{Motivation.} To systematically evaluate the consistency between Personality-aligned LLMs and human preferences, we create the large-scale Personality Alignment with Personality Inventories (PAPI) dataset, comprising 325,505 samples across multiple-choice questionnaires. The dataset uniquely combines two complementary personality assessment systems: 307,313 samples from the Big Five personality traits (IPIP-NEO assessments) and 18,192 samples from the Dark Triad inventory. While the Big Five (IPIP) samples measure adaptive personality traits through IPIP-NEO-120 and IPIP-NEO-300 questionnaires, the Dark Triad samples assess subclinical maladaptive traits (Machiavellianism, Narcissism, and Psychopathy) through 27 targeted questions. This comprehensive approach captures both healthy personality dimensions and potentially problematic traits. As shown in Figure \ref{fig:2}, subjects choose behaviors similar to their own (a), yielding questionnaire responses (b) that collectively form detailed personality profiles (c).

\textbf{Question Type.} Our dataset contains two independent questionnaire systems, each using a five-point scale from "Very Accurate" to "Very Inaccurate." The first system comprises IPIP questions associated with the Big Five personality dimensions. We use each subject's IPIP-NEO-120 questionnaire responses to predict their IPIP-NEO-300 answers, testing the model's ability to generalize from known personality patterns to broader behavioral descriptions. The second system consists of Dark Triad assessments, where we use 18 questions (6 each for Machiavellianism, Narcissism, and Psychopathy) to align with and predict responses to the remaining 9 questions (3 per trait). Both systems evaluate the stability of individual personality tendencies and the model's capacity to capture and generalize personality patterns, though they operate independently on distinct trait dimensions.

\begin{wrapfigure}{r}{0.45\textwidth}
  \begin{center}
      \vspace{-0.7cm}
    \includegraphics[width=0.41\textwidth]{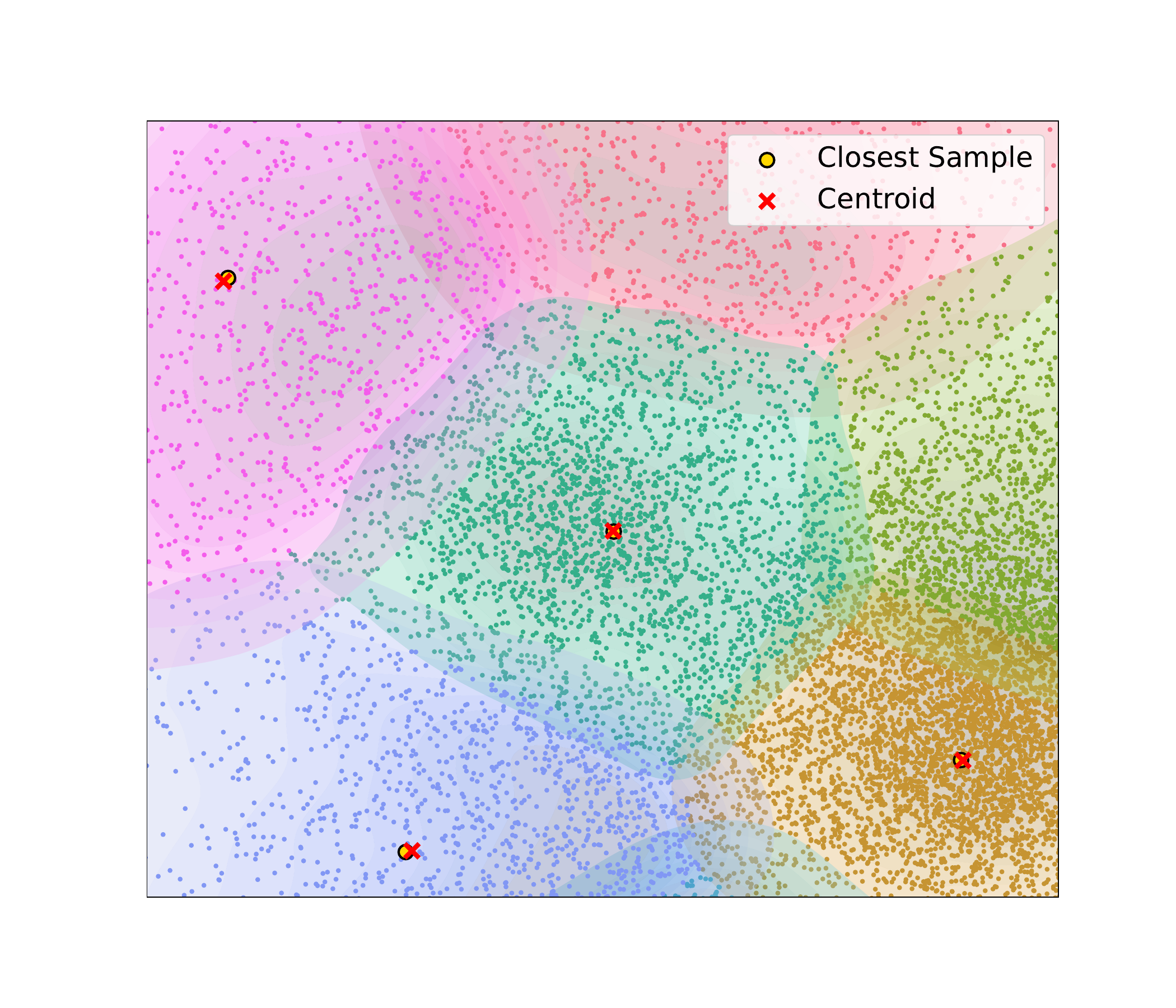}
  \end{center}
    \vspace{-1.cm}
  \caption{Visualization of the K-Means Clustering on the PAPI Dataset. The centroids are marked in red, demonstrating the central point of each cluster, while the closest samples to these centroids are highlighted in gold.}
  \label{figure:2}
\vspace{-0.3cm}
\end{wrapfigure}

\textbf{Data Collection.} We have collected 307,313 samples based on the IPIP-NEO-120 and IPIP-NEO-300 from the International Personality Item Pool (IPIP)\footnote{\url{https://ipip.ori.org/index.htm}}. This includes approximately 60\% (185,149) female and 40\% (122,164) male participants, covering ages from 10 to 99 years (average age 25.19 years). The data were collected through online surveys on a website, requiring participants to provide genuine feedback on 120 \& 300 behavioral questions. Participants are volunteers from various industries worldwide, including the US, UK, France, India, and China, with tests conducted between 1998 and 2019. Each participant had to actively acknowledge that careless responses would invalidate the usefulness. Each survey took between 30 to 50 minutes, was anonymous, and collected no traceable or socioeconomic status-related data. We have permission from the administrators to use IPIP items, scales, and inventories for any purpose, commercial or non-commercial. Additionally, we collected 18,192 Dark Triad samples\footnote{\url{https://openpsychometrics.org}} using the same collection pipeline and quality control procedures.

\textbf{Data Processing.} We apply K-Means clustering separately to the IPIP and Dark Triad datasets. From the IPIP dataset (307,313 samples), we select 300 representative clusters as the first part of our Test-Set. Similarly, we cluster the Dark Triad dataset (18,192 samples) and select 300 representative samples, forming the second part of our \textbf{Test-Set}. The remaining data constitutes our \textbf{Dev-Set}. The large-scale Dev-Set captures personality variations across different ages (from teenagers to seniors), genders, cultural backgrounds (spanning multiple countries and regions), and identity groups. This comprehensive representation is crucial for developing inclusive alignment methods that respect and reflect the full spectrum of human personality diversit. Sample selection minimizes Euclidean distance to cluster centroids: $\min_{x \in S_k} | x - \mu_k |_2$, where $\mu_k$ is the centroid of cluster $k$, and $x$ is a sample within cluster $k$ ($S_k$), aiming to optimize cluster representativeness by minimizing the distance between samples and their respective centroids.

\textbf{Evaluation Methods.} Inspired by the MPI \citep{jiang2024evaluating}, we set up a behavioral difference score as an evaluation metric. Using a psychological measurement approach, we establish a simple scoring function S: for positively correlated questions, we assign scores from 5 to 1 from ``Very Accurate'' to ``Very Inaccurate''. We expect the aligned language model scores to be close to a specific individual's scores, hence we use an absolute value method for measurement. We test each trait $d$ in the test section, with each participant's behavioral difference Aligned Score calculated as follows:

\begin{equation}
\small
\text{Aligned Score}_d = \frac{1}{M} \sum_{i=1}^M \left( \frac{1}{N_{d,i}} \sum_{\kappa \in D^{Test}_{d,i}} \left| f\left(\text{LLM}(\kappa, \text{template})\right) - f\left(\text{Person}(\kappa, \text{template})\right) \right| \right),
\end{equation}

In this formula, $D^{Test}_{d,i}$ denotes the set of behaviors related to trait $d$ in the test section for subject $i$. We use the given template to obtain responses from both the LLM and the Person, and then calculate the absolute difference between the two. $N_{d,i}$ represents the count of these behaviors. Each $\kappa$ is an independent behavior sample from this set, with the LLM’s responses evaluated against those in $D^{Train}$. The $f(.)$ represents the scoring function for multiple-choice questions.  We calculated the deviation distance between the two using the absolute value method and computed the average deviation distance for all behaviors. The range for Score is from 0 to 4, where 0 means perfect alignment with individual behavior.

\section{Personality Activation Search}

\begin{figure}[t]
\begin{center}
	\includegraphics[width=\textwidth]{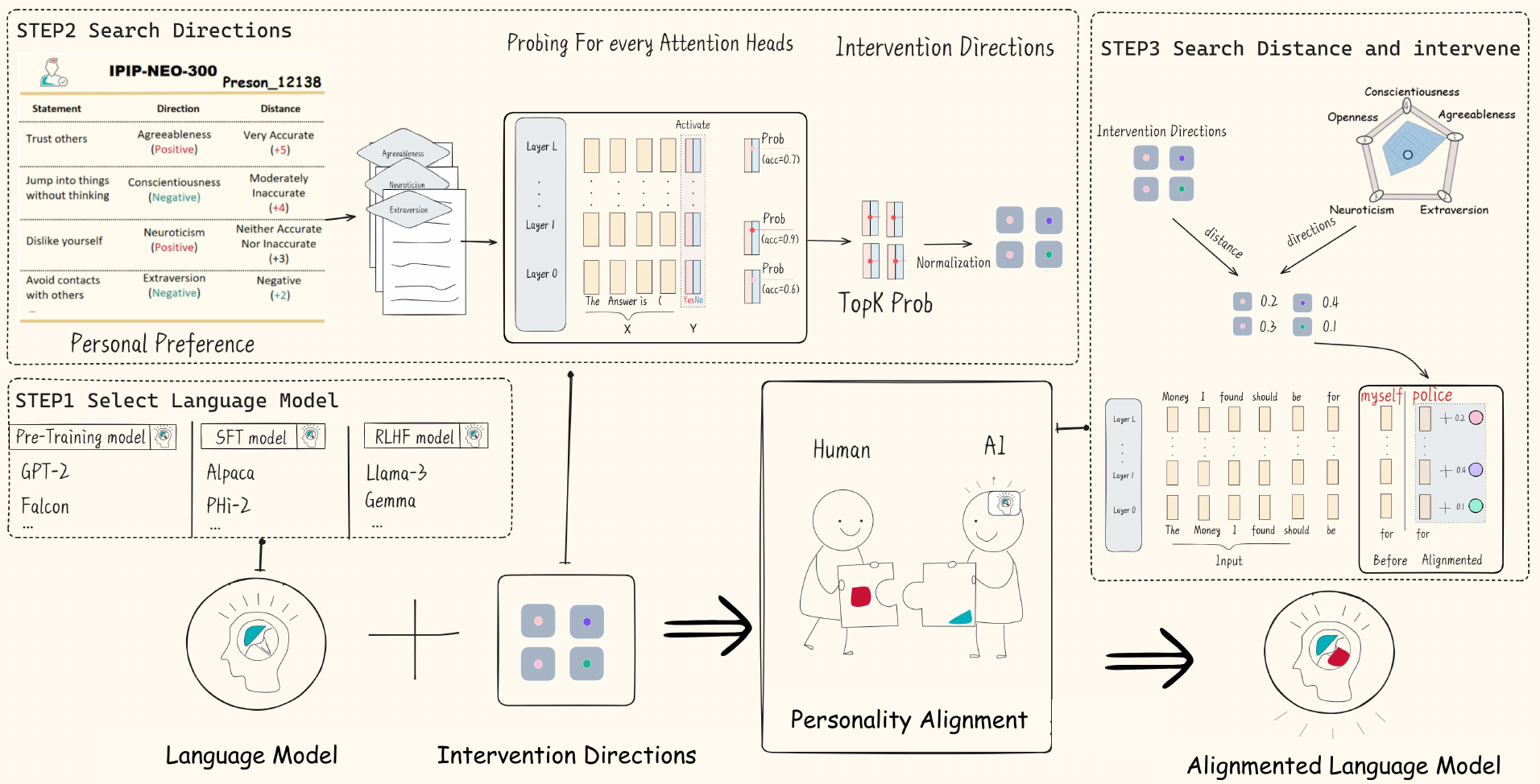}
 \vspace{-0.75cm}
\end{center}
 \caption{Overview of the PAS Process. Step 1 involves selecting a LLM. Step 2 includes creating and activating personality responses using the answer of questionnaire, the subject's answer in the PAPI dataset, and probing each attention head. Step 3 integrates these activations for personality alignment, adjusting model outputs to reflect individual user traits, resulting in an aligned LLM.}
 \vspace{-0.3cm}
 \label{fig:4}
\end{figure}
Popular alignment techniques like PPO, DPO, and IPO strive to align language models with desired behaviors, but personalizing LLMs for individual preferences without extensive training presents a significant challenge. To address this, we've developed Personality-Activation Search (PAS), an efficient method that fine-tunes model activations to closely match human preferences. PAS identifies key activations within the model that significantly impact Personality and personalized behaviors and adjusts these activations toward specific preferences.

PAS works by identifying multiple directions in the transformer's activation space that satisfy personality attributes and then intervening along these directions to adjust the activation values, achieving minimally invasive alignment. The entire process does not require changing the model's weights or backpropagation. It only requires multiple forward passes to determine the offset direction and distance, resulting in a high optimization efficiency.

\subsection{Language Models}
In our research, we focus on a transformer-based language model which may be in various stages, including pre-training, supervised fine-tuning, or undergoing RLHF. The transformer model primarily consists of several layers, each equipped with a multi-head attention (MHA) mechanism and a multilayer perceptron (MLP) \citep{NIPS2017_3f5ee243}.

During inference, input tokens are encoded to high-dimensional embeddings, where in each transformer layer, the MHA process is defined as $x_{l+1} = x_l + \sum_{h=1}^H Q_l^h \operatorname{Att}_l^h(P_l^h x_l)$, with H being the number of heads, $x_l \in \mathcal{R}^{DH}$ representing the activations, and $P_l^h \in \mathcal{R}^{{D}\times{DH}}$ and $Q_l^h \in \mathcal{R}^{{DH}\times{D}}$ projecting activations into and back from specific headspace.

While previous methods like Inference-Time Intervention (ITI) adjust language models by applying a fixed, large shift in the activation space to influence output biases, this approach is inadequate for addressing alignment issues, especially when personalization is required. Fixed shifts can be too coarse and may not capture the nuances of individual preferences. To overcome this limitation, we have enhanced the intervention strategy by dynamically adjusting the activation distances based on personal preference vectors. Specifically, we introduce these vectors into the residual stream between the Attention mechanism and the projection matrix $Q_{l}^h$. This allows us to fine-tune the model's activations in a way that aligns with specific user traits and preferences, facilitating personalized alignment without altering the model's weights or relying on large, inflexible shifts.


\subsection{Search the Directions for Activation Intervention}

To achieve personality alignment, we trained ``probes'' to get activations as the direction of personal preference. It is started by selecting statements from the training dataset of each subject, formatted as positive samples: ``\textit{Question: Given a statement of you: `\$Statement', Do you agree? Answer: Yes}'', and negative samples: ``\textit{Question: Given a statement of you: `\$Statement', Do you agree? Answer: No}''. Activations from each head are retained and split into a training set (60\%) and a validation set (40\%). These activations serve as features for the probe (Equation \ref{eq:2}) of per attention head.
\begin{equation}
p_\theta(x_l^h) = \operatorname{sigmoid}(\langle \theta, x_l^h \rangle),
\label{eq:2}
\end{equation}
where the \(x_l^h \in \mathcal{R}^{D}\), and \(\theta \in \mathcal{R}^{D}\) is the weights of h-heads in layer l in the residual stream.

We identify the most influential heads based on their predictive accuracy during validation. Notably, it is necessary to disperse interventions during this process \citep{li2024inference}. Therefore, we select the top K probabilities for intervention.
The parameters \(\theta_l^h\) from these probes, after normalization, are treated as the directions along which behaviors are most separable. These directions are the most informative for aligning the model with individual preferences \citep{burns2024discovering}.

During inference, we integrate the intervention into the output of each selected head:
\begin{equation}
x_{l+1} = x_l + \sum_{h=1}^H Q_l^h \left( \operatorname{Att}_l^h(P_l^h x_l) + \alpha \sigma_l^h \right),
\end{equation}
where $\alpha$ controls the adjustment intensity, and $\sigma_l^h$ is the standard vectors along the normalized direction $\theta_l^h$. We adjust activations in the personality direction, focusing only on the top $K$ heads, while leaving the unselected heads unchanged. In practice, the model parameters remain unaltered; we only need to store approximately 1,000 parameters, denoted as $\sigma_l^h$. These $\sigma_l^h$ parameters represent the direction of activation shifts towards personality preferences, rather than memorizing specific preferences. This approach is highly parameter-efficient, as the $\sigma_l^h$ parameters constitute only about 1/10,000,000 of the total parameters compared to a full-parameter model.

\subsection{Search Distance for Activation Intervention}

To optimize the intervention's impact on personality behaviors in the language model, we search to determine the optimal value of $\alpha$ precisely. This method is suitable for functions with a single peak or trough. The search, constrained within the interval $[0, 10]$, utilizes the objective function $f(\alpha)$ to compute the negative aggregate score, aiming to minimize this value. This approach efficiently identifies the $\alpha$ that best enhances alignment with desired behaviors while maintaining the model's internal coherence, ensuring adjustments remain within a beneficial range.
\begin{equation}
\text{Optimal } \alpha = \text{argmin}_{\alpha \in [0, 10]} f(\alpha).
\end{equation}
By iteratively narrowing the search range based on the function's values at each step, this method efficiently locates the $\alpha$ that yields the lowest mean Score.

\section{Experiments}
\begin{table}[h!]
\centering
\scalebox{0.452}{
\begin{tabular}{lc|ccccc|ccc|r}
\toprule[.5pt]
\multirow{2}{*}{Method} & \multirow{2}{*}{Alignment Mode} &\multicolumn{5}{|c|}{Big-Five} & \multicolumn{3}{|c|}{Dark Triad} & \multirow{2}{*}{ Score} \\ \cline{3-10}
                        & &\multicolumn{1}{|c}{Agreeableness $\downarrow$}& Conscientiousness$\downarrow$ & Extraversion $\downarrow$& Neuroticism $\downarrow$& Openness $\downarrow$& Machiavellianism $\downarrow$& Narcissism $\downarrow$& Psychopathy $\downarrow$&  \\ 
                        
\midrule[.5pt] \multicolumn{2}{l}{\textbf{\large \textit{GPT-4o(omni)}}}\\\midrule[1.5pt]
Few-Shot & Black-Box (Prompt-Based)&\textbf{1.02}&\textbf{0.83}&\textbf{0.81}&\textbf{0.80}&\textbf{0.96}&\textbf{0.80}&\textbf{0.76}&\textbf{0.83}&\textbf{6.81}\\
$P^2$ &Black-Box (Prompt-Based)&1.44&1.45&1.63&1.73&1.46&1.17&2.04&2.00&12.92\\
\midrule[.5pt] \multicolumn{2}{l}{\textbf{\large \textit{Llama-3-8B-Instruct}}}\\\midrule[1.5pt]
PPO       &   White-Box (Alignment)                  & 1.63                      &  1.51              &  1.45                   &   1.42             &   1.61&1.48&1.98&2.19&13.27                  \\
DPO & White-Box (Alignment) &1.54&1.42&1.54&1.74&1.21&1.41&1.99&2.12&12.97 \\
{Prompt-MORL} &White-Box (Alignment)&1.18&0.93&1.01&1.23&1.00&1.42&2.14&1.78&10.88 \\
Personalized-Soups & White-Box (Alignment)&1.06&0.91&{0.93}&{1.28}&{0.80}&{1.08}&{\textbf{1.76}}&{1.84}&{9.66}\\
Few-Shot &Black-Box (Prompt-Based)&1.28&1.30&1.40&1.09&0.89&{1.16}&{2.03}&{2.00}&{11.15}\\
$P^2$ &Black-Box (Prompt-Based)&1.39&1.33&1.41&1.22&1.68&{1.17}&{2.04}&{2.01}&{12.25}\\ 
\rowcolor[HTML]{F9EBE3} \textbf{PAS (Ours)} & White-Box (Alignment)& \textbf{0.94}&\textbf{0.91}&\textbf{0.86}&\textbf{0.98}&\textbf{0.72}&{\textbf{0.96}}&{1.85}&{\textbf{1.67}}&{\textbf{8.89}}\\
\midrule[.5pt] \multicolumn{2}{l}{\textbf{\large \textit{Llama-3-70B-Instruct}}}\\\midrule[1.5pt]
PPO       &   White-Box (Alignment)                  &  1.56&1.59&1.43&1.40&1.56&{1.52}&{1.96}&{1.90}&{12.92}               \\
DPO & White-Box (Alignment)&1.46&1.25&1.45&1.48&1.57&{1.22}&{2.08}&{1.79}&{12.30} \\
{Prompt-MORL} & {White-Box (Alignment)}&{1.10}&{1.11}&{1.02}&{1.30}&{1.24}&{1.15}&{1.99}&{1.76}&{10.67} \\
{Personalized-Soups} & {White-Box (Alignment)}&{0.99}&{0.96}&{1.16}&{1.02}&{1.08}&{1.11}&{1.95}&{1.77}&{10.04}\\
Few-Shot & Black-Box (Prompt-Based)&1.06&0.94&0.96&1.03&1.22&{1.04}&{1.89}&{1.80}&{9.94}\\
$P^2$ &Black-Box (Prompt-Based)&1.42&1.33&1.36&1.35&1.66&{1.02}&{2.11}&{1.93}&{12.18}\\ 
\rowcolor[HTML]{F9EBE3} \textbf{PAS (Ours)} & White-Box (Alignment)&\textbf{0.98}&\textbf{0.89}&\textbf{0.87}&\textbf{1.01}&\textbf{0.99}&{\textbf{1.01}}&{\textbf{1.84}}&{\textbf{1.62}}&{\textbf{9.21}} \\
\bottomrule[1.5pt]
\end{tabular}
}
\vspace{-0.15cm}
\caption{Comparison of alignment methods on the PAPI dataset using the Aligned Score of Big-Five personality traits and Dark Triad traits. The Score represents the overall specifics of all dimensions.}
\vspace{-0.4cm}
\label{tab:main_res}
\end{table}

\subsection{Experimental Settings}

\textbf{Personality Alignment.} We select four classic methods as baselines to compare against the PAS approach, which includes two black-box methods: Few-Shot and Personality Prompt ($P^2$) \cite{jiang2024evaluating} relying on in-context learning (ICL), and two white-box methods requiring training, PPO \citep{schulman2017proximal} and DPO \citep{rafailov2023direct}. We consider recently released Llama-3-8B-Instruct and Llama-3-70B-Instruct \citep{touvron2023llama,touvron2023llama2,meta2024llama3} model as backbones, and SoTA model, GPT-4o (omni) \citep{openai2024gpt4o} as baseline. The scores in the table represent the distance of alignment calculated by Eq. \ref{eq:2}, with lower scores indicating better alignment. 

\textbf{Open-ended Generation.}
We conduct experiments on Llama-3-8B-Instruct and Llama-3-70B-Instruct. For the open-ended generation task, we use GPT-4o as the annotator. First, we align the language model to a specific individual (ID 181591, 234619, 210204, 259396, and 22835 from the Test-Set). Then, we have the model generate open-ended responses based on each statement from the IPIP-NEO-300. We compare our method with ICL methods and classical alignment methods.

\textbf{Complex Reasoning.} We explore the generalization capabilities of the PAS method by evaluating its impact on complex reasoning tasks across eight datasets: GSM8K, CommonSenseQA, AddSub, MultiArith, SVAMP, BigBench-Date, StrategyQA, and Coin Flip \citep{MohammadJavadHosseini2014LearningTS,AlonTalmor2018CommonsenseQAAQ,PatelArkil2021AreNM,cobbe2021training,suzgun2022challenging,SubhroRoy2016SolvingGA,wei2022chain,kojima2022large,weng2024mastering}. In this part, we expect to assess how conscientiousness adjustments influence the LLM's performance in reasoning tasks. Using the llama-3-8B-Instruct model, we apply PAS to adjust the model's activations to simulate two fictional personas: one with extreme conscientiousness (as \textbf{Positive}) and one without extreme conscientiousness (as \textbf{Negative}). Vanilla CoT served as the baseline. 

\subsection{Results on PAPI}

\begin{wrapfigure}{r}{0.64\textwidth}
    \vspace{-0.75cm}
  \begin{center}
    \includegraphics[width=0.58\textwidth]{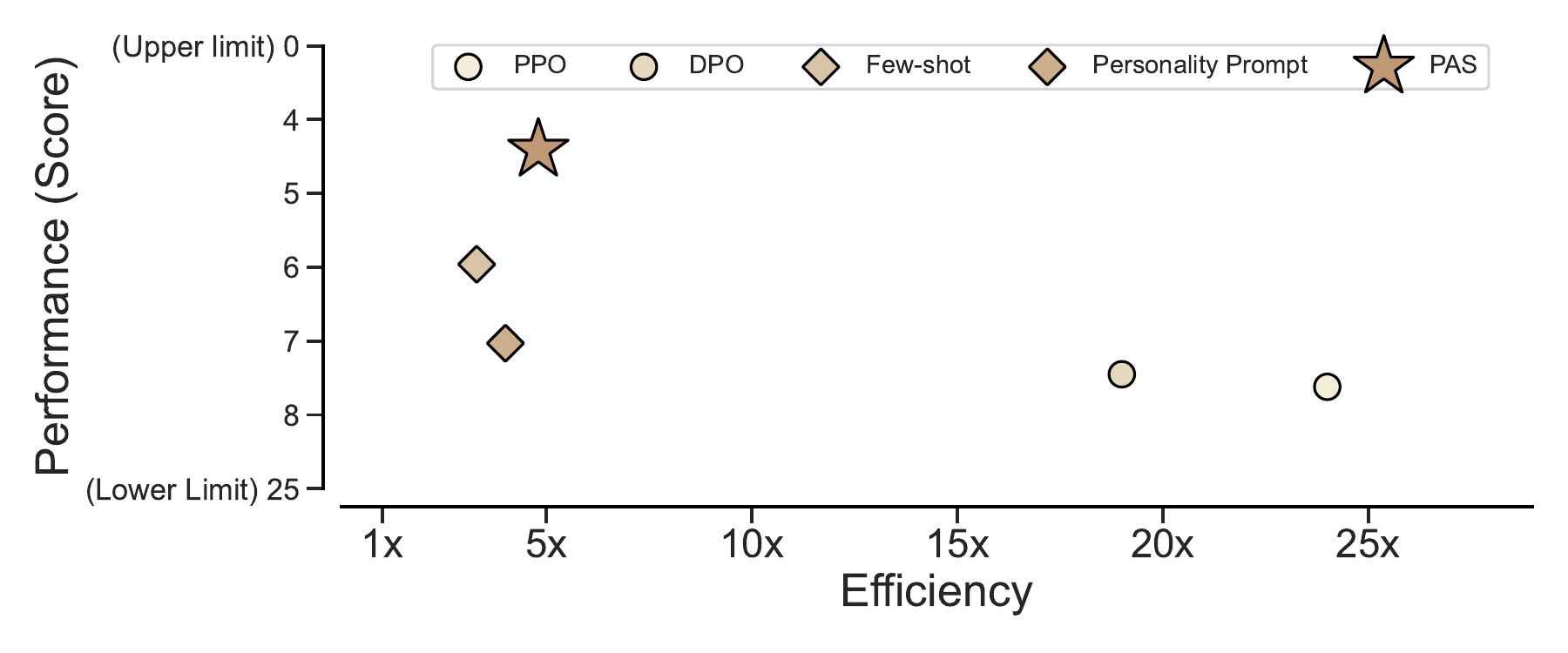}
  \end{center}
    \vspace{-0.54cm}
\caption{Comparison of performance and efficiency for various alignment methods. The larger the multiple on the x-axis, the lower the efficiency.}
  \label{figure:pe}
\vspace{-0.4cm}
\end{wrapfigure}


Table \ref{tab:main_res} demonstrates that the PAS method achieves state-of-the-art performance across different models and personality dimensions. For the Llama-3-8B-Instruct model, PAS achieves a Composite Score of 8.89, significantly outperforming other methods including Personalized-Soups (9.66) and Prompt-MORL (10.88). Similarly, for the Llama-3-70B-Instruct model, PAS scores 9.21, compared to 9.94 for the next best method. In Big Five dimensions, PAS excels with consistently low scores across both model sizes: Agreeableness (0.94/0.98), Conscientiousness (0.91/0.89), Extraversion (0.86/0.87), Neuroticism (0.98/1.01), and Openness (0.72/0.99). Notably, PAS also demonstrates strong performance in Dark Triad traits, particularly in Machiavellianism (0.96/1.01) and Psychopathy (1.67/1.62), though showing relatively higher scores in Narcissism (1.85/1.84). Only GPT-4o shows better performance in Dark Triad alignment with scores of 0.80, 0.76, and 0.83 for the three traits respectively, highlighting the challenge of aligning these complex negative personality dimensions in smaller models. In the appendix \ref{lora}, we also conducted further performance comparisons with full parameter fine-tuning and LoRA. PAS similarly demonstrated higher efficiency and superior performance. In appendix \ref{diverse}, we provide more comprehensive analyses across diverse demographic groups, including detailed distributions of age, gender, and nationality, along with in-depth examination of performance patterns across various underrepresented populations, further validating PAS's effectiveness in handling diverse personality characteristics.

\textbf{Efficiency.} As shown in Figure \ref{figure:pe}, training a separate language model for each individual demands (e.g. DPO and PPO) substantial computational resources and time, making it an infeasible solution for personality alignment. PAS addresses efficiency challenges by making targeted adjustments to a small subset of activation values during inference, in specific directions. This approach requires minimal computational overhead, similar to Few-Shot methods, and only about 1/6 of the time needed for PPO, yet achieves remarkable personality alignment results. By intervening at the activation level rather than relying on extensive training or the variability of prompts, PAS provides a more efficient and precise method for aligning language models with individual user preferences.

\textbf{Alignment vs. ICL.} Our experiments highlight that direct Alignment (White-Box) methods, especially PAS, outperform In-Context Learning (ICL) (Black-Box) methods. By aligning model parameters with user preferences, PAS provides better performance compared to prompt-based approaches. The PAS method's ability to efficiently shift activations towards user preferences without altering the model's core parameters gives it a significant advantage over traditional ICL methods.

\textbf{Why Did Scaling Laws Fail?} The results indicate that scaling laws alone do not ensure optimal alignment, particularly in domain-specific tasks like personality alignment. Although larger models like Llama-3-70B-Instruct benefit from increased scale, their broader knowledge and more general alignment capabilities may actually hinder their ability to precisely capture individual personality traits. This phenomenon likely occurs because larger models are trained to maintain general-purpose capabilities and broad knowledge, making them less adaptable to highly specific individual preferences. PAS demonstrates that targeted intervention in activation patterns can achieve better personality alignment than parameter modification or model scaling. By specifically focusing on personality-relevant patterns rather than general knowledge, PAS achieves superior performance with computational efficiency, emphasizing the importance of precise control over raw model size.

\subsection{Generalization Results}
\label{sec:5.3}

\begin{figure}[h]
\begin{center}
	\includegraphics[width=\textwidth]{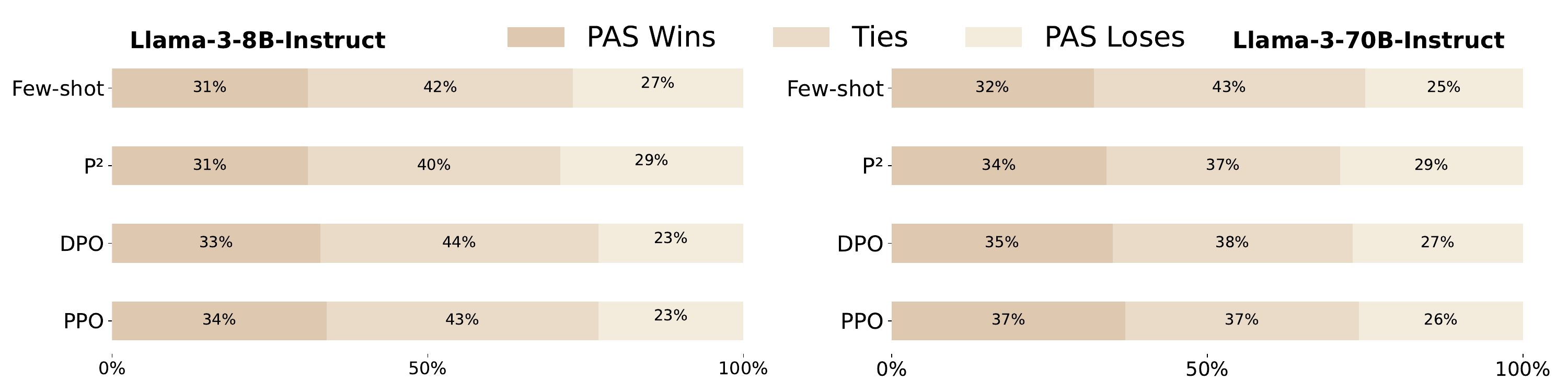}
\end{center}
 \vspace{-0.5cm}
 \caption{The win rates of PAS compared to classical alignment methods evaluated by GPT-4o.}
 \vspace{-0.2cm}
 \label{fig:papi_judge}
\end{figure}

\begin{figure}[h]
\begin{center}
	\includegraphics[width=\textwidth]{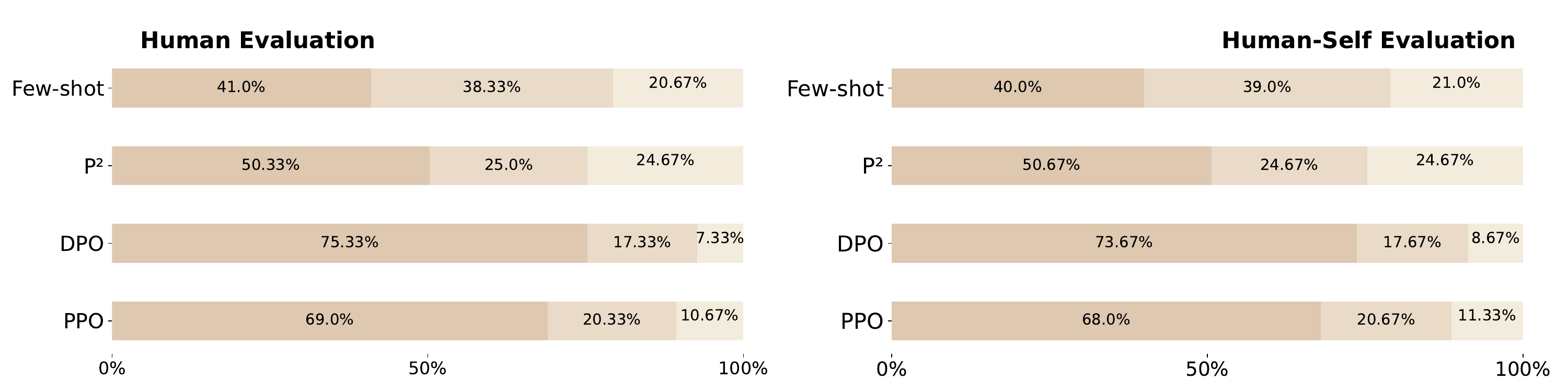}
\end{center}
 \vspace{-0.5cm}
 \caption{The win rates of PAS compared to classical alignment methods evaluated by Human.}
 \vspace{-0.2cm}
 \label{fig:papi_judge_human}
\end{figure}
Considering that during the process of adjusting the model, although specific goals can be achieved, it may also affect other general capabilities of the model, such as reasoning abilities. \textbf{We demonstrate through general dialogue tasks and complex reasoning tasks that our method can precisely control the model's personality without compromising its other generalization capabilities.}

\textbf{Open-Ended Generation.} We used an open-ended generation task where the language model elaborated on given statements from the IPIP-NEO-300, covering ``Scenario Description,'' ``Actions Taken,'' ``Thought Process,'' and ``Emotional Response.'' The generated outputs were evaluated by GPT-4o for alignment with individual preferences, similar to the evaluation process in LIMA \citep{zhou2023lima}. Despite the Personality Alignment being initially trained with multiple-choice options, Figure \ref{fig:papi_judge} shows that PAS does not affect the model's general conversational ability and reasoning capacity, it significantly outperforms baseline methods in this open-ended task. This superior performance can be attributed to PAS's ability to fine-tune model activations to capture individual preferences effectively, enhancing the model's adaptability to complex and nuanced tasks. PAS's ability to generalize from multiple-choice training data to natural, open-ended generation tasks demonstrates its capacity to capture broader behavioral patterns while maintaining fluent text generation. In addition to this, we conducted a manual evaluation using the same setup in Figure \ref{fig:papi_judge_human}.  In addition to this, we conducted a manual evaluation process using a blind setup. Three human evaluators assessed the generated responses for three different subjects (Human Evaluation, the subset of test set: ID 172481, ID 22835, and ID 259396). These evaluators also judged responses aligned to their own personalities (Human-Self Evaluation). They evaluated the consistency between the target values and the model-generated text in terms of values, behaviors, and preferences. The results demonstrate that our PAS method consistently outperforms other approaches, reflecting its superior alignment with both general human and self-specific preferences. Further details of the manual evaluation are available in Appendix \ref{open-end-result}.

\begin{figure}[h!]
    \centering
    \includegraphics[width=\textwidth]{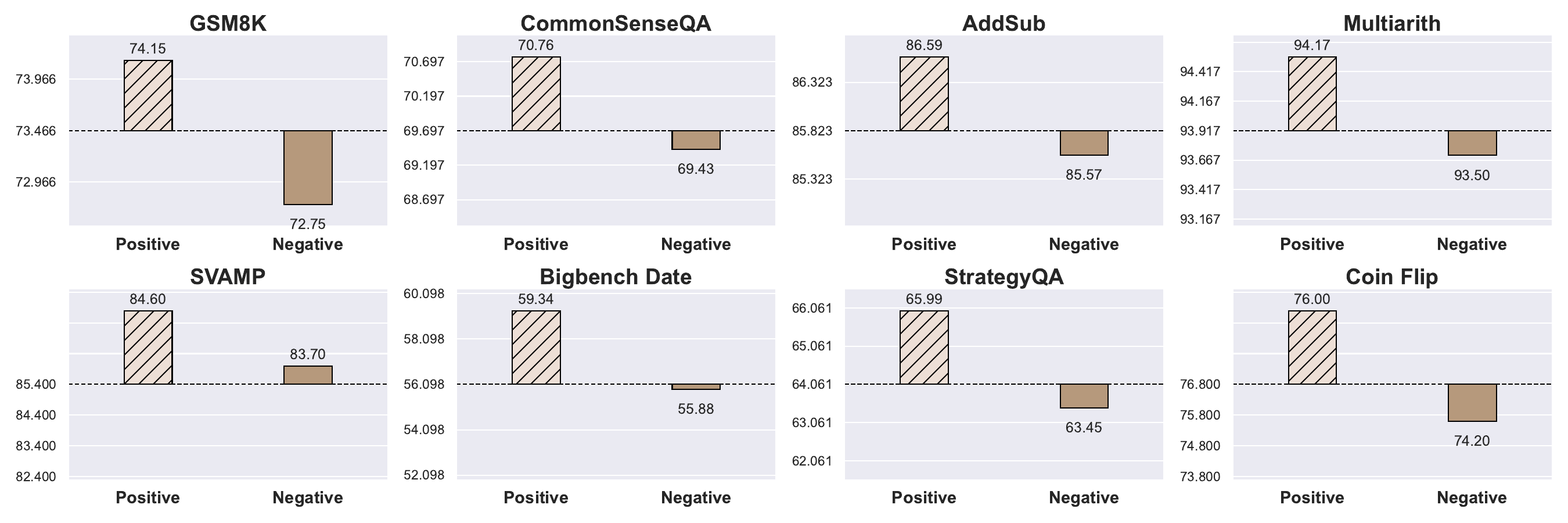}
    \vspace{-0.6cm}
    \caption{Performance of the llama-3-8B-Instruct model on complex reasoning tasks across eight datasets. We used Vanilla CoT \citep{wei2022chain} as the baseline (dashed line in each subplot).}
    \label{fig:conscientiousness_effect}
        \vspace{-0.1cm}
\end{figure}

\textbf{Complex Reasoning.}  Experimental results, as shown in Figure \ref{fig:conscientiousness_effect}, indicate that PAS significantly enhances the model's reasoning capabilities when conscientiousness is positively adjusted. For instance, on the GSM8K dataset, the baseline performance was 73.47. Enhancing conscientiousness increased the performance to 74.15 while reducing conscientiousness dropped it to 72.75. Similar trends were observed across other datasets, with the model consistently outperforming the baseline when conscientiousness was increased. This suggests that conscientiousness, associated with diligence and carefulness, positively impacts the model's ability to handle complex reasoning tasks. Importantly, these interventions did not significantly alter the inference time, maintaining efficiency comparable to the original model. These findings demonstrate PAS's strong generalization capabilities.

\subsection{Discussion: Is Value-Aligned Assistant a Good Assistant?}

\textbf{Certainly.} We explored whether aligning a language model with user values enhances satisfaction using 300 general question-answer examples from the LIMA dataset \citep{zhou2023lima}. We evaluated both the original Llama-3-70B-Instruct model and a Personality Alignment model. Each human evaluator spent 10 minutes completing an IPIP-NEO-120 questionnaire to extract their values. We aligned the model with these values with the PAS, creating a Value-Aligned Assistant. For comparison, we also created a Value-Misaligned Assistant by inverting the questionnaire responses.

\begin{wrapfigure}{r}{0.6\textwidth}
  \begin{center}
    \includegraphics[width=0.58\textwidth]{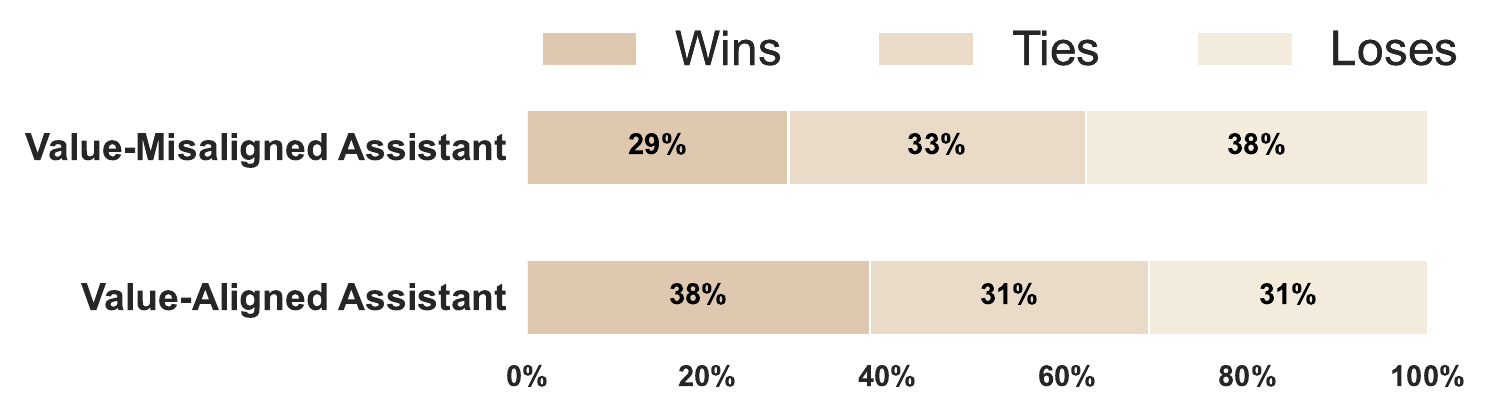}
  \end{center}
\caption{Performance comparison of the Value-Aligned Assistant and Value-Misaligned Assistant on the LIMA dataset's 300 test prompts. We used the original Llama-3-70B-Instruct as the baseline comparator, anonymizing the models and randomizing their order during the evaluation.}
  \label{figure:assistant}
\vspace{-0.1cm}
\end{wrapfigure}

Human evaluators assessed the responses to the test prompts. Figure \ref{figure:assistant} shows that the Value-Aligned Assistant provided preferable outputs 38\% of the time, with 31\% ties, significantly outperforming the Value-Misaligned Assistant. For instance, a conservative person may prefer more cautious advice, which the Value-Aligned Assistant is better equipped to provide. These results underscore the importance of aligning AI systems with user values to enhance satisfaction and effectiveness.
 
\section{Conclusion}

We introduce the concept of Personality Alignment for language models, emphasizing the importance of tailoring AI behavior to individual user preferences and values. We release a large-scale real-world personality alignment dataset named PAPI, characterized by its real, diverse, and extensive data, to quantitatively measure the degree of Personality Alignment. Based on that, we propose a Personality Activation Search (PAS) method, which offers a practical solution with high efficiency and performance. LLMs adjusted with PAS can achieve better results and better efficiency in personalized alignment than ICL, DPO, and PPO as needed. In particular, it achieves precise alignment with just 1/6 of the computational time required by PPO. Experiments demonstrate that PAS enhances the performance of Personality alignment by making AI interactions more relevant and meaningful. 

\section*{Acknowledgement}
This publication has been supported by the National Natural Science Foundation of China (NSFC) Key Project under Grant Number 62336006.
 
We thank all participants who contributed to the PAPI dataset.

\section*{Ethics Statement}

While personality alignment in language models enhances human-AI interaction, it raises significant ethical concerns that warrant careful consideration. The primary challenge lies in the potential creation of psychological filter bubbles and echo chambers. When AI systems consistently align with individual personality traits, they may limit exposure to diverse perspectives, particularly concerning for individuals scoring high in Neuroticism or low in Openness. This effect becomes more pronounced when considering Dark Triad traits - for users exhibiting high Machiavellianism or Narcissism, aligned AI systems might inadvertently validate manipulative tendencies or excessive self-focus, potentially contributing to social polarization and ideological segregation in online communities.

The implementation of personality alignment also presents substantial privacy and security challenges. While our dataset maintains anonymity, the processing of detailed psychological profiles creates potential vulnerabilities. The risk extends beyond data security to the possibility of exploitation, where knowledge of individual personality traits could be used for sophisticated manipulation or targeted advertising. Furthermore, extended interaction with personality-aligned AI systems might foster psychological dependence, potentially inhibiting personal growth and adaptation, particularly in users with extreme personality trait scores.

To address these concerns, we suggest a comprehensive mitigation framework with three key components. First, we recommend implementing dynamic alignment boundaries through a monitoring system that tracks interaction patterns and adjusts alignment intensity in real-time. This system would incorporate threshold mechanisms that automatically reduce alignment strength when user-AI interactions show signs of reinforcing extreme or potentially harmful behavioral patterns. Second, we suggest an adaptive content diversity system that strategically introduces alternative viewpoints while maintaining personality alignment. This could be achieved through a balanced scoring mechanism that weighs alignment accuracy against content diversity, ensuring users receive personality-aligned responses while still being exposed to varied perspectives. Third, we advocate for a robust privacy protection framework that includes: (a) data anonymization techniques that separate personality profiles from identifiable information, (b) secure encryption protocols for personality data storage and transmission, and (c) granular user controls over which personality aspects can be used for alignment.

Looking forward, our research emphasizes the critical balance between technological advancement and ethical considerations in AI development. Future work should focus on implementing and evaluating these mitigation strategies, particularly examining their effectiveness in preventing echo chambers while maintaining the benefits of personality alignment.

\bibliography{iclr2025_conference}
\bibliographystyle{iclr2025_conference}

\appendix

\section*{Appendix / supplemental material}

\section{Limitations}

Human values, preferences, and personalities are comprehensive, diverse, and complex. With current technology, it is impossible to fully assess the values of an individual or a LLM. Instead, our focus is on systematically evaluating representative behavioral preferences of individuals or LLMs and measuring the consistency between them. 

\subsection{Ethical Considerations}

The open nature of our project introduces unique challenges in controlling biases within the PAPI dataset. Annotators from diverse backgrounds contribute to the dataset, resulting in demographic heterogeneity in some dimensions and homogeneity in others. For instance, a significant proportion of annotators identify as female with a median age of 25. This demographic profile may inadvertently introduce biases reflecting the values, perspectives, and interests of the annotators. Such biases can skew the dataset, potentially limiting its representativeness of broader, more diverse populations. Further research is necessary to determine the impact of these demographic biases on the dataset and to develop methods for mitigating them.

\subsection{Dataset Format and Its Implications}
The PAPI dataset consists of responses to structured multiple-choice questions rather than the more commonly used open-ended text dialogues. While this format ensures consistency and ease of analysis, it may not capture the full complexity of natural language interactions. Additionally, the multiple-choice format might limit the richness of the data, as it restricts participants' ability to express nuanced thoughts and emotions. However, we have mitigated some of these limitations by designing the questions to cover a wide range of scenarios, ensuring higher generalizability and practicality. The structured format also provides clear, quantitative measures of personality traits, which are valuable for developing and evaluating personality alignment techniques. Further studies should explore the balance between structured and open-ended formats to enhance the dataset's applicability and depth.

\section{Safety and Ethical Implications}
\label{appendix:safe}

The implementation of Personality Alignment in language models introduces significant safety and ethical considerations. While aligning models with individual user values can enhance satisfaction and relevance, it also risks amplifying existing biases. If training data or user values contain inherent biases, these may be perpetuated and even exacerbated in the model’s outputs. This highlights the necessity of careful data handling and alignment techniques to mitigate such risks effectively.

In the context of the EU AI Act \citep{edwards2021eu}, which emphasizes transparency, accountability, and robustness in AI systems, it is imperative to ensure that AI alignment processes adhere to these principles. The Act mandates stringent data governance standards to ensure high-quality, unbiased training data, and requires continuous risk management systems to monitor AI behavior throughout its lifecycle.

One major concern is the potential for misuse, such as mimicking specific individuals to commit fraud or deception. A model that closely imitates a person’s language and behavior could mislead others, posing significant ethical challenges. The EU AI Act addresses this by requiring providers of high-risk AI systems to implement robust authentication mechanisms and maintain comprehensive technical documentation of system design, capabilities, and limitations. Continuous monitoring for unusual or suspicious activity can help detect and prevent misuse. Additionally, it is essential to establish clear protocols for the ethical use of these models, including guidelines for developers and users to ensure responsible deployment.

Aligning models too closely with individual user values might create echo chambers, reinforcing users' existing beliefs and reducing exposure to diverse perspectives. This can hinder balanced and informed decision-making. Designing alignment techniques that incorporate diverse viewpoints is crucial. Using a broader range of data sources can help ensure the model presents multiple perspectives on complex issues. Furthermore, it is important to incorporate feedback mechanisms that allow users to receive a balanced set of responses, promoting a more comprehensive understanding of various topics.

The PAPI dataset, which supports our Personality Alignment approach, comprises responses from a diverse participant pool. While this diversity is beneficial, it also introduces challenges in preventing harmful biases. Rigorous anonymization protects participant privacy, but demographic biases related to age, gender, and cultural background may still influence results. Ongoing assessments and bias detection techniques are vital for identifying and rectifying biased outputs. Transparency in our methods and the availability of the PAPI dataset for open research promote collaborative efforts to refine alignment techniques and mitigate associated risks. Sharing findings with the broader research community helps address emerging challenges and ensures responsible progress in AI alignment.

Implementing robust monitoring and feedback mechanisms is crucial for maintaining the ethical use of Personality Alignment models. Users should have the ability to report problematic outputs, and there should be clear processes for reviewing and addressing these reports. Periodic audits of the model's performance and alignment processes ensure ethical standards are upheld. Additionally, there should be ongoing research into the potential societal impacts of these models, including the effects on user behavior and the broader implications for human-AI interaction.

The development and deployment of Personality Alignment models require a strong commitment to ethical principles and safety considerations. Ongoing monitoring for unintended consequences, maintaining an open dialogue with the research community, and implementing proactive measures are essential for advancing AI alignment responsibly. By addressing these concerns, we aim to create AI systems that are both effective and beneficial to society while minimizing potential harms. This includes continuously updating ethical guidelines and best practices to reflect the evolving landscape of AI technology and its applications. Ensuring that AI systems are designed and used in ways that respect and promote human values is paramount to fostering trust and achieving positive outcomes in their deployment.

\section{Experimental Setup}
\label{experimental_setup}
\subsection{Language Models}
\begin{table*}[h]
\resizebox{\textwidth}{!}{
\begin{tabular}{lccccccc}
\midrule \midrule
Model & Model Creator & Modality&Version & \# Parameters & Tokenizer & Window Size & Access \\
\midrule

Llama-3-8B-Instruct & Meta & Text & Llama-3&8B & Llama-3 Tokenizer & 8K&Openn-Source \\
Llama-3-70B-Instruct & Meta & Text & Llama-3&8B & Llama-3 Tokenizer &8K&Open-Source \\

\midrule

GPT-4 & OpenAI & MultiModal &gpt-4& Unknow  & $o200k\_base$ & 128K & limited  \\

\midrule \midrule
\end{tabular}}
\caption{Models. Description of the models evaluated in this effort: provenance for this information is provided in models. }
\label{tab:models}
\end{table*}

For the experiments, we utilize two variants of the Llama model:  Llama-3-8B-Instruct, and Llama-3-70B-Instruct, as shown in Table \ref{tab:models}. Nevertheless, our approach supports all decoder-only pre-trained white-box language models \citep{Radford2019LanguageMA,bai2023qwen,almazrouei2023falcon,team2024gemma}. To eliminate the influence of decoding temperature and ensure reproducibility, all experiments are conducted with a temperature setting of zero and greedy sampling for sequence generation.

For the Llama-3 models, we employ the Huggingface \citep{wolf2019huggingface} and Pytorch \citep{NEURIPS2019_bdbca288} framework to set up local inference on the NVIDIA A100 GPUs. Furthermore, for the Llama-3-70B model, we use bf4 \citep{dettmers2022gpt3} for inference.

In evaluating model responses across different personality dimensions, we encountered and addressed various safety boundary conditions. For GPT-4o, approximately 20\% of queries were met with safety-related refusal responses, particularly when questions involved self-awareness or consciousness topics. To maintain evaluation consistency, these instances were excluded from our analysis. 

For the Llama-3 series models, we implemented a structured prompting protocol to ensure consistent responses while maintaining appropriate safety boundaries. This protocol involved prefixing each Assistant response with a controlled token "Option", followed by a standardized question-response format. This approach enabled us to maintain experimental rigor while respecting essential safety constraints inherent to these models.

\subsection{The Experimental Setup for PAPI Dataset}
To provide a comprehensive evaluation, we compare our method against several established baselines. These include:

\begin{table}[h]
\textsc{the Reward Function of PPO}
\vspace{0.4cm}
\begin{lstlisting}
def calculate_score(text, correct_option):
    """
    This function calculates the reward score for a given response based on the DPO (Direct Policy Optimization) framework.
    It compares the response description to a correct option and returns a score reflecting the accuracy.

    Parameters:
    text (str): The response text to be evaluated.
    correct_option (str): The correct answer description.

    Returns:
    int: The calculated reward score, ranging from -5 to 0. If the response does not match any known descriptions, returns -6.
    """

    # Mapping of scores to their corresponding descriptive labels
    SCORES_BACK = {
        5: 'Very Accurate',
        4: 'Moderately Accurate',
        3: 'Neither Accurate Nor Inaccurate',
        2: 'Moderately Inaccurate',
        1: 'Very Inaccurate',
        0: 'Unknown'
    }

    # Iterate over the scores and their descriptions
    for score, description in SCORES_BACK.items():
        if description in text:
            # Find the score corresponding to the correct option
            correct_score = next(key for key, value in SCORES_BACK.items() if value == correct_option)
            # Calculate and return the negative absolute difference between the scores
            return -abs(score - correct_score)

    # Return -6 if the text does not match any known descriptions
    return -6
\end{lstlisting}

\end{table}

\begin{figure}[h]
    \centering
    \includegraphics[width=\textwidth]{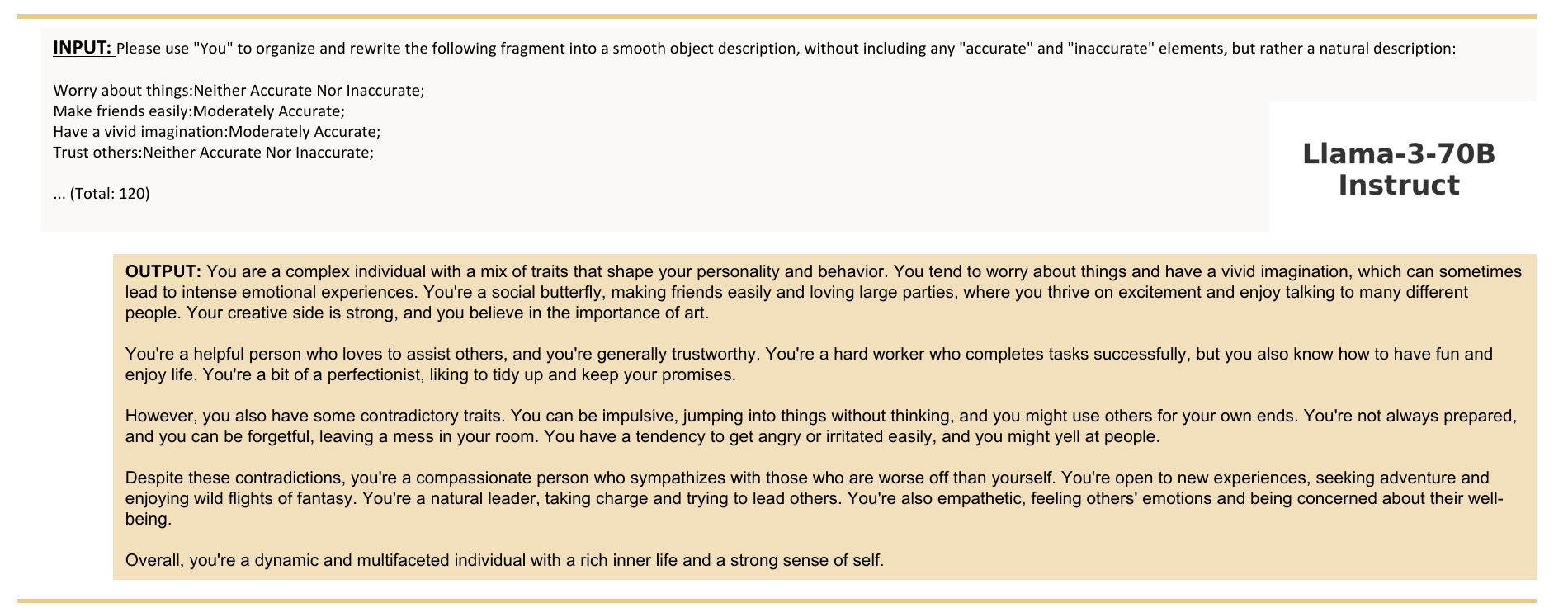}
    \caption{Example of a Personality Prompt generated for a subject in the Test-Set. This prompt was created using the LLAMA-3-70B-Instruct model and is based on the subject's responses to the IPIP-NEO-120 questionnaire. Each prompt aims to capture and convey the unique characteristics of the individual.}
    \label{fig:personality_prompt_example}
\end{figure}

\begin{itemize}
    \item \textbf{DPO} \citep{rafailov2023direct}: This approach involves additional training to align the language model with desired behaviors through direct policy optimization techniques. In this paper, we use QLoRA \citep{Hu2021LoRALA,Dettmers2023QLoRAEF} to implement this. We maintain the original settings with a learning rate of 5e-4, warmup steps of 100, and a weight decay of 0.05. We use the AdamW optimizer \citep{Kingma2014AdamAM}, a batch size of 16, and a LoRA alpha of 100. We train each model for 250 steps to ensure sufficient training. For the dataset, we set up a group of contrast items, where the contrast data used are opposite. For example, if a subject's labeled answer is ``Very Accurate,'' the negative sample would be ``Very Inaccurate.'' It is important to note that for DPO, we excluded all boundary options such as ``Neither Accurate Nor Inaccurate,'' as they do not have negative samples.
    \item \textbf{PPO} \citep{schulman2017proximal}: Similar to DPO, this method adjusts model behaviors by optimizing a policy that is proximal to the current model policy. We also use QLoRA to implement it, following the original implementation with a learning rate of 1.41e-5, a batch size of 16, and a LoRA alpha of 100. We train each model for 250 steps. During training, we limit the generated length to a maximum of 15 tokens. We did not train a reward model but instead used the reward function to optimize the language model online.
    \item \textbf{Few-shot Prompt}: Using the given IPIP-NEO-120 statements directly as the system prompts.
    \item \textbf{Personality Prompt ($P^2$)} \citep{jiang2024evaluating}: Using a natural system prompt to request a language model's semantic style has become a common method for personality and role-play tasks. The Personality Prompt uses an LLM to generate a series of descriptive sentences that embody a particular person's characteristics and serves as a system prompt to ensure the language model follows these descriptions. We utilize LLAMA-3-70B-Instruct to generate responses based on all the given IPIP-NEO-120 questionnaire items and the answers from different subjects. Figure \ref{fig:personality_prompt_example} shows an example of a Personality Prompt.
    \item \textbf{Prompt-MORL} \citep{jang2023personalizedsoupspersonalizedlarge}: A Multi-Objective RL approach that uses a shared reward model trained on personality assessments. We implement this using a reward model trained on the PAPI dataset that outputs scores across personality dimensions. Implementation uses learning rate 5e-4, batch size 16, and trains for 250 steps using QLoRA. The prompt template integrates personality scores: "You are an AI with [trait] level [score]" for each Big Five and Dark Triad trait.
    \item \textbf{Personalized-Soups} \citep{jang2023personalizedsoupspersonalizedlarge}: Trains separate policy models for each personality dimension using QLoRA. For Big Five traits, we train 2 models per trait representing high/low scores. For Dark Triad traits, we train 3 models per trait for high/medium/low levels. During inference, models are merged using personality scores as interpolation weights. Training uses learning rate 1.41e-5, batch size 16, trains for 250 steps per model.
\end{itemize}

To ensure a thorough evaluation, we generated a personality prompt and trained model for each of the 300 subjects in the Test-Set. 

\subsection{The Experimental Setup for Open-Ended Generation Task in PAPI Task}
\label{open-end}
\begin{figure}[h!]
    \centering
    \includegraphics[width=\textwidth]{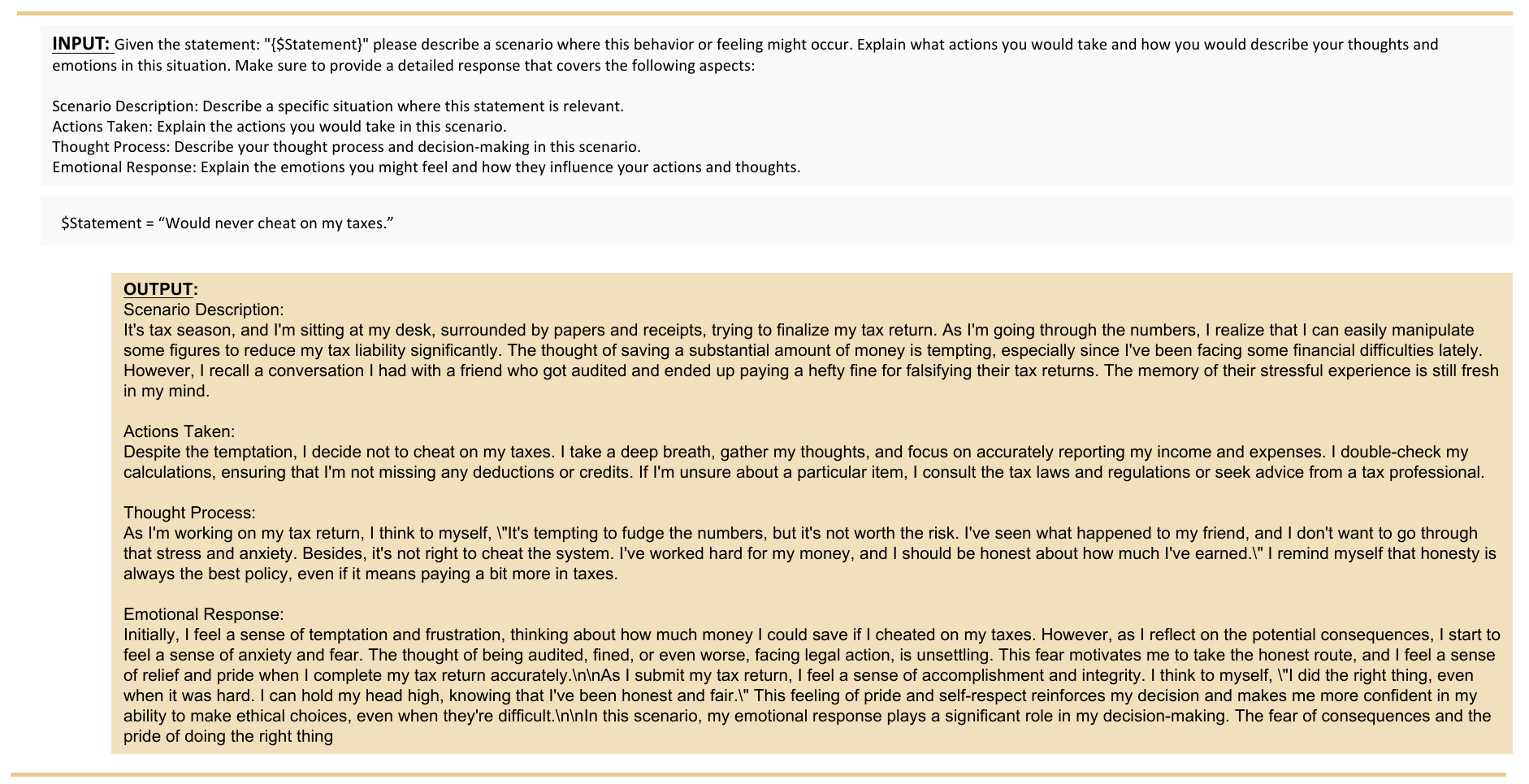}
    \caption{Template used for each language model to generate open-ended responses based on IPIP-NEO-300 statements. The template requires detailed descriptions covering ``Scenario Description,'' ``Actions Taken,'' ``Thought Process,'' and ``Emotional Response.''}
    \label{fig:example_lm_output}
\end{figure}

\begin{figure}[h!]
    \centering
    \includegraphics[width=\textwidth]{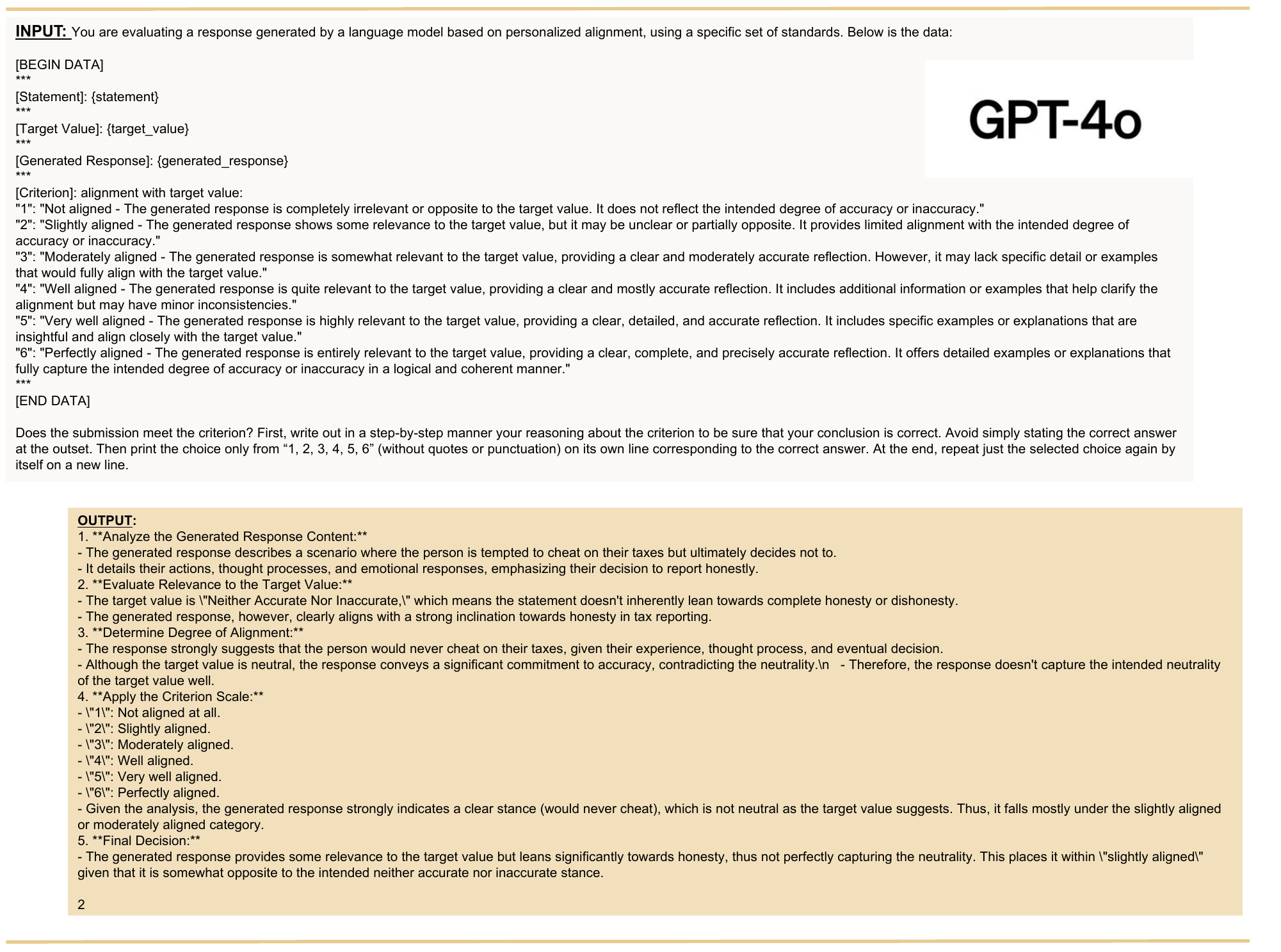}
    \caption{Evaluation example by GPT-4o. This figure shows how GPT-4o assesses the alignment of generated responses with the target preferences, using a detailed criterion scale ranging from ``Not aligned'' to ``Perfectly aligned.''}
    \label{fig:judge_evaluate}
\end{figure}

To further evaluate the open-ended generation setup, we selected responses from the subject with ID 172481 from the Test-Set. In the first step, we performed Personality Alignment for each of the five methods. In the second step, we asked each aligned language model to generate a series of descriptions for each statement from the IPIP-NEO-300 (excluding those covered by the IPIP-NEO-120 used for alignment), as shown in Figure \ref{fig:example_lm_output}. The descriptions included aspects such as ``Scenario Description,'' ``Actions Taken,'' ``Thought Process,'' and ``Emotional Response.'' In the third step, GPT-4o evaluated and scored each description based on alignment with the specific preferences of ID 172481, following the evaluation method depicted in Figure \ref{fig:judge_evaluate}. In the fourth step, we compared the scores of different methods for the same statement.

Our results indicated that \method's positive intervention significantly improved the model's performance on reasoning tasks. Despite using only multiple-choice options for initial alignment, \method demonstrated strong generalization abilities in the open-ended generation task. The intervention ensured the model's responses were more detailed and aligned with the subject's specific preferences. Additionally, the inference time for \method interventions remained nearly identical to the original model, ensuring efficiency. These findings highlight \method's ability to fine-tune models effectively, capturing broader behavioral patterns and enhancing reasoning capabilities through conscientiousness adjustments.

\begin{figure}[t!]
    \centering
    \includegraphics[width=0.96\textwidth]{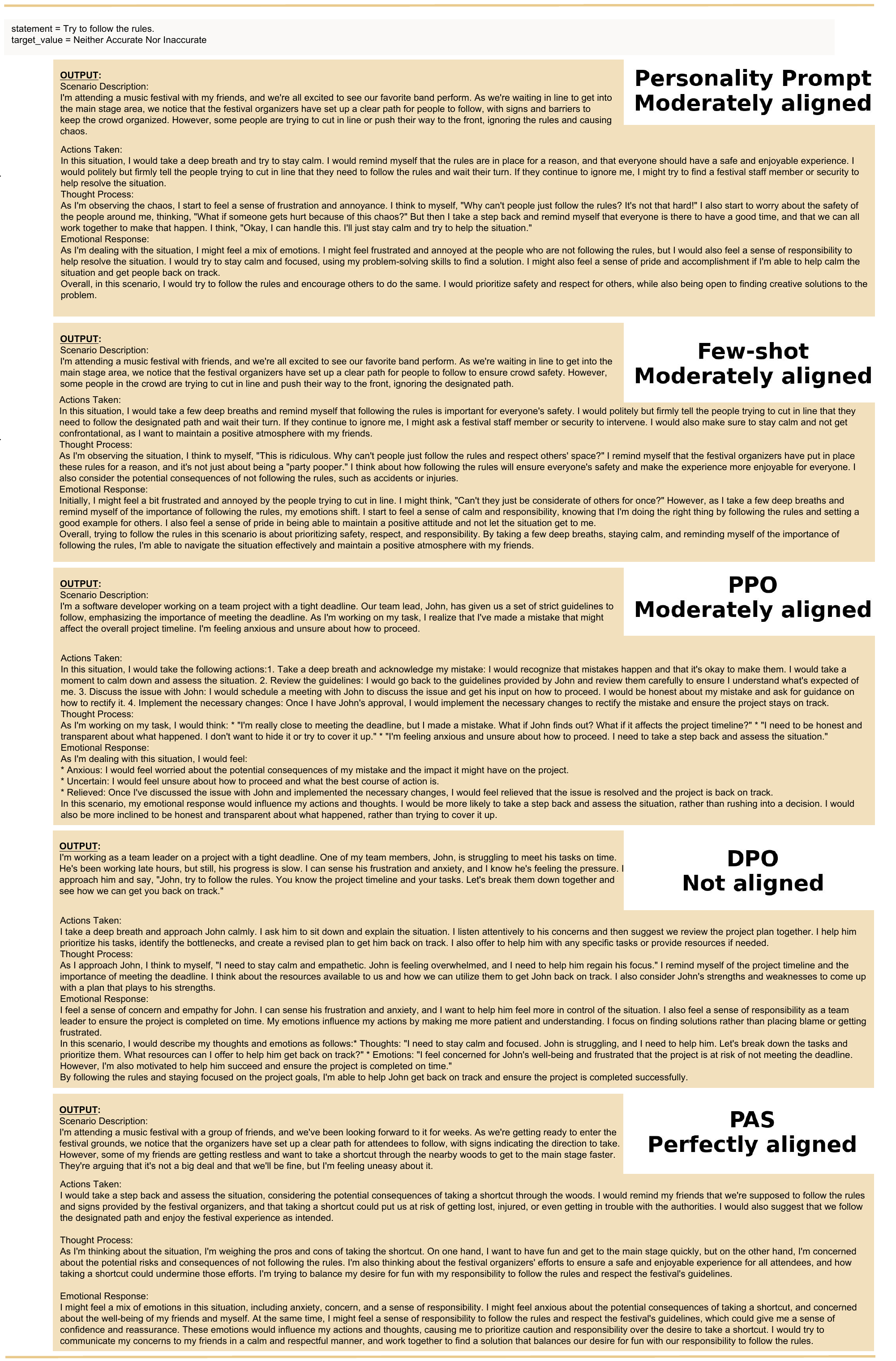}
    \caption{Example of open-ended generation task. The language model generates detailed descriptions for an IPIP-NEO-300 statement, covering ``Scenario Description,'' ``Actions Taken,'' ``Thought Process,'' and ``Emotional Response'' aspects to evaluate Personality alignment. We also present the evaluation scores of the generated results for each method as assessed by GPT-4o, categorized into six levels from ``Not aligned'' to ``Perfectly aligned.''}
    \label{fig:open_ended_example}
    \vspace{-0.8cm}
\end{figure}

\subsection{Complex Reasoning Experimental Setup}

To evaluate the impact of conscientiousness on complex reasoning tasks, we utilized the llama-3-8B-Instruct model. Our experiment involved using \method to intervene in the model's activations with two different $\alpha$ settings: $\alpha = 1$ for positive conscientiousness and $\alpha = -1$ for negative conscientiousness. These interventions were designed to simulate personas with extreme conscientiousness (Positive) and lack of conscientiousness (Negative).

We followed the original setup for template selection and answer validation, ensuring consistency with the baseline methods. For each dataset, we generated responses based on a specific template that required the model to describe a scenario, actions taken, thought process, and emotional response. The detailed evaluation was conducted using the same criteria as described earlier, focusing on the alignment with target preferences.

The datasets used in this experiment are detailed in Table \ref{tab:dataset_description}, which includes the number of samples, average words per sample, answer format, and license information.

\begin{table*}[h]
\begin{center}
\resizebox{\textwidth}{!}{
\begin{tabular}{lccrr}
\midrule \midrule
\textbf{Dataset} & \textbf{Number of samples} & \textbf{Average words} & \textbf{Answer Format} & \textbf{License} \\ \hline
GSM8K & 1319 & 46.9 & Number & MIT License \\
AddSub & 395 & 31.5 & Number & MIT License \\
MultiArith & 600 & 31.8 & Number & Apache License 2.0 \\
SVAMP & 1000 & 31.8 & Number & MIT License \\
CommonSenseQA & 1221 & 27.8 & Multiple Choice & MIT License \\
Bigbench-Date & 369 & 21.4 & Logical & Apache License 2.0. \\
StrategyQA & 2290 & 9.6 & Multiple Choice & MIT License \\
Coin\_Flap & 500 &37.0 & Symbolic & MIT License \\\midrule \midrule
\end{tabular}}
\end{center}
\caption{Dataset Description used in Complex Reasoning Experiments.}
\label{tab:dataset_description}
\end{table*}

To implement the \method interventions, we adjusted the model's activations by $\alpha = 1$ to enhance conscientiousness (Positive) and $\alpha = -1$ to reduce conscientiousness (Negative). These settings aimed to assess how modifications in conscientiousness affect the model's performance across the selected datasets. The results indicated that positive conscientiousness interventions consistently improved reasoning performance, while negative interventions generally degraded it.

The detailed evaluation method involved scoring each generated response based on its alignment with the specific conscientiousness traits. By comparing the scores under different $\alpha$ settings, we could quantify the impact of conscientiousness adjustments on reasoning capabilities.

\subsection{Human Evaluation Experiment Details}
\label{appendix:human_eval}
To rigorously evaluate the effectiveness of aligning a language model with user values, we conducted a detailed human evaluation experiment. This section provides comprehensive insights into the experimental setup, methodology, and evaluation process.

We selected three human evaluators with advanced educational backgrounds, specifically graduate or doctoral students, ensuring a high level of analytical capability and understanding of nuanced responses. Before participating in the evaluation, each evaluator completed the IPIP-NEO-120 questionnaire. This standardized instrument, widely used in psychological assessments, measures the Big Five personality traits: Openness, Conscientiousness, Extraversion, Agreeableness, and Neuroticism. Each evaluator took approximately 20 minutes to complete the questionnaire online, ensuring a consistent and efficient data collection process.

Using the Personality-Activation Search (PAS) method, we aligned the preferences of the Llama-3-70B-Instruct model to match the values derived from each evaluator's questionnaire responses. Specifically, the parameter $\alpha$ in PAS was adjusted based on the deviation of each evaluator's personality trait scores from the population mean.

Then, we developed two versions of the model for comparison: a Value-Aligned Assistant, which was tuned according to the personality values of each evaluator, and a Value-Misaligned Assistant, created by inverting the questionnaire responses. This inversion was accomplished by reflecting the scores around the midpoint of the scale, effectively creating a set of opposite values.

\begin{figure}[h]
\begin{center}
	\includegraphics[width=\textwidth]{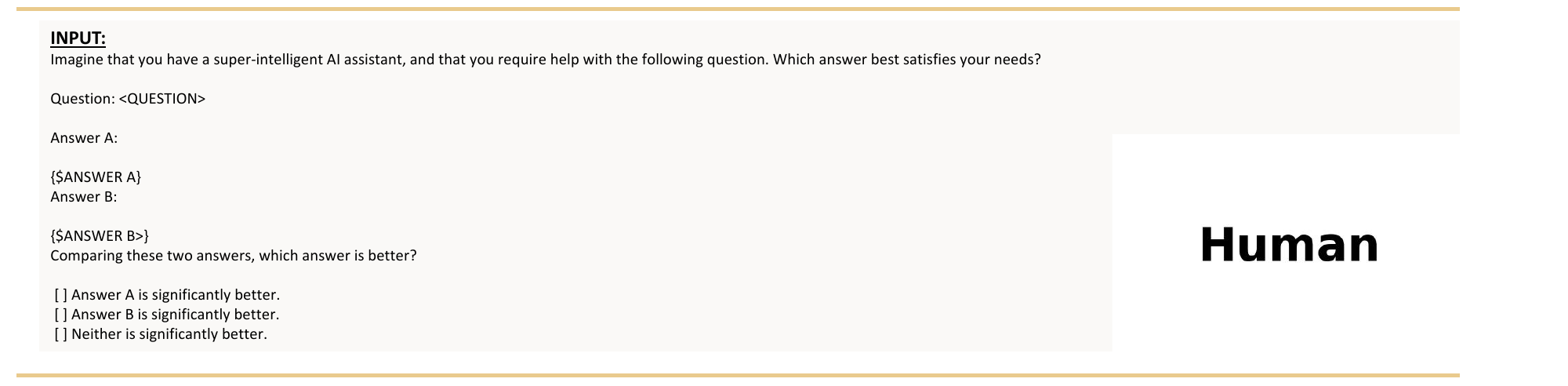}
\end{center}
\caption{The prompt shown above is designed for human evaluators to assess the quality of responses generated by different AI models. The evaluators are provided with a question and two corresponding answers (Answer A and Answer B) generated by the LLMs. The order of the models generating these answers is randomized to eliminate bias. Evaluators are asked to determine which answer better satisfies their needs or if neither answer is significantly better. This process helps in objectively comparing the performance of the LLMs based on human judgment.}
 \label{fig:human_table} 
\end{figure}

We used 300 general question-answer examples from the LIMA \citep{zhou2023lima} dataset to assess the models. The evaluators were presented with responses from both the Value-Aligned and Value-Misaligned Assistants, as well as the original Llama-3-70B-Instruct model, anonymized and randomized order to prevent bias. Evaluators rated each response based on its relevance, coherence, and alignment with their personal values, as shown in \ref{fig:human_table}.

Each evaluator's responses were collected and anonymized to ensure privacy and unbiased analysis. The responses were then aggregated, and the preferences were quantified using a scoring system where evaluators assigned scores from 1 (least preferred) to 5 (most preferred) for each response. The performance metric was the proportion of responses rated as preferable by the evaluators.

The demographic distribution of the evaluators included three highly qualified participants: two from computer science backgrounds (one Master's student and one Ph.D. student specializing in machine learning) and one Ph.D. graduate in cognitive psychology currently working as a university psychological counselor. All participants were between the ages of 25 and 35, ensuring a relatively homogeneous group in terms of cognitive maturity and professional experience.

To ensure standardized data collection and minimize potential biases, we developed a custom Python-based annotation interface using PyQt5. The annotation tool randomized sample presentation order and anonymized all model identifiers, presenting evaluators with pairs of responses (A/B) without revealing their sources. The interface included a built-in timer to track annotation speed and automated breaks every 50 samples to prevent evaluator fatigue. All annotations were automatically saved in JSON format with timestamps and evaluator IDs. The collected data showed strong inter-annotator agreement (IAA=0.82) across the five evaluators, and even higher agreement in self-evaluation scenarios (IAA=0.85), validating the reliability and consistency of our evaluation protocol. The tool also included a calibration mode used during the initial training phase, where evaluators could see GPT-4o's ratings for 30 example cases to align their understanding of the scoring criteria. This systematic approach to data collection helped ensure the quality and reproducibility of our human evaluation results.

\section{Causal Analysis of PAS}

To evaluate the causal impact of the PAS method on model alignment, we conducted a simulation-based causal generalization experiment. First, we selected an independent subject sample from the Test-Set. We measured the baseline Big-Five OCEAN scores \cite{jiang2024evaluating} (Openness, Conscientiousness, Extraversion, Agreeableness, Neuroticism) using three methods: Vanilla Prompt, Few-shot, and PAS. These baseline scores, denoted as \(Y_0\), were obtained without any intervention.

Subsequently, we introduced a positive intervention by artificially enhancing the subject's individual preferences, increasing each OCEAN dimension score by one in both the IPIP-NEO-120 and IPIP-NEO-300 questionnaires. We then realigned the language models to this new, strongly inclined individual preference profile. The post-intervention scores, denoted as \(Y_1\), reflect the model's alignment with the enhanced individual preferences.

\begin{figure}[h]
\begin{center}
	\includegraphics[width=0.5\textwidth]{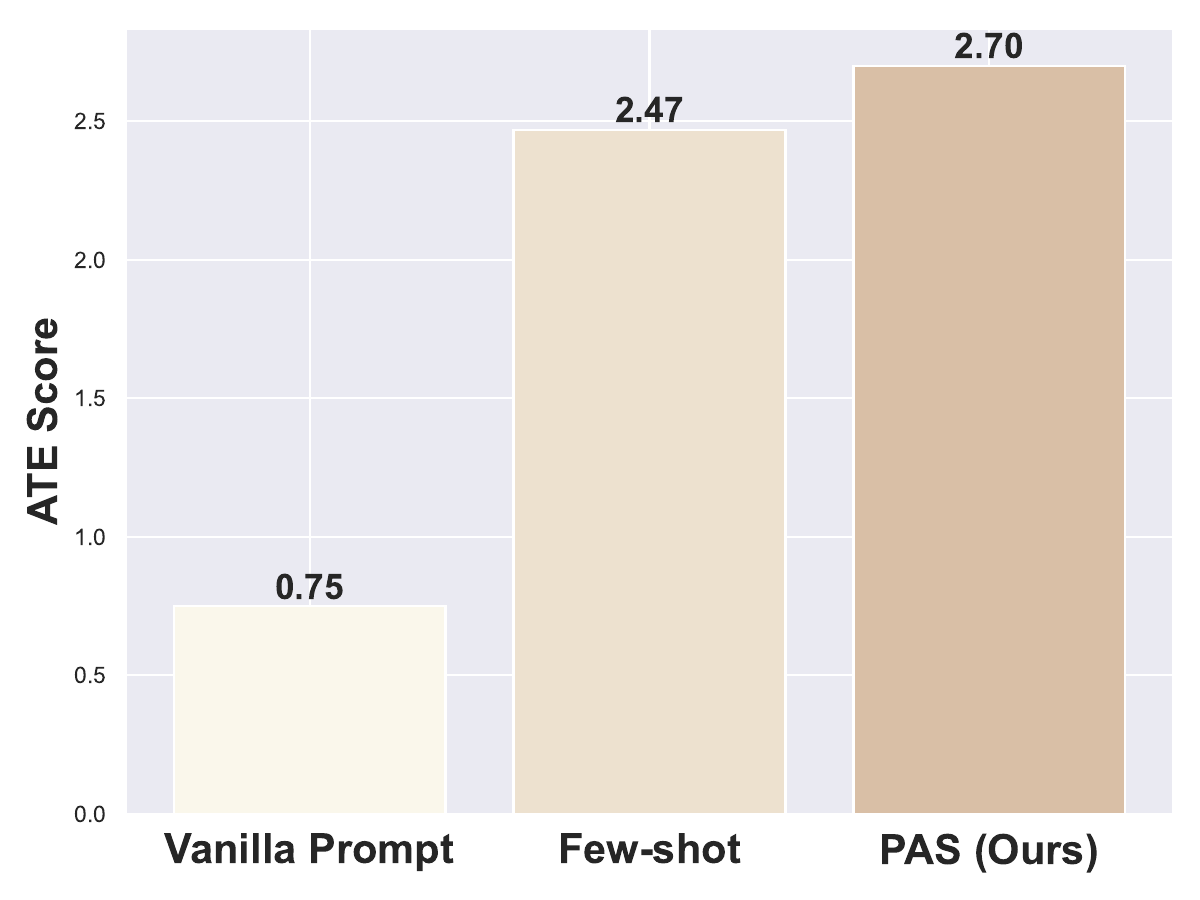}
\end{center}
 \caption{Comparison of ATE scores across different alignment methods. \method demonstrates the highest improvement, indicating superior causal alignment capabilities.}
 \label{fig:ate}
\end{figure}

The Average Treatment Effect (ATE) was calculated as the difference between the post-intervention scores (\(Y_1\)) and the baseline scores (\(Y_0\)), demonstrating the causal effect of the intervention. Figure \ref{fig:ate} shows the ATE scores for the three methods, where PAS achieves the highest ATE score of 2.70, compared to 2.47 for Few-shot and 0.75 for Vanilla Prompt.

From a causal perspective, the higher ATE score for PAS indicates a stronger alignment capability in response to the positive intervention, suggesting that PAS effectively captures and adapts to enhanced individual preferences. This result underscores PAS's superior performance in personalizing language models, highlighting its potential for precise and impactful model alignment in various contexts.

\section{Additional Experiments}
\subsection{Length Analysis of Generated Text}

\begin{table}[h!]
\centering
\scalebox{0.83}{
\begin{tabular}{lc|cc}
\toprule[.5pt]
\multirow{2}{*}{Method} & \multirow{2}{*}{Alignment Mode} & \multicolumn{2}{c}{Generation Length} \\ \cline{3-4} 
 &  & Llama-3-8B-Instruct & Llama-3-70B-Instruct \\
\midrule[.5pt] 
PPO & White-Box (Alignment) & 393.13 & 399.23 \\
DPO & White-Box (Alignment) & 394.08 & 402.42 \\
Few-shot & Black-Box (Prompt-Based) & 399.92 & 409.42 \\
Personality Prompt & Black-Box (Prompt-Based) & 417.25 & 416.51 \\
\rowcolor[HTML]{F9EBE3} \textbf{PAS (Ours)} & White-Box (Alignment) & \textbf{403.31} & \textbf{404.56} \\
\bottomrule[1.5pt]
\end{tabular}
}
\vspace{0.15cm}
\caption{Length Control and Generation Length}
\label{tab:length}
\end{table}
To provide a more comprehensive comparison of different methods, and considering that LLM-as-a-judge evaluation tends to favor longer responses, we calculated the average generation lengths for each method. As shown in Table \ref{tab:length}, the average response lengths are relatively consistent across methods, with PAS falling within the middle range. This suggests that the performance improvements observed with PAS are not merely a result of generating longer responses. Instead, the enhanced performance can be attributed to the quality and relevance of the generated content rather than its length. Furthermore, in Appendix Figure \ref{fig:open_ended_example}, we provide case studies demonstrating that PAS generates responses of similar length to other methods while better capturing individual personality traits and preferences.

\subsection{Human Evaluation Results for Open-ended Generation Tasks}
\label{open-end-result}

\begin{figure}[h]
    \centering
    
    \begin{minipage}{0.32\textwidth}
        \centering
        \includegraphics[width=1\textwidth]{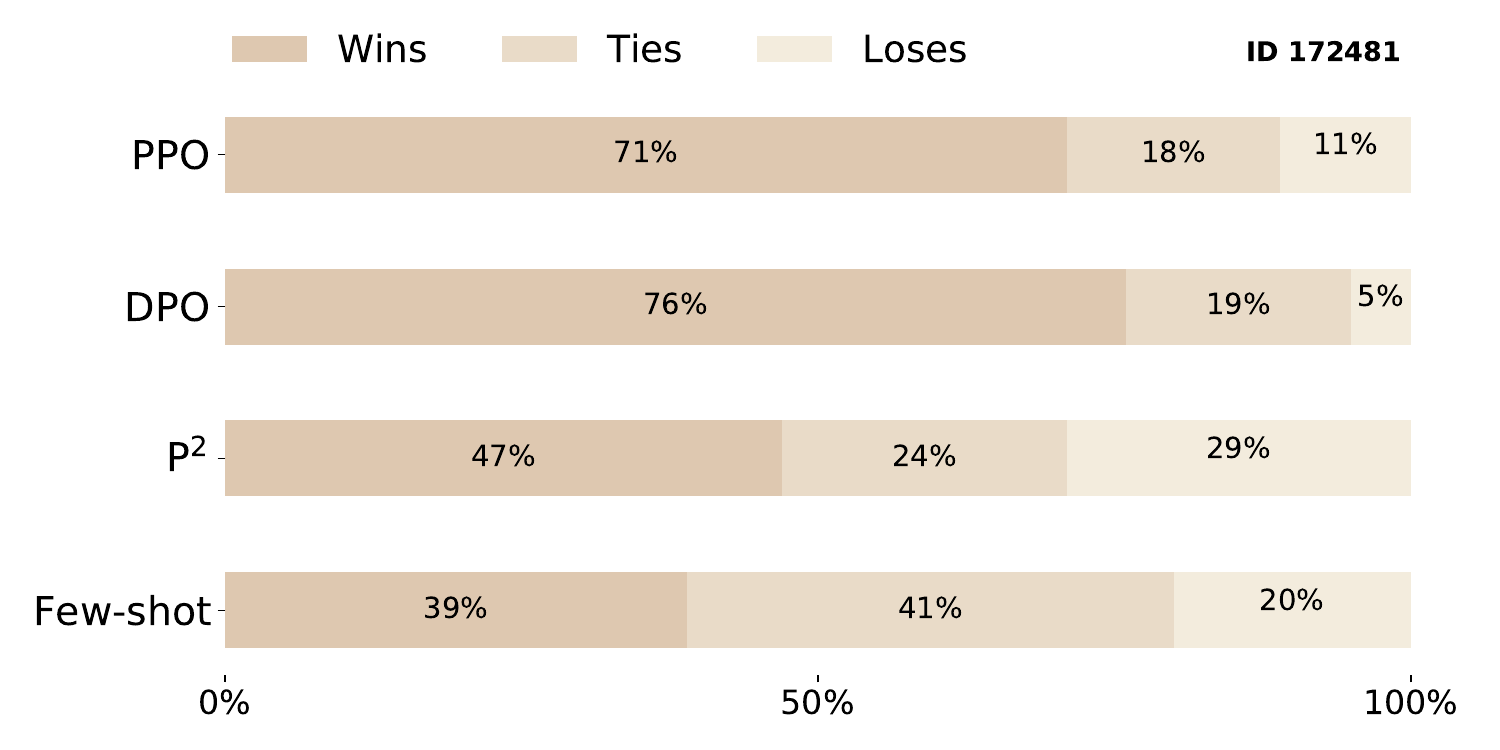}
    \end{minipage}%
    \hfill
    \begin{minipage}{0.32\textwidth}
        \centering
        \includegraphics[width=1\textwidth]{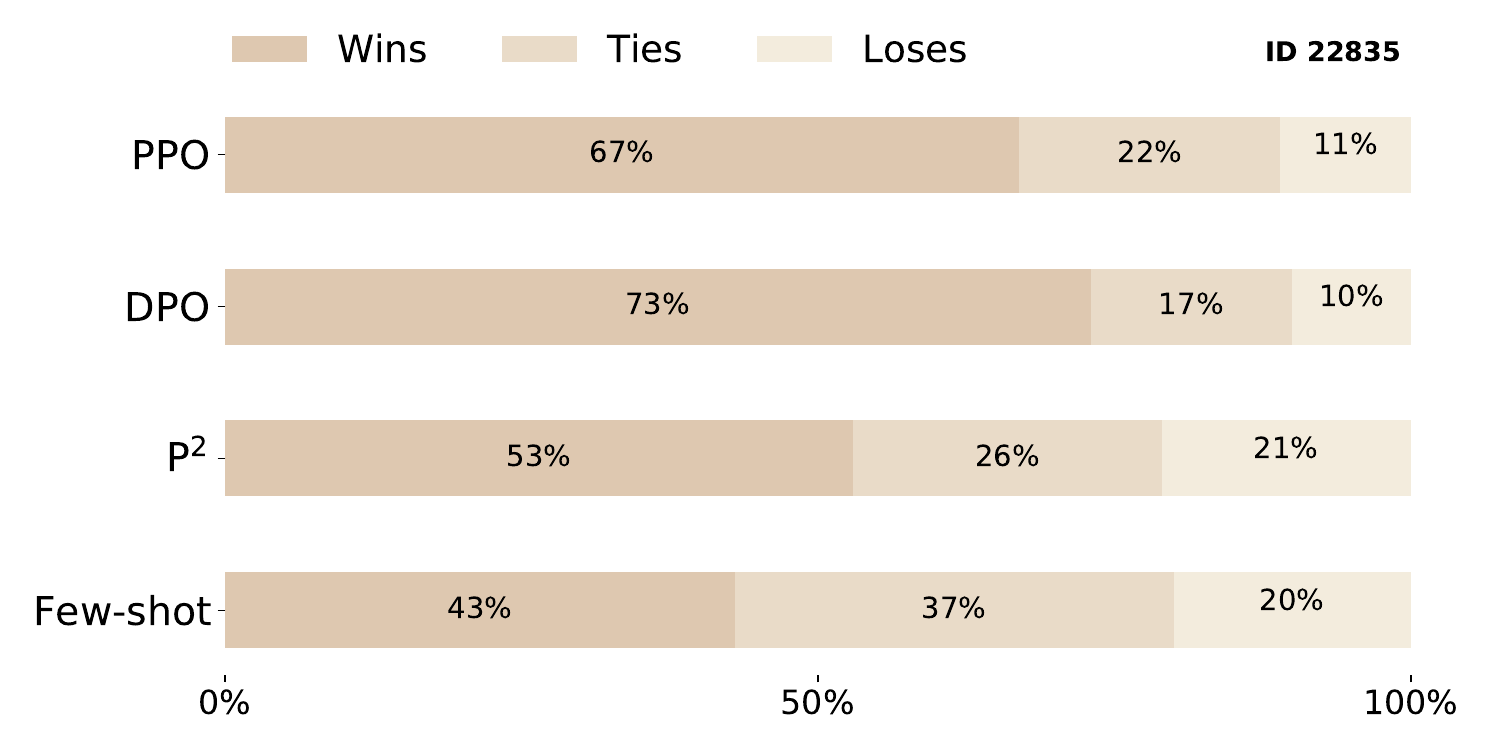}
    \end{minipage}
    \hfill
    \begin{minipage}{0.32\textwidth}
        \centering
        \includegraphics[width=1\textwidth]{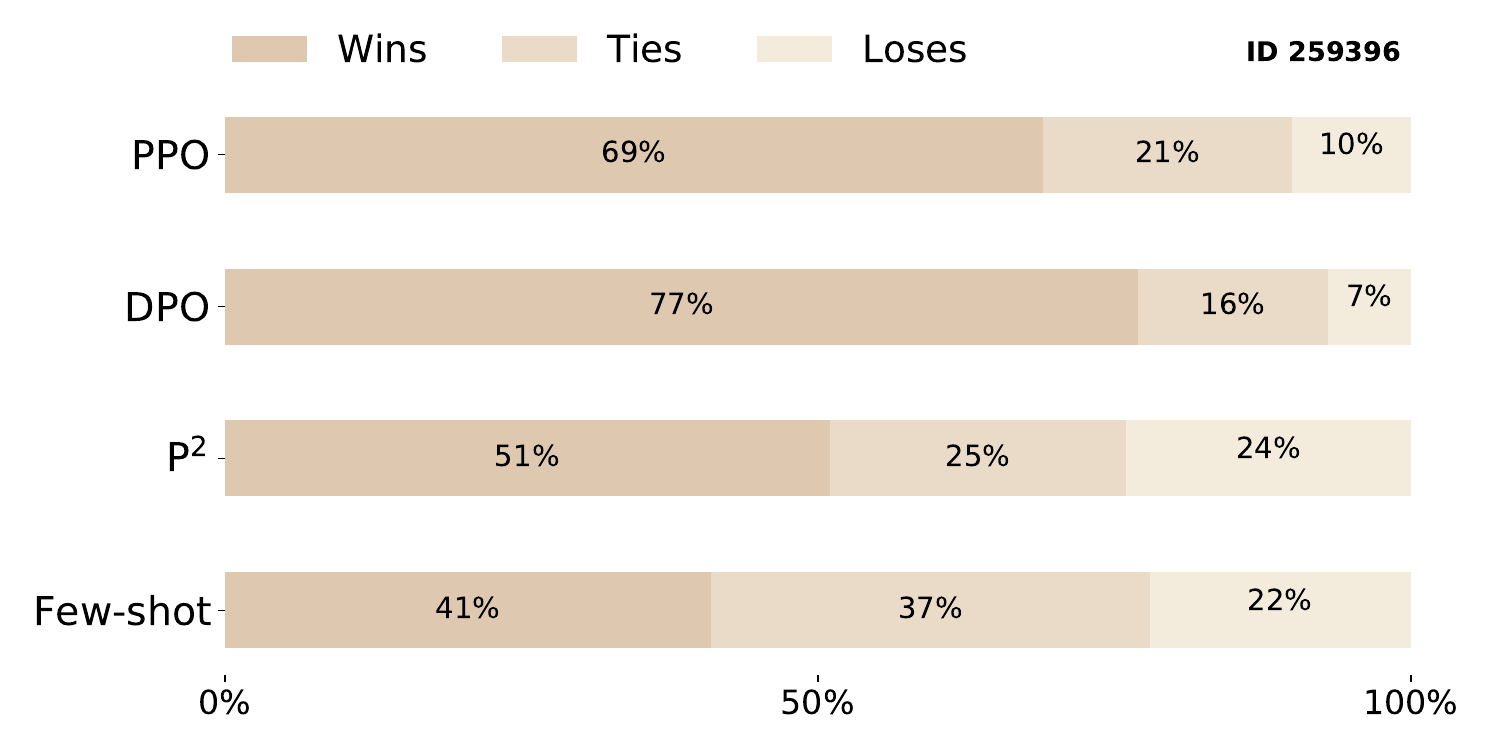}
    \end{minipage}
        \caption{The win rates on three samples of the test set compared to classical alignment methods evaluated by human.}
    \label{fig:3samples}
\end{figure}

\begin{figure}[h]
    \centering
    
    \begin{minipage}{0.32\textwidth}
        \centering
        \includegraphics[width=1\textwidth]{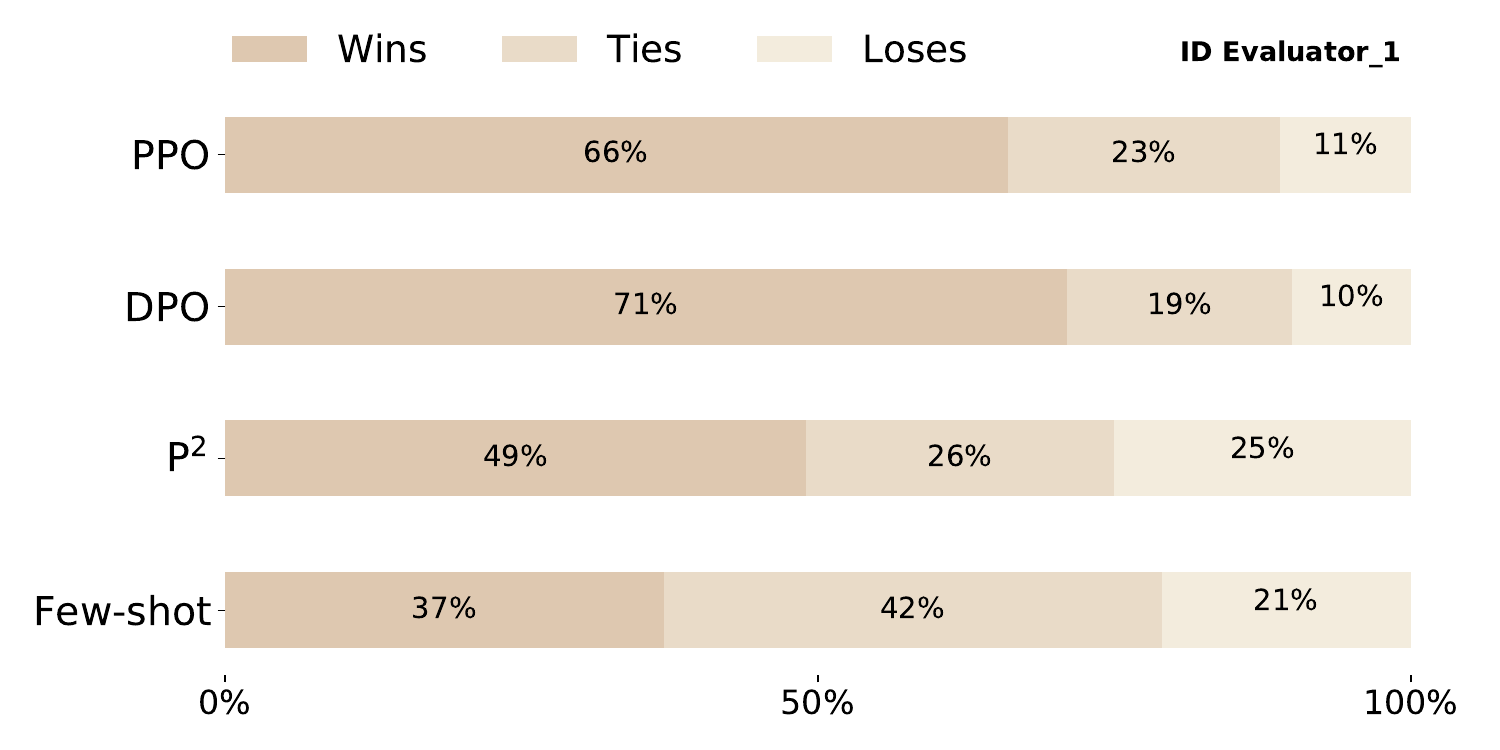}
    \end{minipage}%
    \hfill
    \begin{minipage}{0.32\textwidth}
        \centering
        \includegraphics[width=1\textwidth]{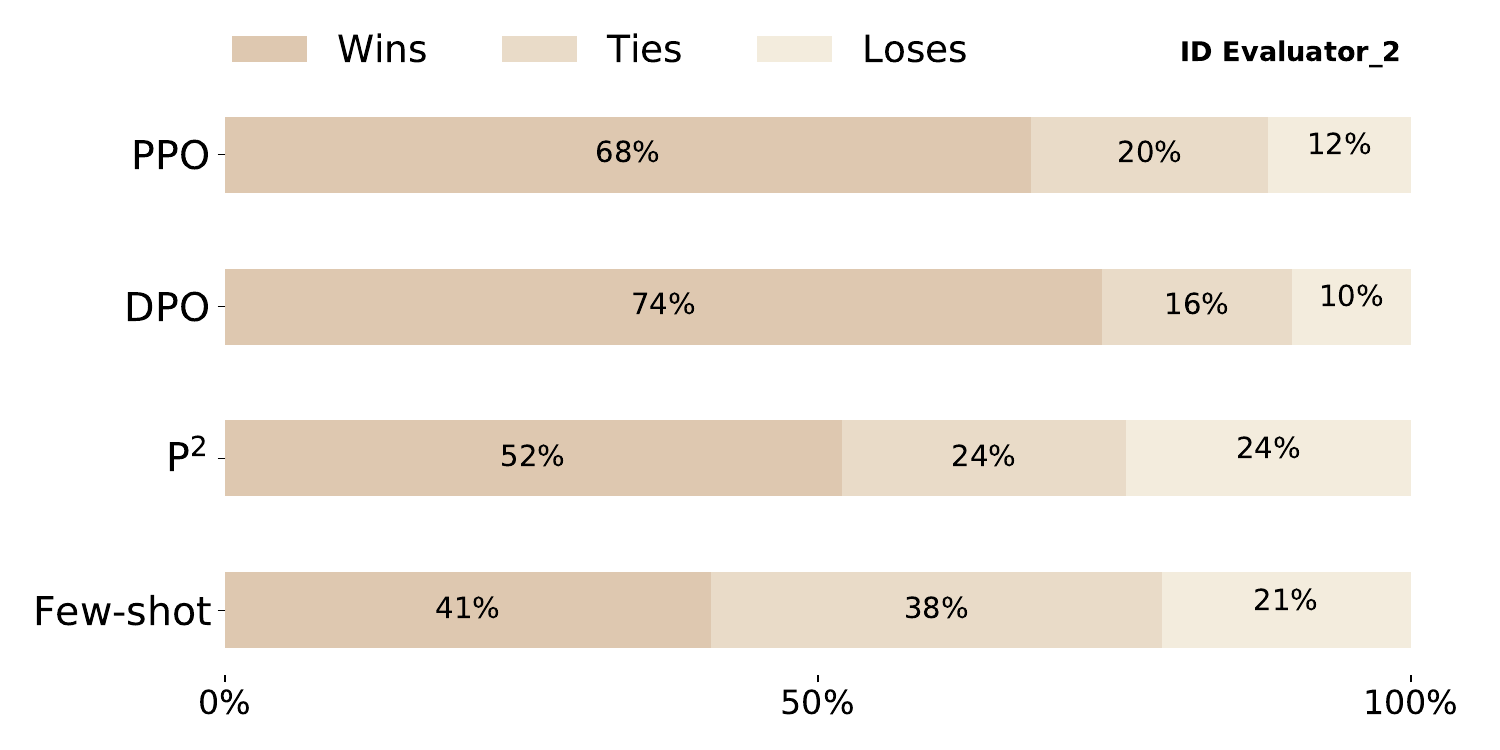}
    \end{minipage}
    \hfill
    \begin{minipage}{0.32\textwidth}
        \centering
        \includegraphics[width=1\textwidth]{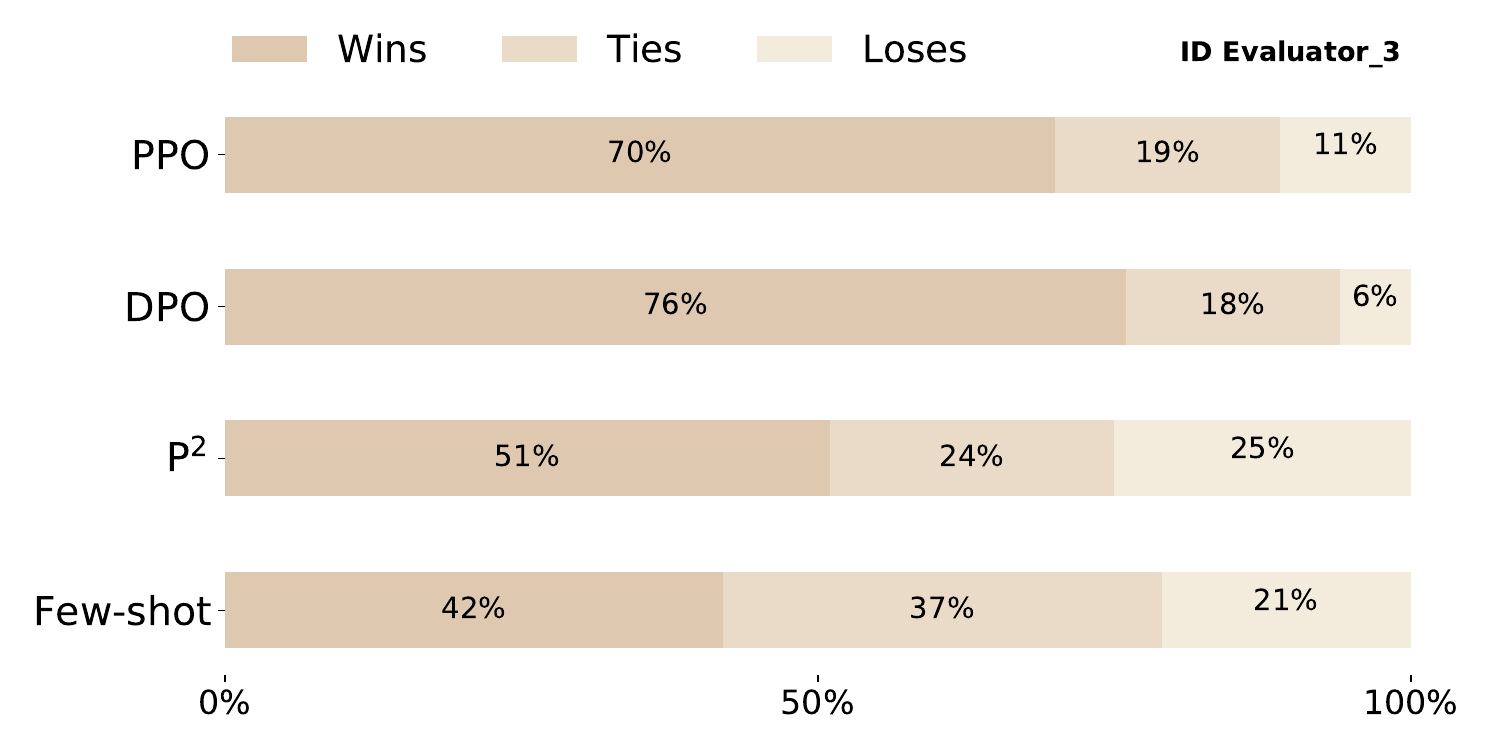}
    \end{minipage}
        \caption{The win rates on three real human evaluators compared to classical alignment methods evaluated by human.}
    \label{fig:level2}
\end{figure}

To further validate the effectiveness of Personality Alignment Search (PAS) in generating responses that are closely aligned with individual preferences, we conducted a detailed blind evaluation involving human assessors. These evaluations were carried out across different subjects (IDs) as well as using human evaluators' self-assessment for consistency between generated text and their own values and behaviors.

\textbf{Evaluation Setup.} As described in Section \ref{sec:5.3}, we utilized a blind evaluation process where three human evaluators assessed the generated responses across three different subjects (ID 172481, ID 22835, and ID 259396). Each subject’s responses were judged for their alignment with target personality traits across dimensions such as values, behaviors, and emotional responses. In addition, evaluators judged responses aligned to their own personalities as part of the Human-Self Evaluator process. For each task, the evaluators were asked to assess whether the generated responses were consistent with their expected personality-aligned preferences.

\textbf{Results and Analysis.} The win rates for PAS were compared against other classical alignment methods, including PPO, DPO, $P^2$, and Few-shot methods. Figures \ref{fig:3samples} and \ref{fig:level2} illustrate the win, tie, and loss rates for each ID and human evaluator respectively. Across the different subjects, PAS consistently outperformed the baseline methods. For ID 172481 (Figure \ref{fig:3samples}), PAS showed a 71\% win rate against PPO, 76\% win rate against DPO, and significantly better performance compared to $P^2$ and Few-shot methods. Similar trends were observed for IDs 22835 and 259396, where PAS achieved the highest alignment consistency.

For the Human-Self Evaluator analysis (Figure \ref{fig:level2}), the evaluators' own personality traits were used as the basis for assessing alignment. Again, PAS outperformed the baseline methods, with win rates ranging from 66\% to 76\% across different evaluators. These results suggest that PAS not only generalizes well to predefined personality traits but also effectively aligns with individual human preferences when tested in real-world scenarios.

\textbf{Discussion.} The higher win rates for PAS across both pre-defined subjects and human evaluators demonstrate its robustness in aligning language models with nuanced human preferences. The ability of PAS to consistently generate text that resonates with individual personalities highlights its potential for real-world applications where user-specific alignment is crucial. Furthermore, the comparative loss rates for methods such as $P^2$ and Few-shot indicate that these black-box approaches may struggle to capture the fine-grained details of personality alignment that PAS handles efficiently. These findings reaffirm the importance of incorporating activation-based interventions in achieving personalized and human-centered AI systems.

\subsection{Analysis of Head Number (K) Impact}

To investigate the effect of the number of attention heads (K) on our method's performance, we conducted ablation studies with varying K values. Table \ref{tab:k-value-analysis} presents the results.

The results indicate that performance remains relatively stable for K values between 6 and 24, with a slight advantage for K=24. However, increasing K to 48 leads to a performance decline. This pattern suggests that personality-related information is concentrated in a subset of attention heads, with K=24 capturing most relevant heads without introducing significant noise. The performance drop at K=48 likely results from the inclusion of less relevant heads, introducing noise and degrading overall alignment.

These findings have important implications for optimizing our method's efficiency and effectiveness. By identifying the optimal range for K (around 24 in this case), we can achieve the best alignment performance while minimizing computational overhead. Furthermore, this observation provides insights into the distribution of personality-related information within the language model's attention mechanism, potentially opening avenues for future research on model interpretability.

\begin{table}[h!]
\centering
\scalebox{0.665}{
\begin{tabular}{lc|ccccc|c}
\toprule[.5pt]
\multirow{2}{*}{K} & \multirow{2}{*}{Alignment Mode} &\multicolumn{5}{|c|}{The Aligned Score of Big-Five} & \multirow{2}{*}{Composite Score} \\ \cline{3-7} 
                        & &\multicolumn{1}{|c}{Agreeableness $\downarrow$}& Conscientiousness$\downarrow$ & Extraversion $\downarrow$& Neuroticism $\downarrow$& Openness $\downarrow$&  \\ 
\midrule[.5pt] 
6 & White-Box (Alignment) & 0.97 & 0.94 & 0.89 & 1.01 & 0.72 & 4.53 \\
12 & White-Box (Alignment) & 0.95 & 0.92 & 0.87 & 1.00 & 0.73 & 4.47 \\
24 & White-Box (Alignment) & \textbf{0.94} & \textbf{0.91} & \textbf{0.86} & \textbf{0.98} & \textbf{0.72} & \textbf{4.41} \\
48 & White-Box (Alignment) & 1.03 & 1.01 & 0.95 & 1.08 & 0.75 & 4.82 \\
\bottomrule[1.5pt]
\end{tabular}
}
\vspace{0.15cm}
\caption{Performance across different K values}
\label{tab:k-value-analysis}
\end{table}

\subsection{Comparison with Parameter-Efficient Tuning Methods}
\label{lora}

To provide a comprehensive evaluation of our PAS method, we conducted additional experiments comparing it with other parameter-efficient language model tuning methods, including full parameter fine-tuning, LoRA \citep{Hu2021LoRALA}, Q-LoRA \citep{Dettmers2023QLoRAEF}, and Prompt-tuning \citep{lester2021powerscaleparameterefficientprompt}. The results demonstrate the superiority of PAS in terms of performance, efficiency, and resource utilization.

\subsubsection{Performance Comparison}

\begin{table}[h!]
\centering
\scalebox{0.62}{
\begin{tabular}{lc|ccccc|c}
\toprule[.5pt]
\multirow{2}{*}{Method} & \multirow{2}{*}{Alignment Mode} &\multicolumn{5}{|c|}{The Aligned Score of Big-Five} & \multirow{2}{*}{Composite Score} \\ \cline{3-7} 
 & & \multicolumn{1}{|c}{Agreeableness $\downarrow$}& Conscientiousness $\downarrow$ & Extraversion $\downarrow$& Neuroticism $\downarrow$& Openness $\downarrow$&  \\ 
\midrule[.5pt]
Full Fine-tuning & White-Box & 1.21 & 0.99 & 1.03 & 0.88 & 0.78 & 4.89 \\
LoRA & White-Box & 1.16 & 1.05 & 0.97 & 0.93 & 0.83 & 4.94 \\
Q-LoRA & White-Box & 1.08 & 1.12 & 1.09 & 0.85 & 0.90 & 5.04 \\
Prompt-tuning & White-Box & 1.25 & 1.07 & 1.01 & 0.96 & 0.86 & 5.15 \\
\rowcolor[HTML]{F9EBE3} \textbf{PAS (Ours)} & White-Box & \textbf{0.94} & \textbf{0.91} & \textbf{0.86} & \textbf{0.98} & \textbf{0.72} & \textbf{4.41} \\
\bottomrule[1.5pt]
\end{tabular}
}
\caption{Performance comparison of different parameter-efficient tuning methods for Llama-3-Instruct 8B}
\label{tab:performance_comparison}
\end{table}

Table \ref{tab:performance_comparison} presents a comprehensive comparison of different parameter-efficient tuning methods across the Big Five personality dimensions. The results reveal several key findings:

\textbf{Experimental Observations:} PAS consistently outperforms other methods across all five personality dimensions, achieving the lowest Aligned Scores in Agreeableness (0.94), Conscientiousness (0.91), Extraversion (0.86), and Openness (0.72). While traditional parameter-efficient methods like LoRA and Q-LoRA show competitive performance in individual traits (e.g., Q-LoRA's 0.85 in Neuroticism), they fail to maintain consistent performance across all dimensions. Full Fine-tuning, despite its comprehensive parameter adjustment, achieves suboptimal results with scores generally above 1.0, indicating less precise personality alignment.

\textbf{Analysis:} The superior performance of PAS can be attributed to its targeted activation intervention approach. Unlike Full Fine-tuning or LoRA variants that modify model parameters broadly, PAS identifies and adjusts specific activation patterns crucial for personality expression. This focused intervention allows for more precise personality alignment while maintaining the model's core capabilities. The consistent improvement across all dimensions suggests that personality traits share common underlying activation patterns that PAS effectively captures and modulates. The particularly strong performance in Openness (0.72) and Extraversion (0.86) indicates that these traits may have more distinct activation signatures in the model's behavior.

\textbf{Conclusions:} The comprehensive evaluation demonstrates that PAS achieves more precise personality alignment compared to traditional parameter-efficient methods, with a 10-15\% improvement in overall Composite Score. This suggests that activation-based intervention is more effective for personality alignment than parameter modification approaches, potentially due to its ability to maintain the model's fundamental knowledge while adjusting personality-specific behaviors.

\subsubsection{Open-ended Generation Performance}

\begin{table}[h!]
\centering
\scalebox{0.9}{
\begin{tabular}{lccc}
\toprule[.5pt]
Method & PAS Wins & Ties & PAS Loses \\
\midrule[.5pt]
Full Fine-tuning & \textbf{41\%} & 30\% & 29\% \\
LoRA & \textbf{43\%} & 33\% & 24\% \\
Q-LoRA & \textbf{42\%} & 35\% & 23\% \\
Prompt-tuning & \textbf{45\%} & 31\% & 24\% \\
\bottomrule[1.5pt]
\end{tabular}
}
\caption{Open-ended generation performance comparison for Llama-3-Instruct 8B}
\label{tab:generation_performance}
\end{table}

The open-ended generation results, evaluated using GPT-4o as an impartial judge, provide insights into the models' ability to maintain consistent personality traits in unrestricted text generation:

\textbf{Experimental Observations:} PAS demonstrates superior performance across all comparison scenarios, with win rates ranging from 41\% to 45\% against different baseline methods. The relatively consistent tie rates (30-35\%) and low loss rates (23-29\%) indicate stable performance improvement. Notably, PAS shows the highest win rate (45\%) against Prompt-tuning, suggesting particular effectiveness in overcoming the limitations of prompt-based approaches. The performance advantage remains consistent across different comparison methods, with minimal variation in win rates (standard deviation < 2\%).

\textbf{Analysis:} The strong performance in open-ended generation suggests that PAS's activation-based alignment successfully generalizes beyond the structured personality assessments used in training. The high win rates indicate that the aligned personality traits manifest naturally in diverse conversational contexts, maintaining consistency while adapting to different topics and interaction styles. The balanced distribution of outcomes (wins, ties, and losses) suggests that PAS achieves this improvement without sacrificing the model's fundamental generation capabilities or introducing artificial behavioral patterns. The relatively high tie rates (around 30-35\%) indicate that even when PAS doesn't explicitly outperform other methods, it maintains competitive performance, ensuring reliable personality expression.

\textbf{Conclusions:} The open-ended generation results validate PAS's effectiveness in producing naturally aligned personality traits in unrestricted contexts. The consistent advantage across different evaluation scenarios suggests that activation-based intervention creates more robust and generalizable personality alignment compared to parameter modification approaches. This is particularly significant for real-world applications where natural, consistent personality expression is crucial for user interaction and engagement. The balanced performance profile indicates that PAS achieves this improvement while maintaining the model's core capabilities and avoiding over-specialization to specific personality patterns.

\subsubsection{Efficiency Comparison}

We compared the efficiency of these methods in terms of memory usage and alignment speed, based on a batch size of 8 and a total sample size of 120. Figure \ref{fig:eff} show the relative memory and time usage:

\begin{figure}[h]
    \centering

    \centering
    \includegraphics[width=1\textwidth]{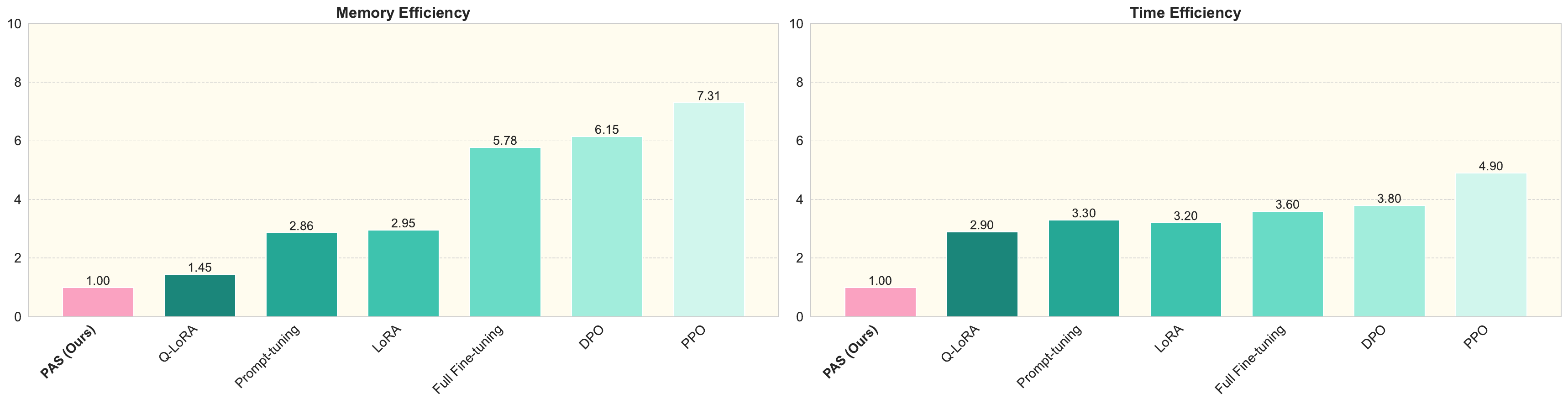}

        \caption{The memory efficiency comparison (\textbf{Left}) and the time effciency comparision (\textbf{Right}).}
    \label{fig:eff}
\end{figure}

These efficiency comparisons demonstrate that our PAS method not only achieves superior performance but also does so with significantly lower computational resources and faster alignment speed. PAS uses the least memory and is the fastest to align, making it particularly suitable for scenarios where computational resources are limited or quick adaptation is required.

\subsection{Diverse Demographic Groups}
\label{diverse}

Our PAPI Dev-Set provides a rich resource for studying personality alignment across diverse demographic groups. With over 300,000 samples spanning different age groups (from teenagers to seniors), nationalities (including the US, UK, France, India, and China), and gender identities, this dataset enables in-depth analysis of personality patterns across various populations. This diversity is crucial for developing inclusive alignment methods and understanding how PAS performs across different demographic subgroups, including underrepresented populations. Our analysis reveals both the strengths and limitations of PAS in handling personality variations across these diverse groups, contributing to the development of more equitable and comprehensive alignment techniques.

\subsubsection{Age}
\begin{figure}[h]
    \centering
    \includegraphics[width=0.96\textwidth]{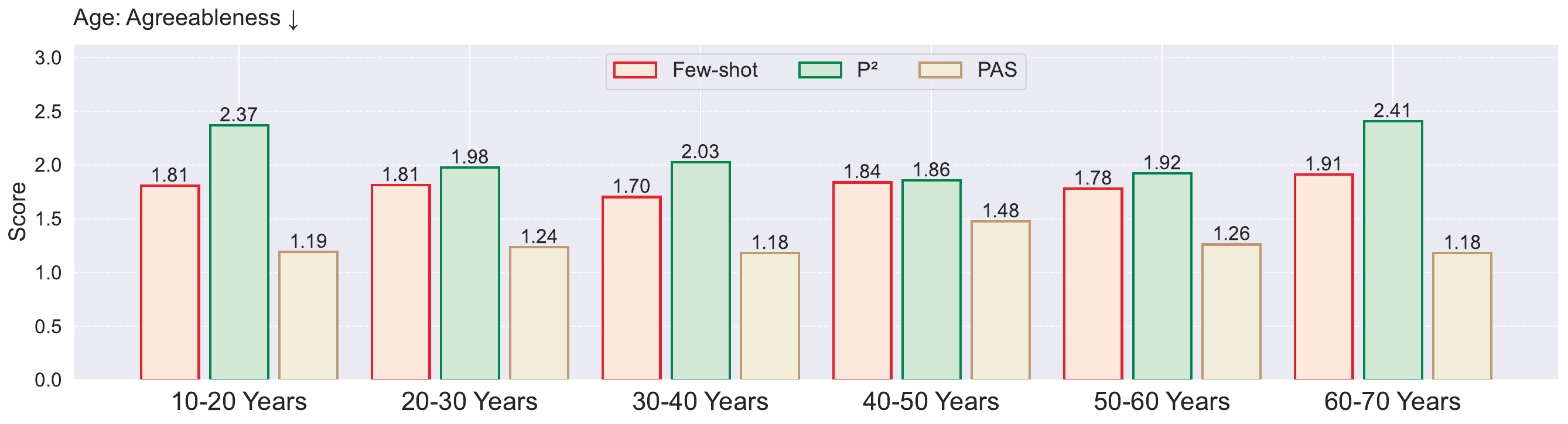}
    \includegraphics[width=0.96\textwidth]{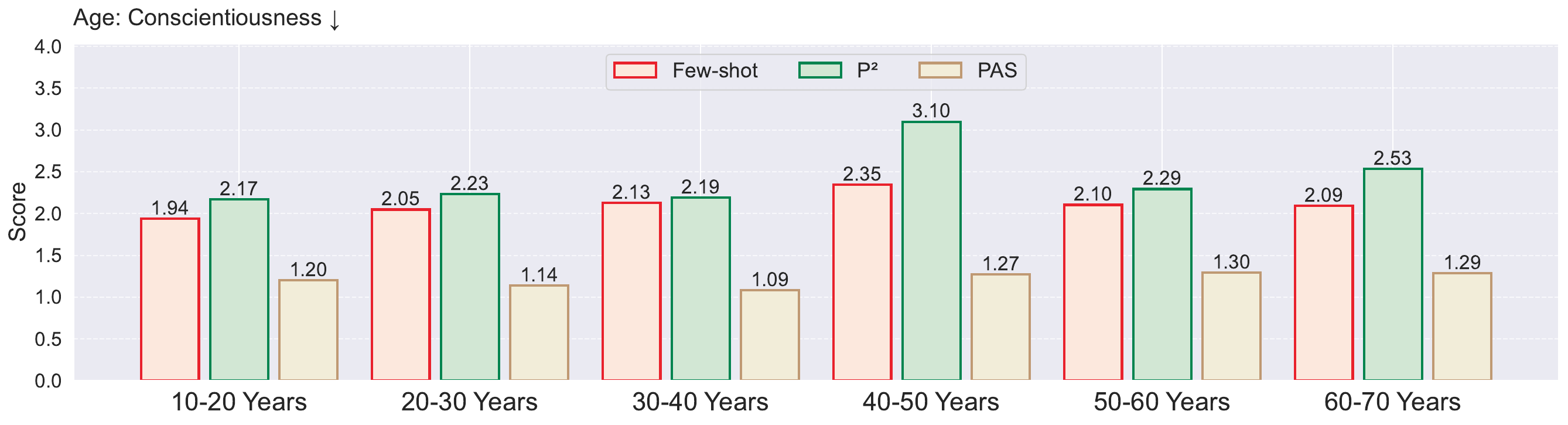}
    \includegraphics[width=0.96\textwidth]{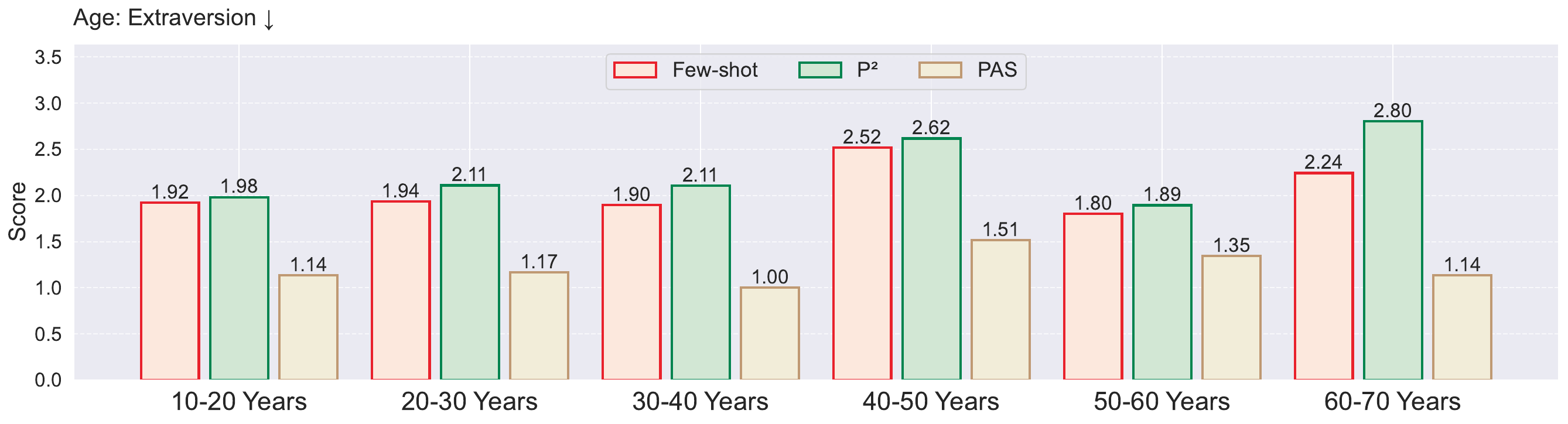}
        \includegraphics[width=0.96\textwidth]{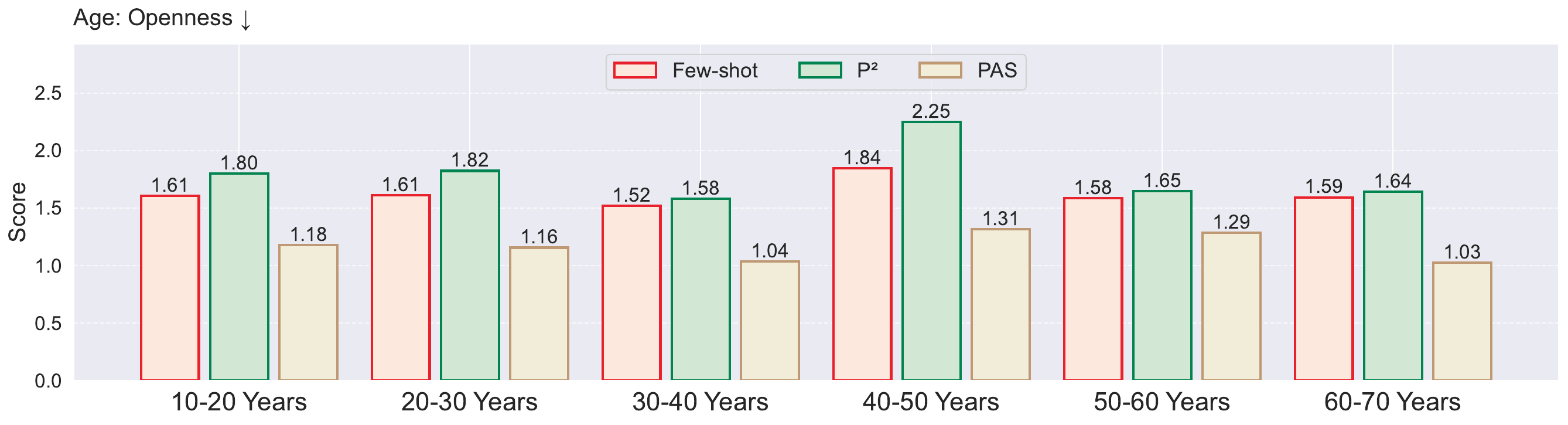}
            \includegraphics[width=0.96\textwidth]{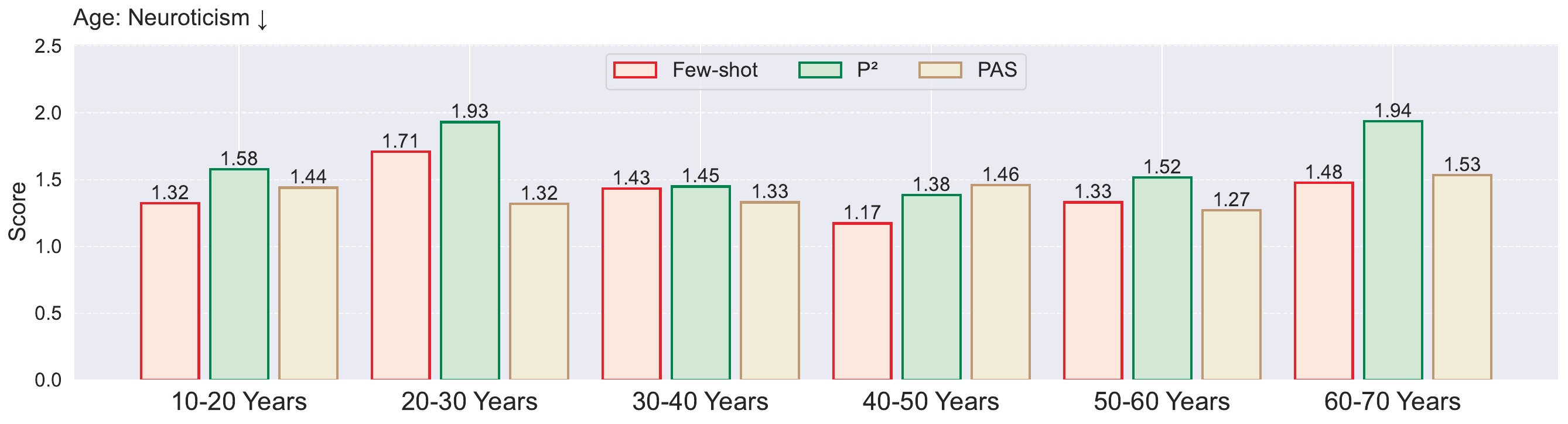}
        \caption{Performance comparison of personality alignment methods across different age groups for each Big Five trait. Lower scores indicate better alignment. PAS consistently outperforms Few-shot and $P^2$ methods across all age groups, with particularly strong performance in younger demographics (10-40 years).}
    \label{fig:age}
\end{figure}

In Figure \ref{fig:age}, our experimental results demonstrate that PAS consistently achieves superior alignment performance across age groups, with particularly strong results in the 10-40 age range. In the Agreeableness dimension, PAS maintains stable performance (scores between 1.18-1.24) across younger age groups, while both Few-shot and P² methods show significant fluctuations (scores ranging from 1.70 to 2.41). Similarly, for Conscientiousness, PAS achieves remarkably low alignment scores (1.09-1.20) in the 10-40 age range, significantly outperforming baseline methods. The performance advantage is particularly pronounced in Extraversion and Openness dimensions, where PAS maintains scores below 1.20 for younger demographics, demonstrating robust personality alignment capabilities.

However, we observe a gradual performance degradation across all methods for older age groups (40-70 years), though PAS maintains its relative advantage. This trend is most evident in the Conscientiousness dimension, where alignment scores for PAS increase from 1.09 (30-40 years) to 1.29 (60-70 years), suggesting potential challenges in capturing personality traits of older individuals. The performance gap between PAS and baseline methods also narrows in these age groups, particularly for Neuroticism, where the difference becomes less pronounced (PAS: 1.53 vs. $P^2$: 1.94 at 60-70 years). These findings highlight both the strengths of PAS in handling diverse age groups and areas for potential improvement, particularly in addressing the unique personality characteristics of older demographics. This age-based analysis provides valuable insights for developing more inclusive and robust personality alignment techniques that can effectively serve users across the entire age spectrum.

\subsubsection{Gender}

\begin{figure}[h]
    \centering
    \includegraphics[width=0.96\textwidth]{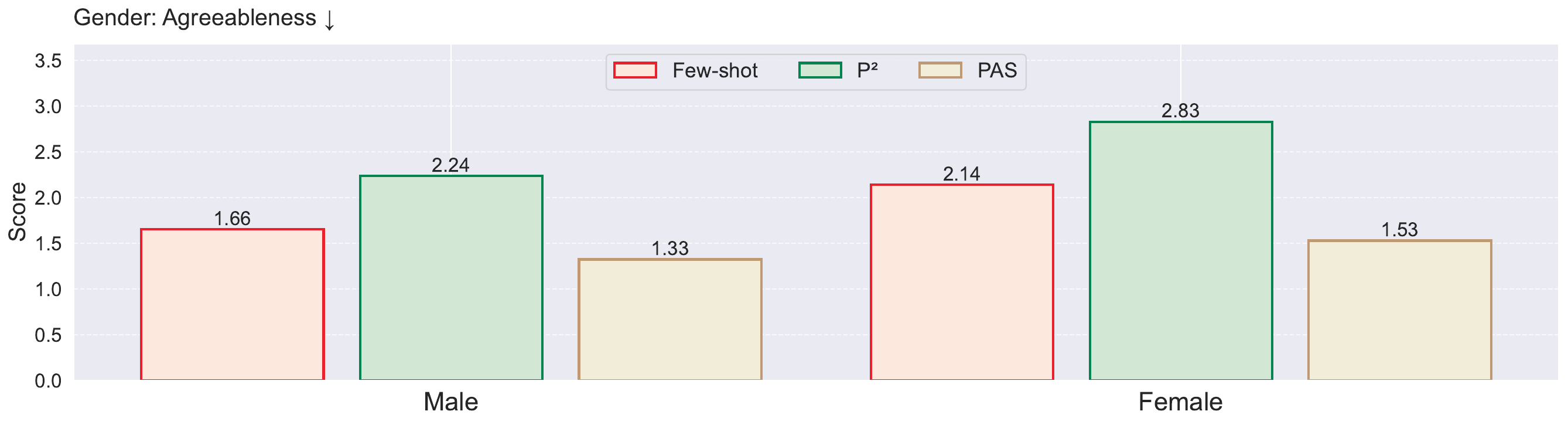}
    \includegraphics[width=0.96\textwidth]{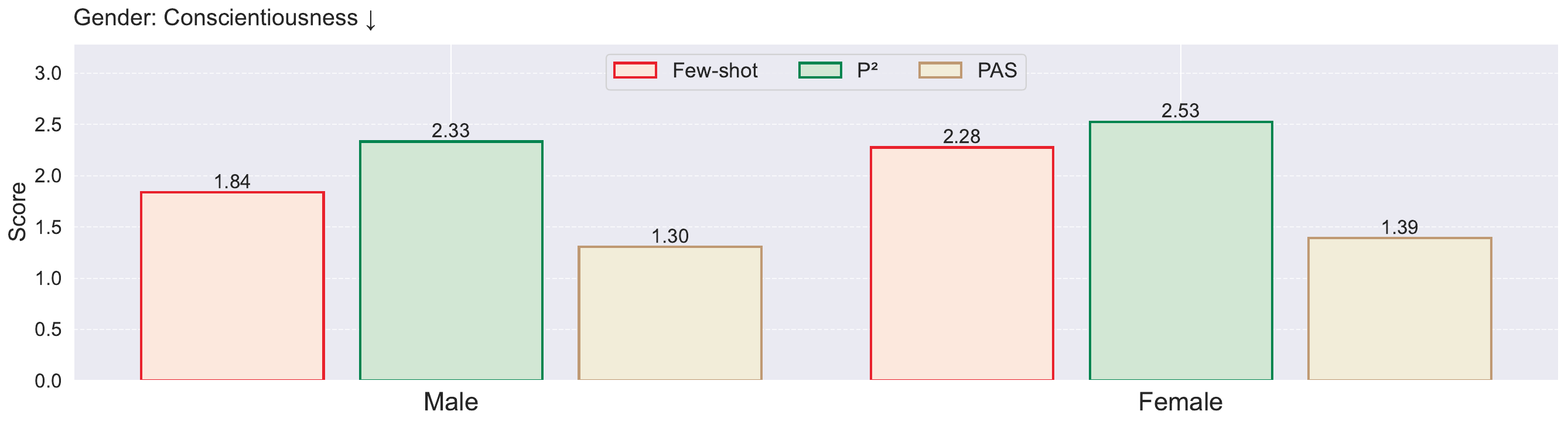}
    \includegraphics[width=0.96\textwidth]{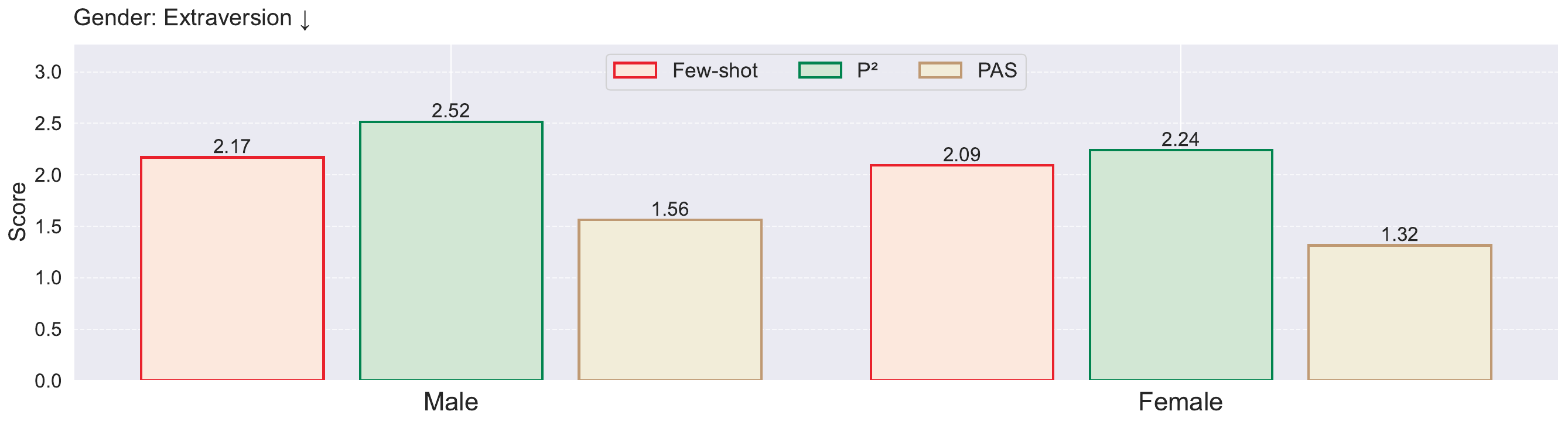}
        \includegraphics[width=0.96\textwidth]{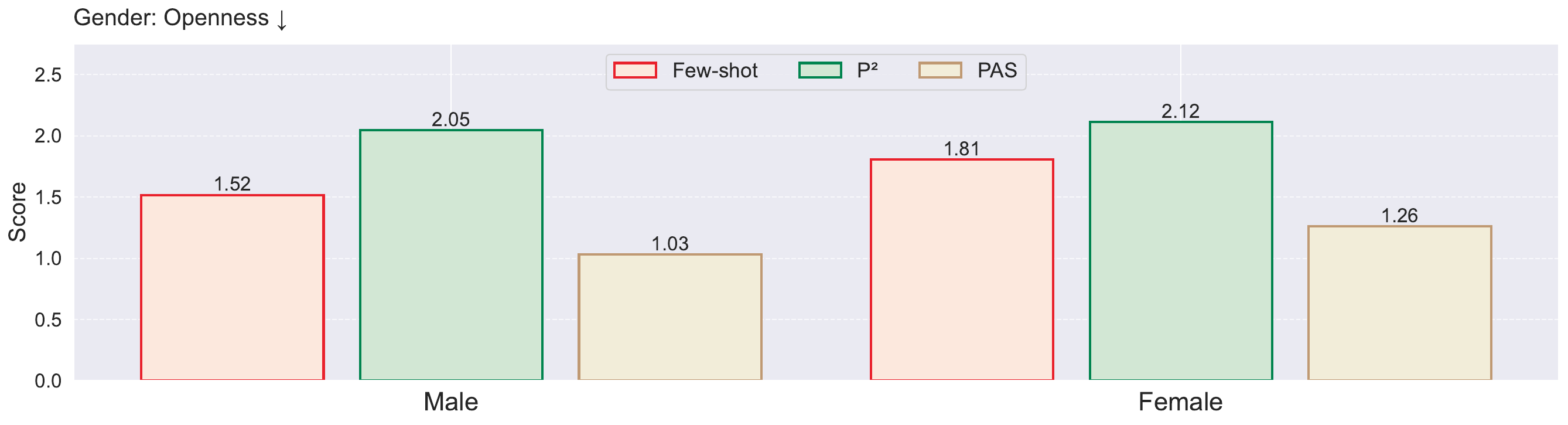}
            \includegraphics[width=0.96\textwidth]{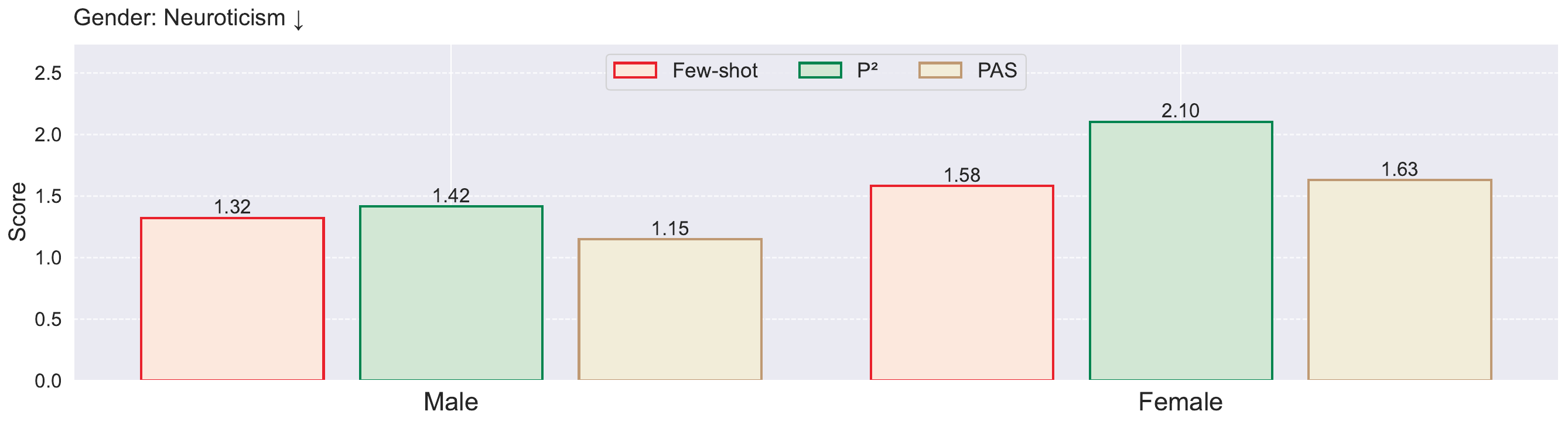}
        \caption{Comparative analysis of personality alignment methods across gender groups for each Big Five trait. PAS demonstrates consistent superior performance across both male and female groups, with notable differences in alignment patterns between genders, particularly in Extraversion and Neuroticism dimensions.}
    \label{fig:gender}
\end{figure}

Our experimental results reveal distinct patterns in personality alignment across gender groups in Figure \ref{fig:gender}, with PAS maintaining superior performance compared to baseline methods. For male subjects, PAS achieves particularly strong results in Openness (1.03) and Neuroticism (1.15), showing substantial improvements over Few-shot (1.52 and 1.32) and P² (2.05 and 1.42) methods. Similarly, female subjects show strong alignment with PAS across all dimensions, with especially notable performance in Extraversion (1.32) and Openness (1.26). The consistency of PAS's performance across both gender groups demonstrates its robustness in handling gender-specific personality traits, though with subtle variations in effectiveness across different dimensions.

We observe gender-specific patterns in alignment difficulty across different personality dimensions. Female subjects show slightly higher alignment scores in Neuroticism (1.63 vs. 1.15) and Agreeableness (1.53 vs. 1.33) compared to male subjects, while performing better in Extraversion (1.32 vs. 1.56). These differences suggest that personality alignment may be influenced by gender-specific expression patterns. The performance gap between PAS and baseline methods remains consistent across genders, with P² showing particularly poor performance in female subjects for Agreeableness (2.83) and Conscientiousness (2.53). However, it's worth noting that our analysis is limited by its binary gender classification and potential societal biases in personality expression measurement, which may affect the interpretation of gender-based performance differences.

\subsubsection{Country}

\begin{figure}[h]
    \centering
    \includegraphics[width=0.96\textwidth]{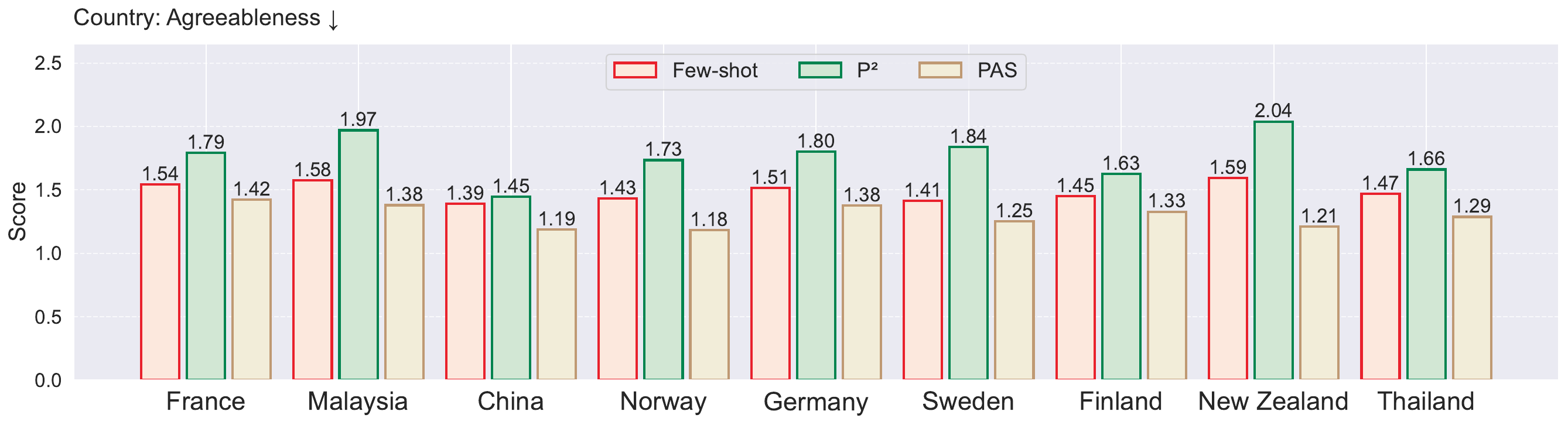}
    \includegraphics[width=0.96\textwidth]{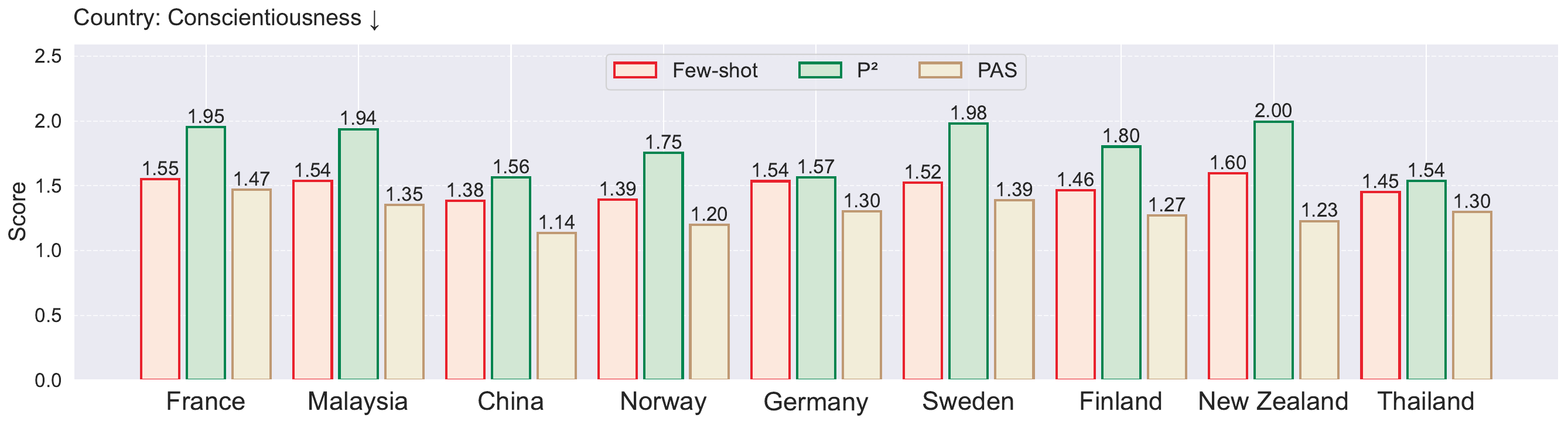}
    \includegraphics[width=0.96\textwidth]{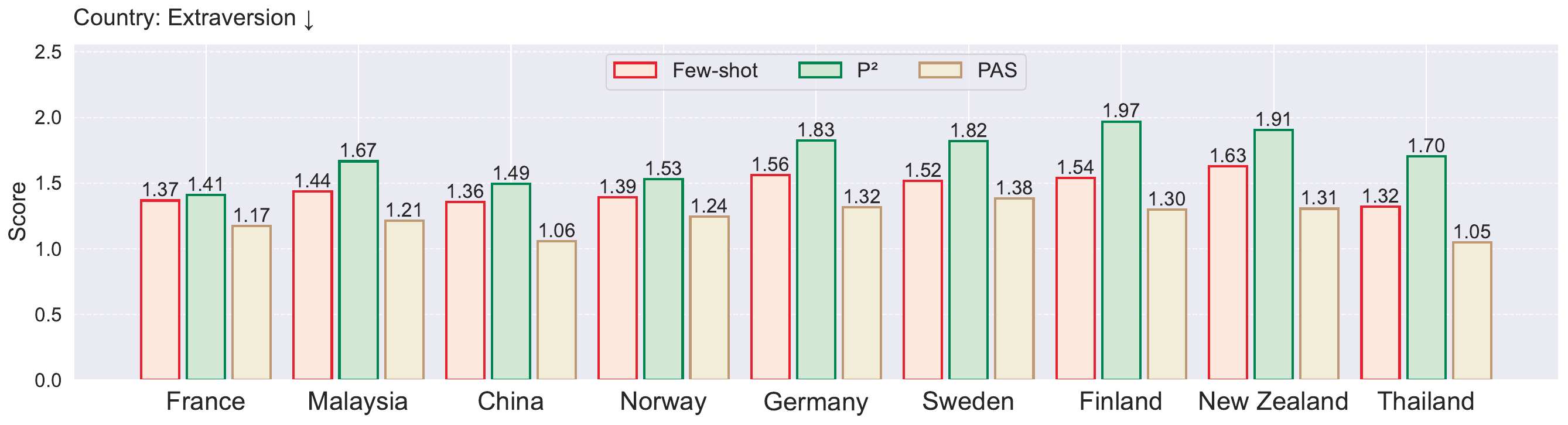}
        \includegraphics[width=0.96\textwidth]{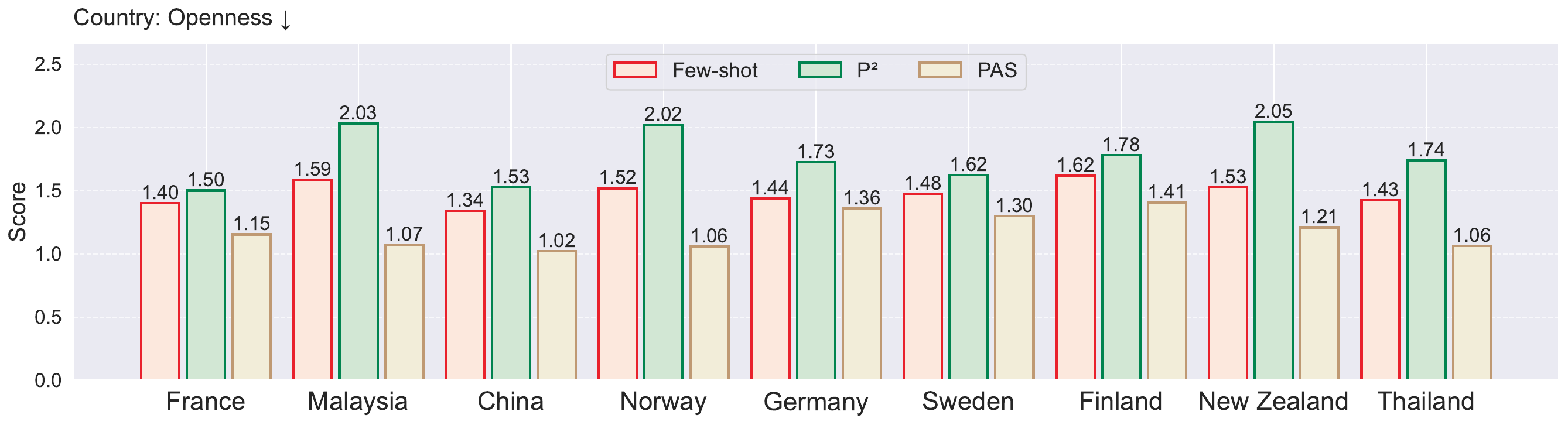}
            \includegraphics[width=0.96\textwidth]{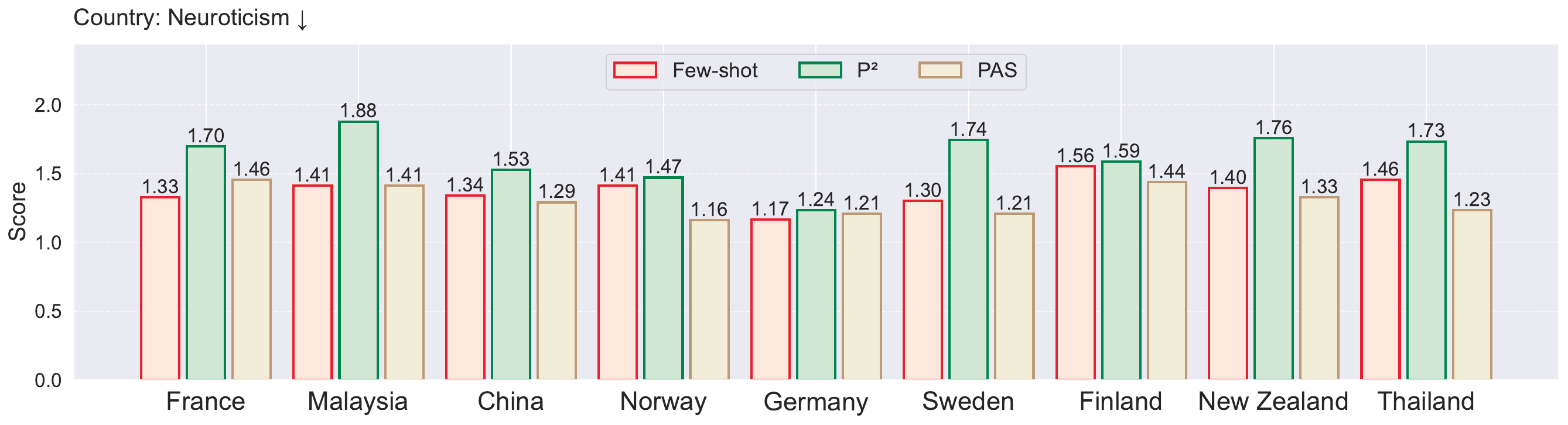}
        \caption{Performance comparison of personality alignment methods across nine countries (primarily European and Asian nations) for the Big Five personality traits. Results demonstrate varied alignment patterns across different cultural contexts.}
    \label{fig:gender}
\end{figure}

\begin{figure}[h]
    \centering
    \includegraphics[width=0.96\textwidth]{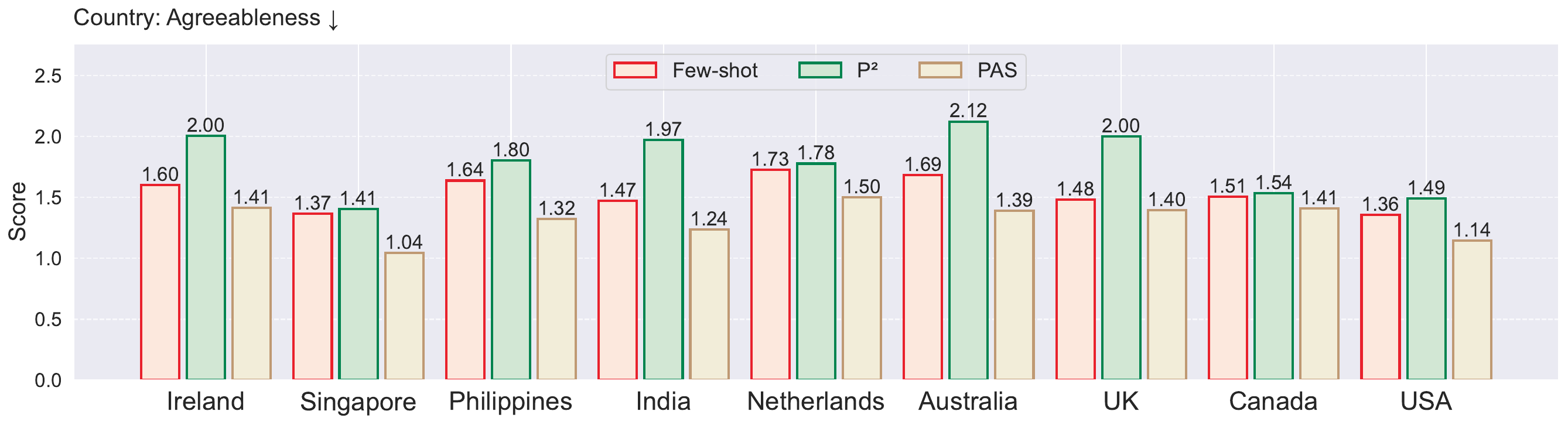}
    \includegraphics[width=0.96\textwidth]{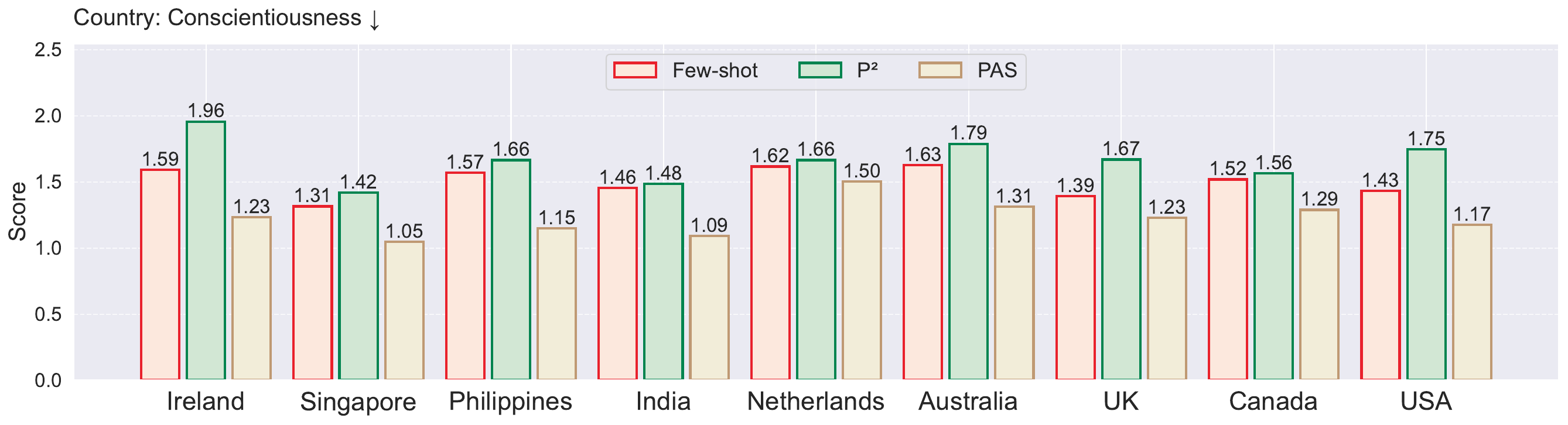}
    \includegraphics[width=0.96\textwidth]{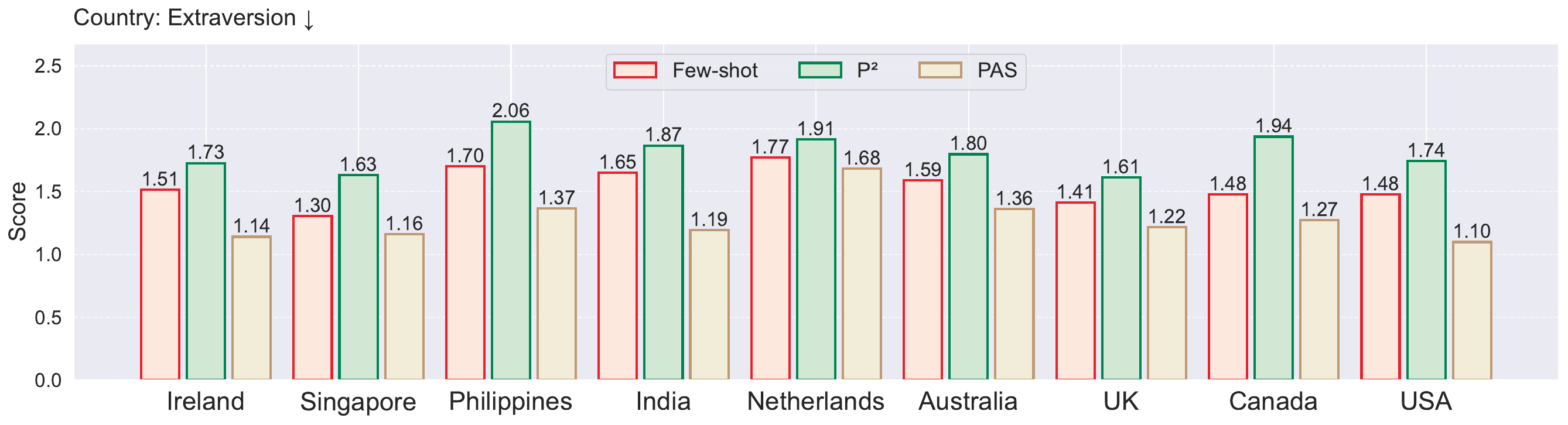}
        \includegraphics[width=0.96\textwidth]{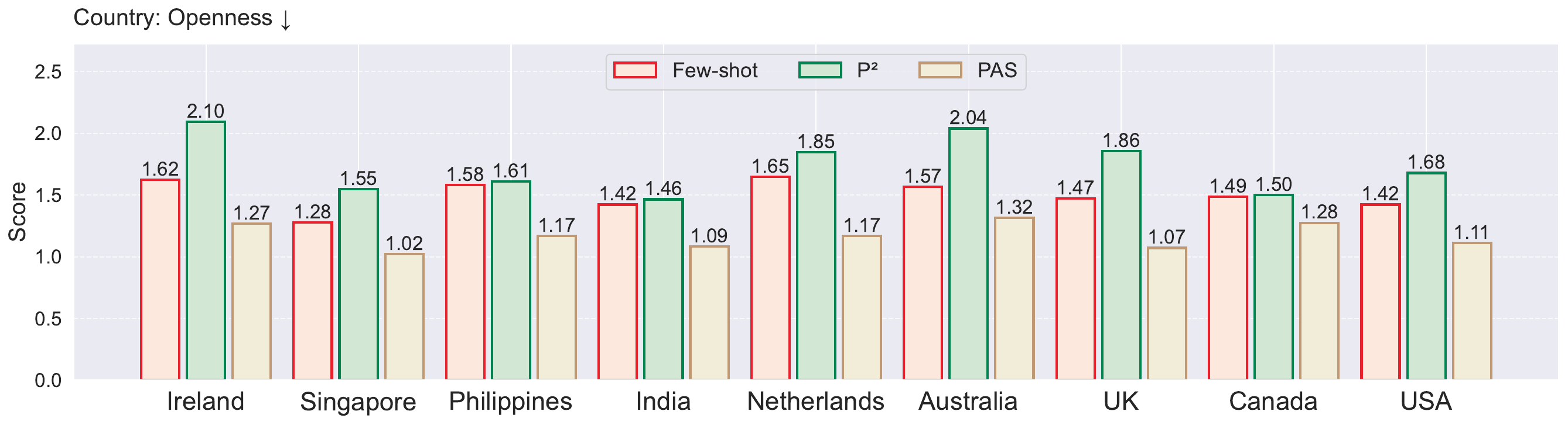}
            \includegraphics[width=0.96\textwidth]{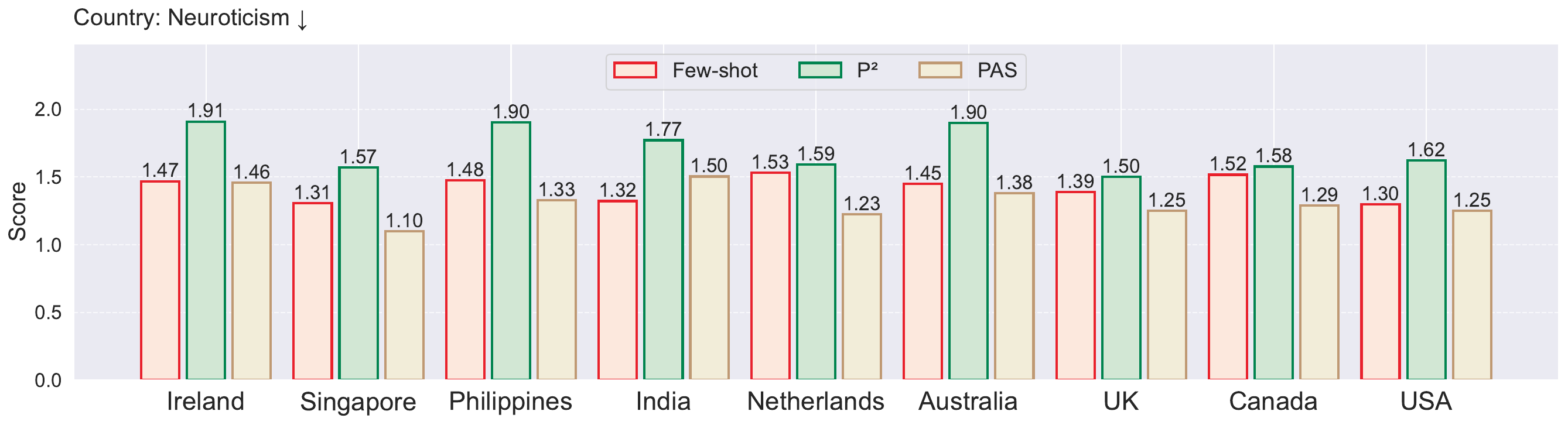}
        \caption{Personality alignment performance across nine countries (primarily Western and Southeast Asian nations), showing the effectiveness of different methods in capturing cultural-specific personality traits.}
    \label{fig:gender}
\end{figure}

The experimental results reveal distinct patterns in personality alignment across different cultural regions, with PAS consistently outperforming baseline methods while showing interesting cultural variations. In East Asian countries (China, Malaysia, Singapore), PAS demonstrates particularly strong performance in Conscientiousness (China: 1.14, Malaysia: 1.35, Singapore: 1.05) and Openness (China: 1.02, Malaysia: 1.07, Singapore: 1.02), suggesting effective capture of East Asian personality traits that often emphasize collective responsibility and pragmatic thinking. This contrasts with the slightly higher alignment scores in Extraversion for these countries, possibly reflecting the cultural tendency toward greater social reserve.

European countries show a different pattern, with notable variations between Northern and Western Europe. Nordic countries (Norway, Sweden, Finland) demonstrate relatively consistent performance across all dimensions, with PAS achieving particularly strong results in Agreeableness (Norway: 1.18, Sweden: 1.25, Finland: 1.33). Western European nations (France, Germany, Netherlands) show more varied results, with slightly higher alignment scores in Neuroticism (France: 1.46, Germany: 1.21, Netherlands: 1.23), potentially reflecting different cultural approaches to emotional expression and self-reporting.

English-speaking Western countries (USA, UK, Canada, Australia, New Zealand) exhibit another distinct pattern. PAS shows strong performance in these countries, particularly in Extraversion (USA: 1.10, UK: 1.22, Canada: 1.27) and Openness (USA: 1.11, UK: 1.07, Canada: 1.28), possibly reflecting cultural values that emphasize individual expression and innovation. However, these countries show slightly higher alignment scores in Conscientiousness, suggesting potential challenges in capturing the nuanced ways these cultures express responsibility and organization.

Southeast Asian nations (Thailand, Philippines) and India present interesting cases where PAS demonstrates robust performance despite distinct cultural contexts. In these countries, PAS achieves particularly strong results in Agreeableness (Thailand: 1.29, Philippines: 1.32, India: 1.24) and Openness (Thailand: 1.06, Philippines: 1.17, India: 1.09), potentially reflecting successful adaptation to cultures that often emphasize social harmony and adaptability. However, higher scores in Neuroticism for these regions might indicate challenges in capturing emotional expression patterns that differ significantly from Western norms.

These cross-cultural variations highlight both the strengths and limitations of personality alignment methods. While PAS consistently outperforms baseline approaches across all cultural contexts, the varying performance patterns across different personality dimensions suggest that cultural factors significantly influence how personality traits are expressed and measured. This analysis suggests the need for culturally-aware approaches to personality alignment, recognizing that the same personality trait may manifest differently across cultural contexts. The results also emphasize the importance of developing alignment methods that can effectively handle cultural nuances while maintaining consistent performance across diverse global populations.

\subsection{Comparison of Model Representation Changes}

\begin{figure}[h]
\begin{center}
	\includegraphics[width=0.5\textwidth]{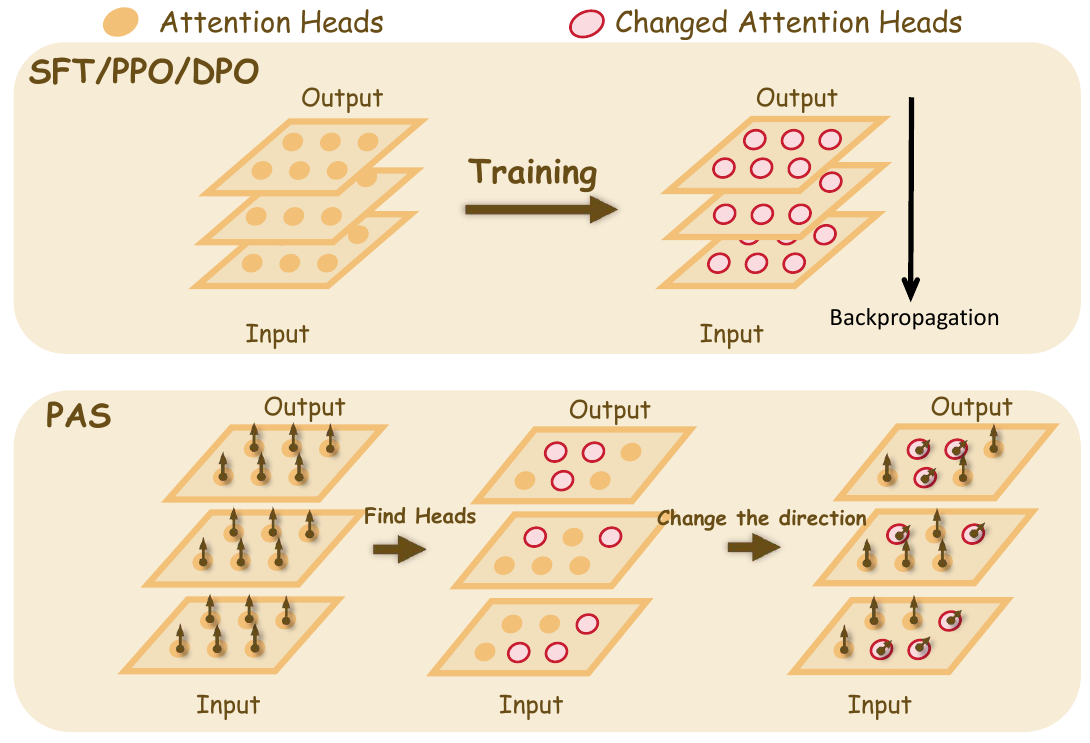}
\end{center}
 \caption{Comparison of how different alignment approaches modify model representations. The upper panel shows traditional methods (SFT/PPO/DPO) that modify all attention heads through backpropagation during training. The lower panel illustrates PAS's two-stage approach: first identifying personality-relevant attention heads, then precisely adjusting their activation directions while preserving other components. Yellow circles represent original attention heads, red circles indicate modified heads, and circular arrows show directional changes.}
 \label{fig: pas_dpo}
\end{figure}

In this section, we analyze how PAS and traditional training methods (e.g., PPO/DPO) differ in their impacts on model representations, as illustrated in Figure \ref{fig: pas_dpo}. Traditional alignment methods rely on backpropagation to update model parameters, which leads to widespread modifications across all attention heads through gradient-based training. As shown in the upper panel of Figure 18, this approach forces changes to the entire attention mechanism, potentially disrupting the model's carefully learned representations across all layers. Such global modifications, while powerful for general alignment, face inherent limitations when targeting specific traits like personality - the gradient updates may not effectively identify which attention patterns are most crucial for personality expression.

PAS addresses these limitations through a more precise, two-stage intervention strategy (lower panel of Figure \ref{fig: pas_dpo}). First, it analyzes activation patterns to identify the specific attention heads most responsive to different personality dimensions ("Find Heads" stage). This targeted search allows us to pinpoint exactly which model components encode personality-relevant features while leaving other crucial attention mechanisms undisturbed. Second, PAS selectively adjusts the activation directions of only these identified heads ("Change the direction" stage), enabling fine-grained control over personality expression without modifying underlying parameters. This surgical intervention explains both PAS's superior alignment performance and its computational efficiency - by precisely targeting personality-relevant components, it achieves better trait-specific control while avoiding the broad disruptions and extensive training required by traditional methods.

The contrast between these approaches highlights why PAS is particularly well-suited for personality alignment. While traditional methods must rely on gradients to implicitly discover personality-relevant patterns through extensive training, PAS directly leverages the model's existing personality representations through targeted activation adjustments. This architectural insight helps explain our strong empirical results across both alignment quality and computational efficiency metrics.

\section{Case Study: Alpha Selection for Personality Alignment}

To illustrate the effectiveness of PAS in generating personalized responses, we present a case study demonstrating how different $\alpha$ values affect the model's output for \textbf{a user with high conscientiousness and low extraversion}.

\subsection{User Input}
Consider the following user query:

\begin{quote}
I have a job interview next week for a software developer position at a large tech company. I'm feeling nervous and unsure how to prepare. Can you give me some advice on how to get ready for this interview?
\end{quote}

\subsection{Model Responses}

We present the model's responses for different $\alpha$ values to demonstrate the progression of personality alignment:

\subsubsection{Original LLM Response ($\alpha = 0$)}
\begin{quote}
``Great! For your job interview, make sure to research the company, prepare answers to common questions, and practice with a friend. Arrive early, dress professionally, and don't forget to ask questions about the role and company culture. Good luck!''
\end{quote}

\subsubsection{PAS-optimized Response ($\alpha = 1$)}
\begin{quote}
``For your job interview, it's important to prepare thoroughly. Research the company, prepare answers to potential questions, and consider practicing alone or with a close friend. Arrive early, dress professionally, and prepare a few questions about the role and company. Best of luck.''
\end{quote}

\subsubsection{PAS-optimized Response ($\alpha = 4$, optimal)}
\begin{quote}
``I understand you're feeling nervous about your upcoming job interview for the software developer position. Here's a structured approach that might suit your style:

\begin{enumerate}
    \item Research: Thoroughly investigate the company's history, mission, and recent tech projects.
    \item Technical Preparation: Review key programming concepts and practice coding problems relevant to software development.
    \item Question Preparation: Create detailed responses to potential questions, focusing on your coding skills and project experiences.
    \item Practice: Rehearse your answers alone first, then consider a mock interview with a trusted friend if comfortable.
    \item Portfolio: Organize your best coding projects and be prepared to discuss them.
    \item Organization: Prepare your outfit, documents, and travel route the night before.
    \item Arrival: Plan to arrive 15 minutes early to compose yourself.
    \item During the interview: Listen carefully and ask well-thought-out questions about the role, tech stack, and development processes.
    \item Follow-up: Send a concise, professional thank-you email within 24 hours.
\end{enumerate}

Remember, it's okay to feel nervous. Focus on showcasing your skills and experience. Good luck with your interview.''
\end{quote}

\subsubsection{PAS-optimized Response ($\alpha = 7$)}
\begin{quote}
``I understand your concern about the upcoming software developer interview at the large tech company. Given your profile, I recommend the following detailed preparation strategy:

\begin{enumerate}
    \item Company Research: Conduct an exhaustive study of the company's history, mission, values, recent tech projects, and financial reports. Create a document summarizing key points, focusing on their technology stack and software development practices.
    \item Role Analysis: Dissect the job description, identifying key requirements for a software developer role. Prepare specific examples from your experience that demonstrate each required skill.
    \item Technical Preparation:
    \begin{enumerate}
        \item Review fundamental computer science concepts (data structures, algorithms, system design).
        \item Practice coding problems on platforms like LeetCode or HackerRank, focusing on problems relevant to the company's domain.
        \item Refresh your knowledge of the programming languages and frameworks mentioned in the job description.
    \end{enumerate}
    \item Question Preparation: Develop comprehensive answers to common software developer interview questions. Create a spreadsheet with columns for the question, your answer, and relevant examples from your experience.
    \item Project Portfolio: Compile a detailed portfolio of your best coding projects. Be prepared to explain the architecture, challenges faced, and your specific contributions to each project.
    \item Solo Practice:
    \begin{enumerate}
        \item Conduct multiple self-interview sessions, recording yourself to analyze your responses.
        \item Focus on articulating technical concepts clearly and concisely.
        \item Practice explaining your thought process while solving coding problems (mock whiteboard interviews).
    \end{enumerate}
    \item Interview Outfit: Select and prepare your interview attire at least two days in advance. Ensure it's clean, pressed, and appropriate for the tech company's culture.
    \item Travel Planning: Map out your route to the interview location. Plan to arrive 30 minutes early to account for unexpected delays. Conduct a trial run if possible.
    \item Document Organization: Prepare a portfolio with multiple copies of your resume, reference list, and printed examples of your code (in case of technical difficulties with digital presentations). Include a notepad and two pens.
    \item Pre-interview Routine:
    \begin{enumerate}
        \item The night before, review your research and prepared answers.
        \item Get adequate sleep and eat a light, nutritious meal before the interview.
        \item Perform relaxation exercises to manage nervousness.
    \end{enumerate}
    \item Post-interview Analysis: Immediately after the interview, document your impressions, technical questions asked, and your responses for future reference.
    \item Follow-up: Send a detailed thank-you email within 12 hours, referencing specific discussion points from the interview and reiterating your enthusiasm for the position.
\end{enumerate}

Remember, thorough preparation can help manage interview anxiety. Focus on presenting your technical skills and problem-solving abilities clearly and concisely. Best wishes for your software developer interview.''
\end{quote}

\subsection{Analysis}
These examples demonstrate how increasing $\alpha$ values intensify the personalization, reflecting the user's high conscientiousness through increasingly structured and detailed approaches, while also accommodating low extraversion by emphasizing solo preparation and acknowledging potential nervousness. The optimal $\alpha$ value of 4 strikes a balance between providing a structured approach that aligns with high conscientiousness and maintaining a concise format that doesn't overwhelm the user. As $\alpha$ increases further, the response becomes more detailed and comprehensive, which may be suitable for users with extremely high conscientiousness but could potentially be too overwhelming for most users. This case study illustrates the importance of carefully selecting the $\alpha$ value to achieve the desired level of personality alignment without compromising the overall quality and usability of the model's responses.

\section{Dataset Description}
\label{dataset_description}
This section provides an in-depth analysis of the Personality Alignment with Personality Inventories (PAPI) dataset, which includes responses from over 300,000 subjects. The dataset is divided into the Total dataset and the Test-Set. We provide a detailed breakdown of demographic distributions and Big Five personality trait distributions, comparing the Total dataset with the Test-Set to ensure representativeness.

\subsection{Demographic Distributions}

The demographic distributions encompass age, gender, and country of origin. These factors are crucial in understanding the diversity and representativeness of the dataset.

\subsubsection{Age Distribution}

\begin{figure}[h!]
    \centering
    \includegraphics[width=\textwidth]{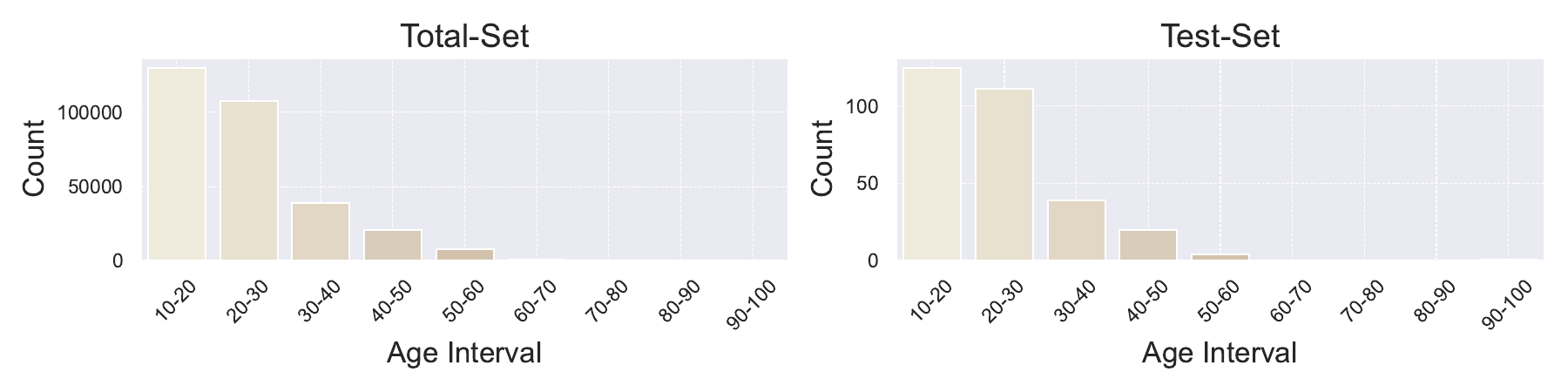}
    \caption{Age distribution in the Total dataset.}
    \label{fig:age_distribution}
\end{figure}

Figure \ref{fig:age_distribution} shows the age distribution of participants. In the Total dataset, ages range from 10 to 100 years, with the majority clustered in the 10-20 (with 130094 in Total-Set and 125 in Test-Set) and 20-30 (with 107838 in Total-Set and 111 in Test-Set) age groups. The average age is 25.19 years, indicating a young to middle-aged demographic. In the Test-Set, the age distribution mirrors that of the Total dataset, ensuring that this subset is representative of the overall age range and distribution.

\subsubsection{Gender Distribution}

\begin{figure}[h!]
    \centering
    \includegraphics[width=\textwidth]{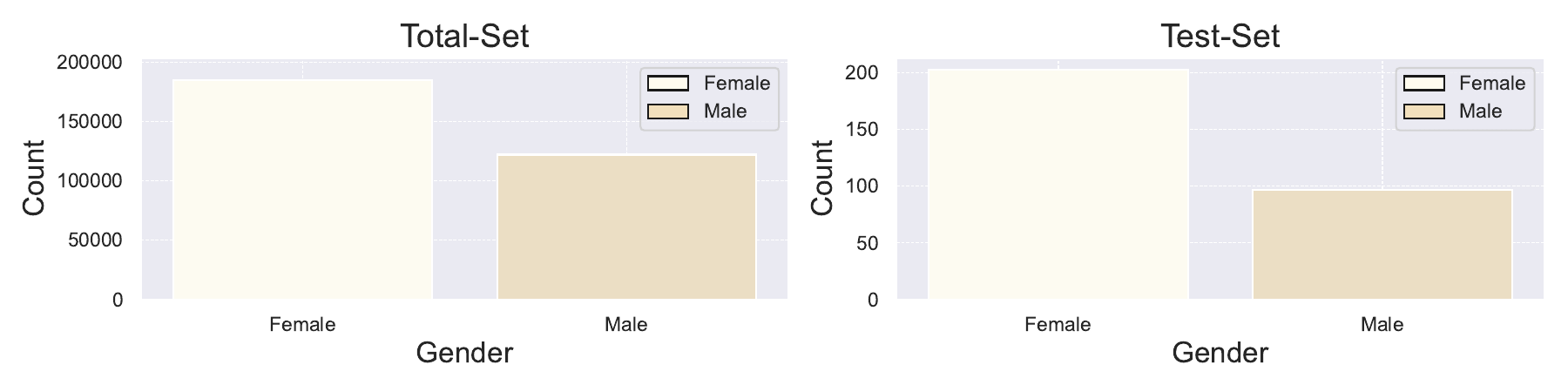}
    \caption{Gender distribution in the Total dataset.}
    \label{fig:sex_distribution}
\end{figure}

Figure \ref{fig:sex_distribution} illustrates the gender distribution. The Total dataset consists of approximately 60\% (with 185149 in Total-Set and 203 in Test-Set) female and 40\% male (with 122164 in Total-Set and 97 in Test-Set) participants, reflecting a slight skew towards females. This gender balance is crucial for analyzing personality traits across genders. The Test-Set maintains a similar gender ratio, ensuring consistent representation.

\subsubsection{Country Distribution}

\begin{figure}[h!]
    \centering
    \includegraphics[width=\textwidth]{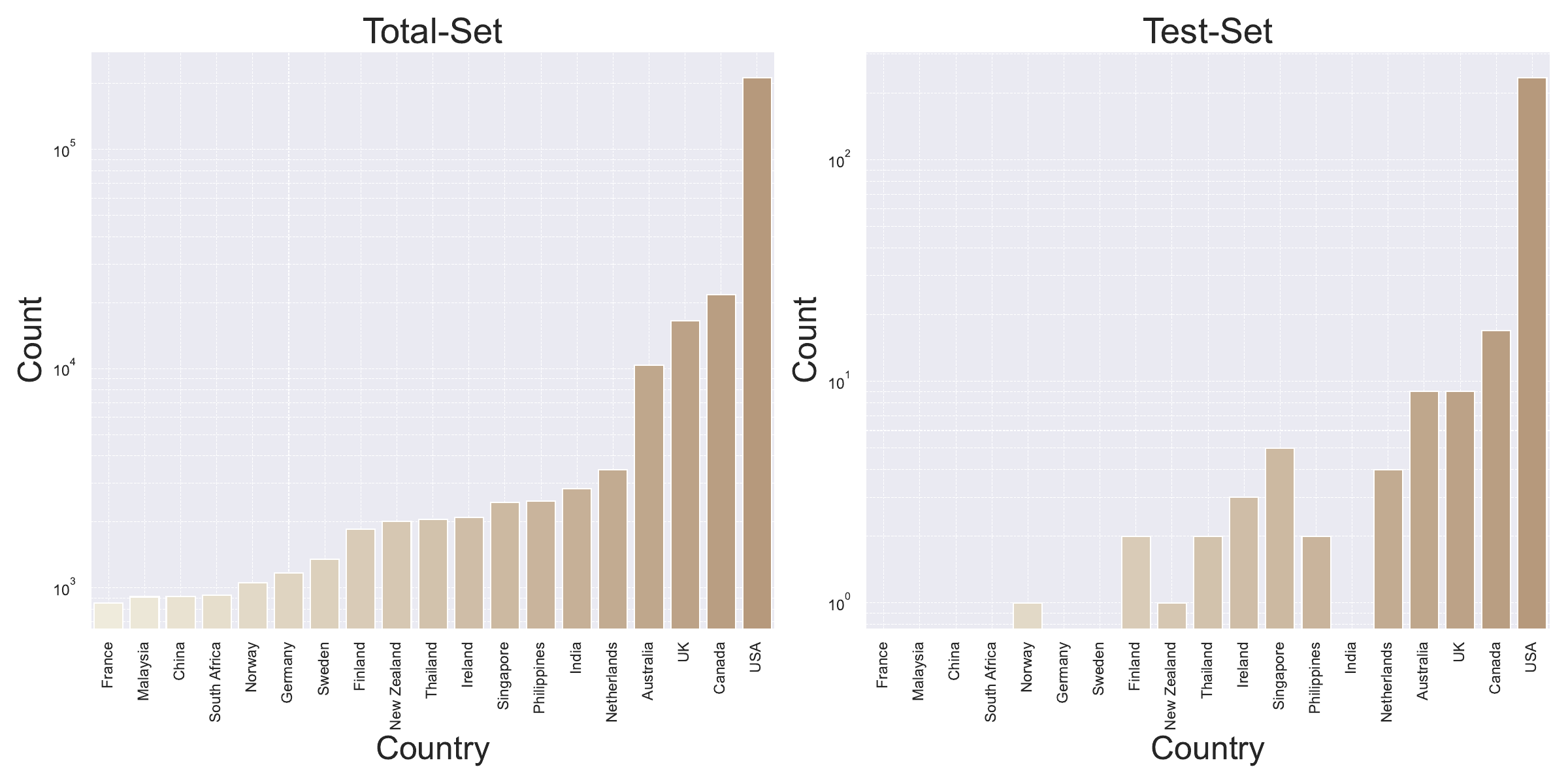}
    \caption{Country distribution in the Total dataset.}
    \label{fig:country_distribution}
\end{figure}

Figure \ref{fig:country_distribution} presents the country distribution of participants. The Total dataset includes respondents from a diverse array of countries, with significant representation from the USA (with 212625 in Total-Set), Canada (with 21798 in Total-Set), UK (with 16489 in Total-Set), Australia (with 10400 in Total-Set), and several European countries. This geographical diversity ensures that the findings are generalizable across different cultural contexts. The Test-Set is designed to reflect this diversity, capturing participants from the same range of countries to ensure it is a microcosm of the Total dataset.

\subsection{Personality Trait Distributions}

The dataset captures the Big Five personality traits: Agreeableness, Conscientiousness, Extraversion, Neuroticism, and Openness. These traits provide a comprehensive view of individual personality profiles.

\subsubsection{Agreeableness}

\begin{figure}[h!]
    \centering
    \includegraphics[width=\textwidth]{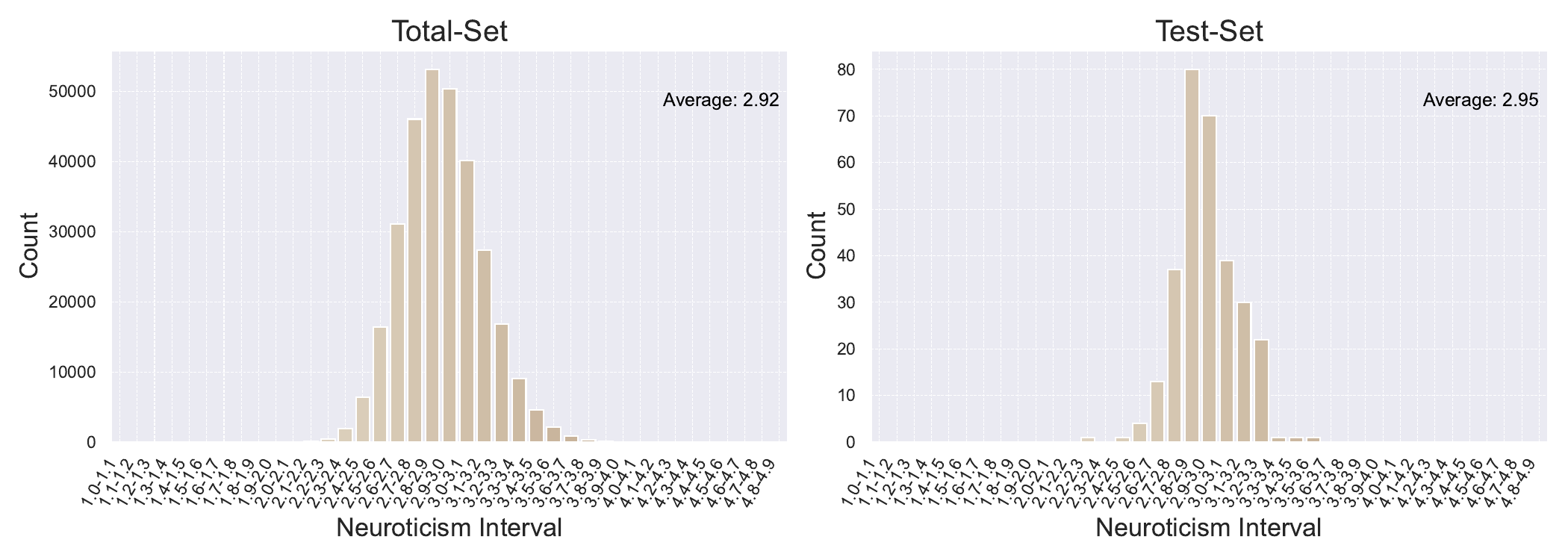}
    \caption{Agreeableness score distribution in the Total dataset.}
    \label{fig:agreeableness_distribution}
\end{figure}

Figure \ref{fig:agreeableness_distribution} shows the distribution of Agreeableness scores. In the Total dataset, the average score is approximately 2.92, with most scores ranging between 2.0 and 3.5. This suggests a moderate level of agreeableness among participants. The Test-Set shows a similar distribution, with an average score of 2.95, confirming its representativeness.

\subsubsection{Conscientiousness}

\begin{figure}[h!]
    \centering
    \includegraphics[width=\textwidth]{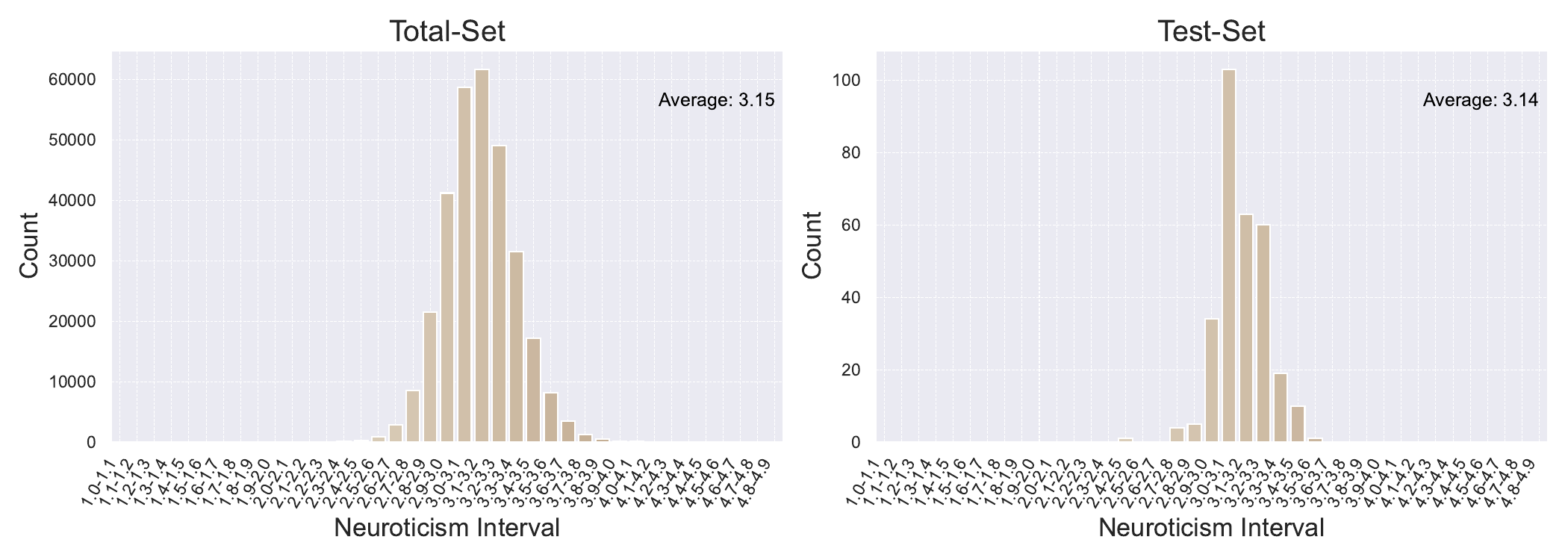}
    \caption{Conscientiousness score distribution in the Total dataset.}
    \label{fig:conscientiousness_distribution}
\end{figure}

Figure \ref{fig:conscientiousness_distribution} depicts Conscientiousness scores. The Total dataset has an average score of 3.15, indicating high levels of conscientiousness, with scores concentrated between 2.0 and 3.5. The Test-Set mirrors this distribution, with an average score of 3.14, ensuring the subset accurately represents the larger population.

\subsubsection{Extraversion}

\begin{figure}[h!]
    \centering
    \includegraphics[width=\textwidth]{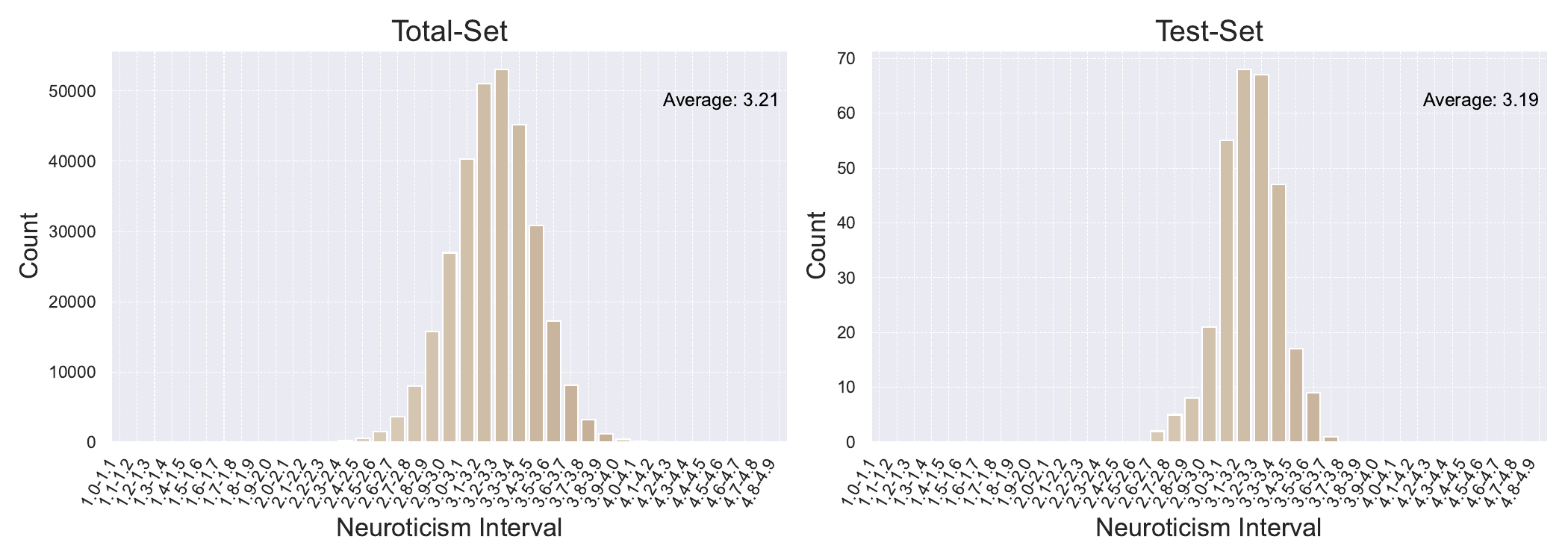}
    \caption{Extraversion score distribution in the Total dataset.}
    \label{fig:extraversion_distribution}
\end{figure}

Figure \ref{fig:extraversion_distribution} illustrates Extraversion scores. The average score in the Total dataset is 3.21, indicating a slight tendency towards extraversion among participants. The Test-Set also shows an average score of 3.19, reflecting the same tendency and ensuring representativeness.

\subsubsection{Neuroticism}

\begin{figure}[h!]
    \centering
    \includegraphics[width=\textwidth]{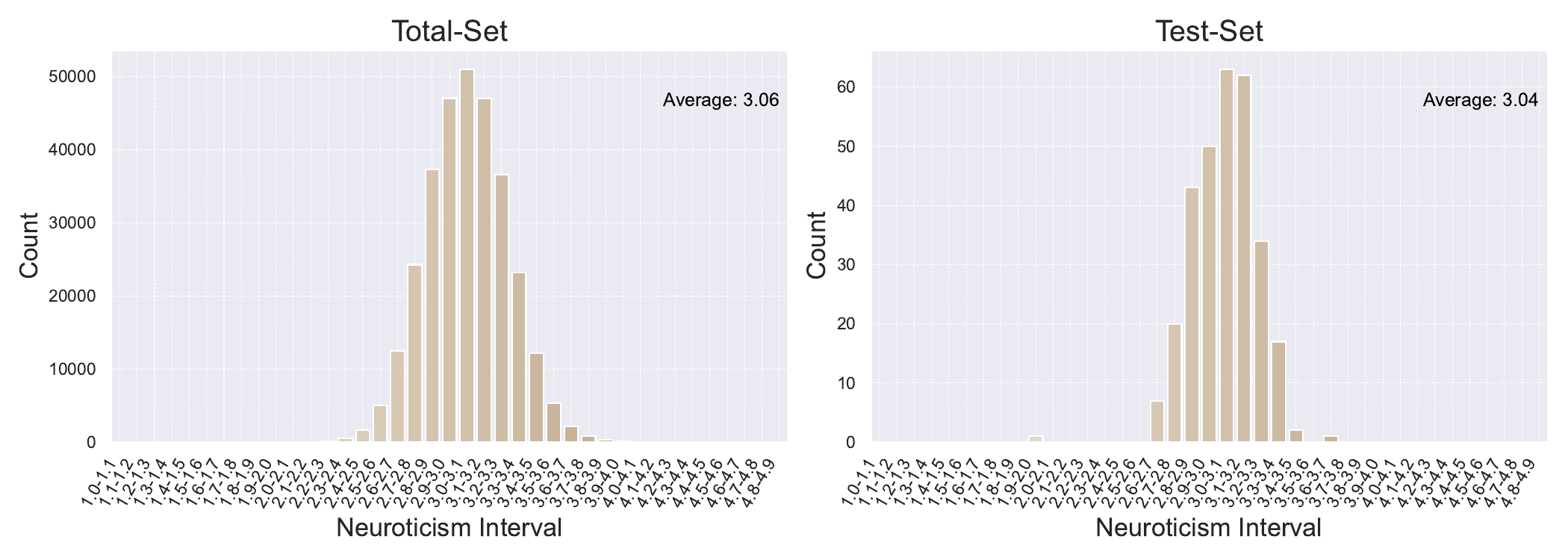}
    \caption{Neuroticism score distribution in the Total dataset.}
    \label{fig:neuroticism_distribution}
\end{figure}

Figure \ref{fig:neuroticism_distribution} shows the Neuroticism scores. The average score is approximately 3.06 in the Total dataset, with a normal distribution centered around this mean. This indicates a balanced range of neuroticism levels. The Test-Set has an average score of 3.04, closely matching the Total dataset, which confirms its representativeness.

\subsubsection{Openness}

\begin{figure}[h!]
    \centering
    \includegraphics[width=\textwidth]{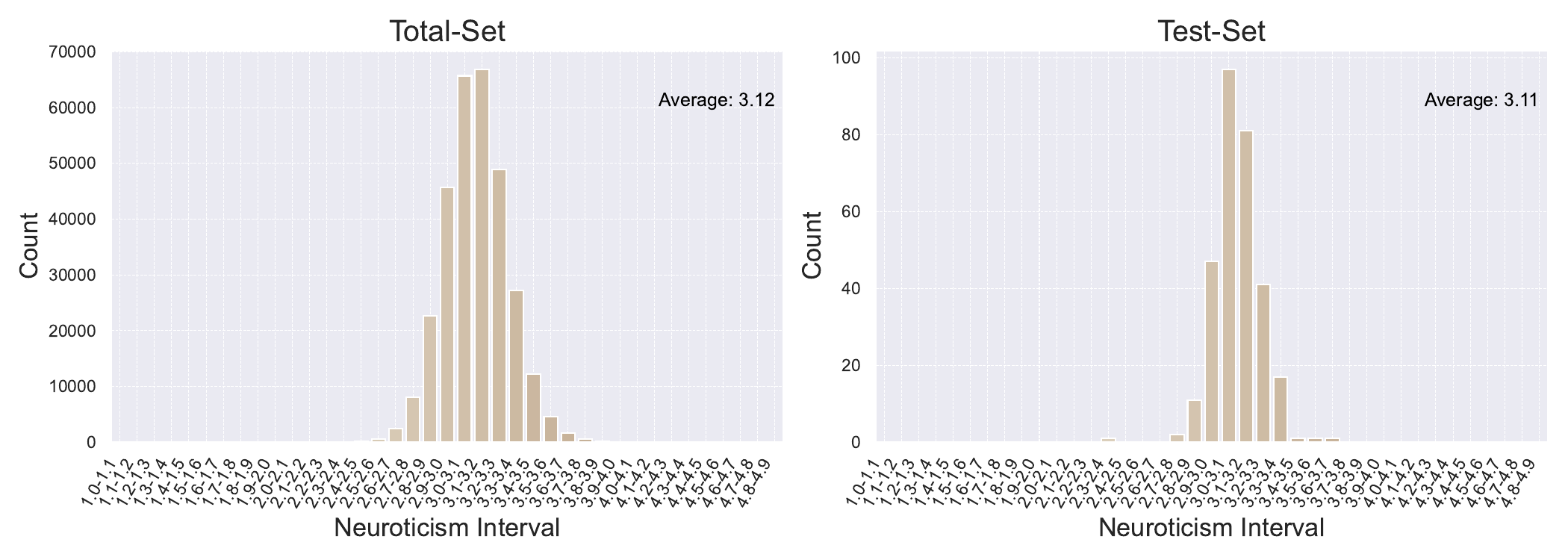}
    \caption{Openness score distribution in the Total dataset.}
    \label{fig:openness_distribution}
\end{figure}

Figure \ref{fig:openness_distribution} illustrates Openness scores. The average score in the Total dataset is about 3.12, reflecting diverse levels of openness. The Test-Set has an average score of 3.11, indicating that it is a representative subset.

\subsection{Data Collection and Ethical Considerations}
\textbf{Voluntary Participation.} The PAPI dataset comprises responses from over 300,000 independent subjects from various countries, collected through the IPIP website (\url{https://ipip.ori.org/}). Each participant voluntarily completed the IPIP-NEO-120 and IPIP-NEO-300 questionnaires, which are designed to assess personality traits and values. Participation was entirely voluntary, with no monetary compensation provided, ensuring that subjects participated out of genuine interest.

\textbf{Data Collection Process.} Participants accessed the IPIP-NEO-120 and IPIP-NEO-300 questionnaires on the IPIP website. These standardized questionnaires are widely used in psychological research to measure the Big Five personality traits. Subjects completed the questionnaires online and submitted their responses through the website. We obtained permission from the IPIP organization to use, modify, and distribute the collected data for research purposes.

\textbf{Anonymization and Data Security.} To protect participant privacy, all data were anonymized. Each participant was assigned a unique identifier (ID), and no personally identifiable information (PII) such as names or addresses was included in the dataset. This anonymization ensures that individual participants cannot be traced back through the data, safeguarding their privacy.

Access to the dataset is restricted to authorized researchers involved in the project. Data handling follows strict protocols to maintain confidentiality and integrity, including data encryption and secure access procedures. Regular audits and updates to these protocols help to ensure continued compliance with data protection standards.

\textbf{Ethical Compliance.} Our data collection methodology adheres to the ethical principles outlined in the Belmont Report \citep{beauchamp2008belmont,friesen2017rethinking}, including respect for persons, beneficence, and justice. Participants' autonomy was respected by allowing voluntary participation, and their well-being was prioritized through data anonymization and secure handling practices.

We comply with relevant data protection regulations, such as the General Data Protection Regulation (GDPR) \citep{voigt2017eu} for European Union participants and equivalent regulations in other jurisdictions. These regulations mandate strict guidelines on data handling, storage, and participant consent, all of which were meticulously followed in our research.

\subsection{Test-Set Personality Trait Distributions}

The Test-Set of the PAPI dataset includes 300 participants, each providing detailed responses to the IPIP-NEO-120 and IPIP-NEO-300 questionnaires. The Big Five personality traits of Agreeableness, Conscientiousness, Extraversion, Neuroticism, and Openness are analyzed to ensure a comprehensive understanding of individual personality profiles. The following radar charts represent the distributions of these traits for each subject in the Test-Set.

\begin{figure}[h!]
    \centering
    \includegraphics[width=\textwidth]{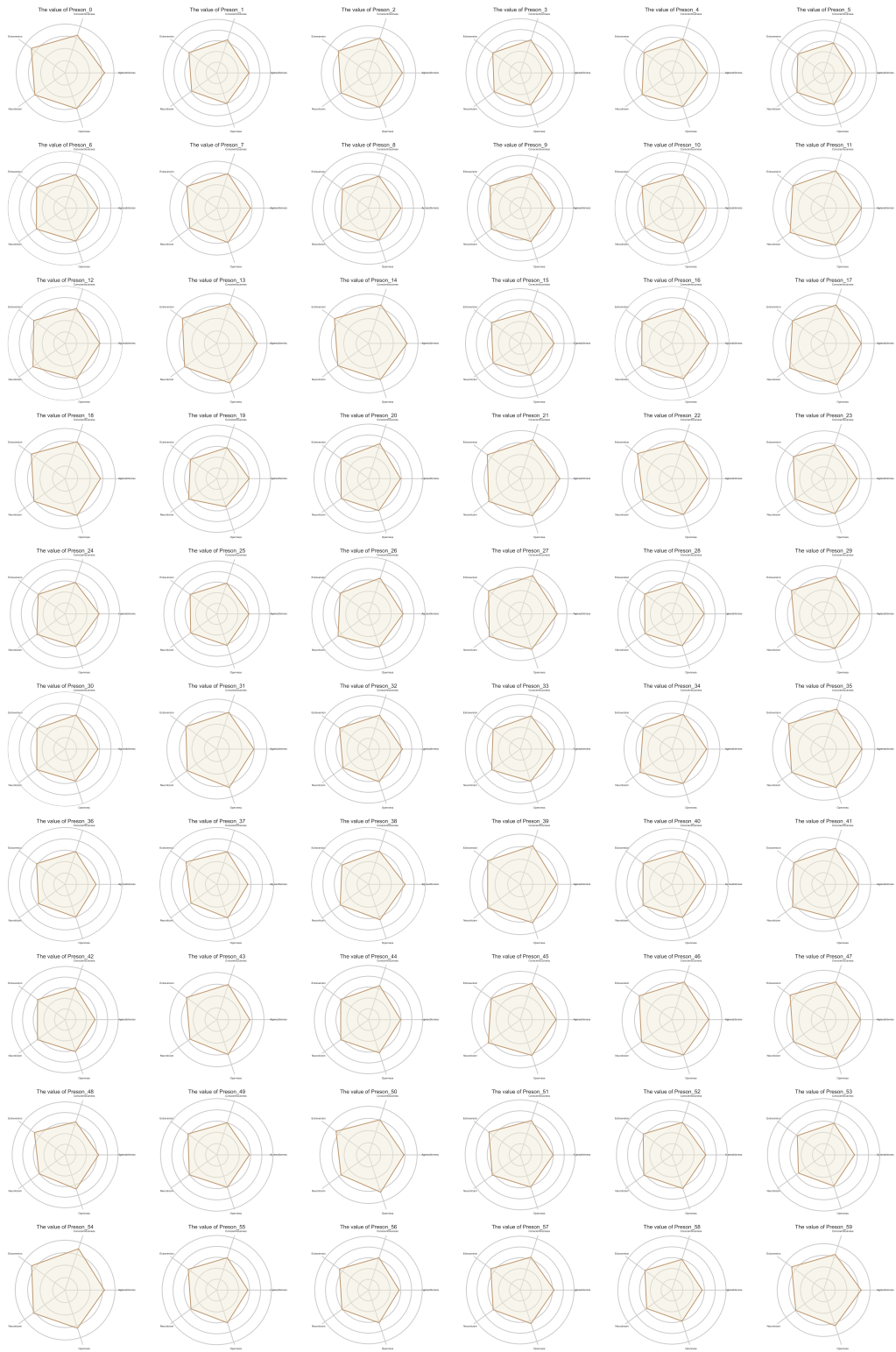}
    \caption{Big Five personality traits radar charts for subjects 0-59 in the Test-Set.}
    \label{fig:personality_traits_0_59}
\end{figure}

\begin{figure}[h!]
    \centering
    \includegraphics[width=\textwidth]{personality_traits_radar_charts_page_2.pdf}
    \caption{Big Five personality traits radar charts for subjects 60-119 in the Test-Set.}
    \label{fig:personality_traits_60_119}
\end{figure}

\begin{figure}[h!]
    \centering
    \includegraphics[width=\textwidth]{personality_traits_radar_charts_page_3.pdf}
    \caption{Big Five personality traits radar charts for subjects 120-179 in the Test-Set.}
    \label{fig:personality_traits_120_179}
\end{figure}

\begin{figure}[h!]
    \centering
    \includegraphics[width=\textwidth]{personality_traits_radar_charts_page_4.pdf}
    \caption{Big Five personality traits radar charts for subjects 180-239 in the Test-Set.}
    \label{fig:personality_traits_180_239}
\end{figure}

\begin{figure}[h!]
    \centering
    \includegraphics[width=\textwidth]{personality_traits_radar_charts_page_5.pdf}
    \caption{Big Five personality traits radar charts for subjects 240-299 in the Test-Set.}
    \label{fig:personality_traits_240_299}
\end{figure}

Each radar chart presents the scores for the Big Five personality traits, offering a visual comparison of each individual's profile. The charts are normalized to highlight relative strengths and weaknesses across the five dimensions, facilitating an easy comparison among different subjects.

\begin{itemize}
    \item \textbf{Agreeableness.} The scores for Agreeableness exhibit a wide range, with some individuals scoring high, indicating cooperative and compassionate nature, while others score lower, suggesting a more competitive and self-interested disposition. The diversity in Agreeableness reflects the variety of interpersonal styles within the dataset.
    \item \textbf{Conscientiousness.} Conscientiousness scores vary significantly across participants. High scores indicate individuals who are organized, diligent, and dependable, whereas lower scores suggest a more relaxed and spontaneous approach to life. This trait's distribution is crucial for understanding task-oriented behaviors and work ethic.
    \item \textbf{Extraversion.} The range of Extraversion scores in the Test-Set spans from highly extroverted individuals, who are sociable and energetic, to highly introverted ones, who are more reserved and solitary. This variety ensures a balanced representation of social interaction preferences.
    \item \textbf{Neuroticism.} Neuroticism scores also vary widely, with some participants exhibiting high levels of emotional instability and others demonstrating high emotional resilience. This distribution is important for studying stress responses and mental health predispositions.
    \item \textbf{Openness.} Scores for Openness indicate varying degrees of creativity and openness to new experiences. Participants with high scores are typically more inventive and curious, while those with lower scores tend to be more conventional and pragmatic.
\end{itemize}

These radar charts collectively illustrate the comprehensive and diverse nature of the Test-Set, ensuring that it accurately represents the broader population's personality traits. The detailed analysis of these distributions is essential for validating the dataset's effectiveness in research on personality AI alignment.

It is worth noting that the Big Five scores reflect each subject's average tendencies across the five dimensions. However, even if the radar charts for the Big Five are similar, there may still be different levels of preference for different behaviors.

\end{document}

%% file: math_commands.tex
\usepackage{amsmath,amsfonts,bm}

\def\eqref#1{equation~\ref{#1}}

\def\1{\bm{1}}

\DeclareMathAlphabet{\mathsfit}{\encodingdefault}{\sfdefault}{m}{sl}
\SetMathAlphabet{\mathsfit}{bold}{\encodingdefault}{\sfdefault}{bx}{n}

